\documentclass[11pt]{article}

\usepackage{acl}

\usepackage{times}
\usepackage{latexsym}

\usepackage[T1]{fontenc}

\usepackage[utf8]{inputenc}
\usepackage{booktabs}
\usepackage{microtype}

\usepackage{inconsolata}

\usepackage{graphicx}

\usepackage[most]{tcolorbox}

\tcbset {
  base/.style={
    bottomtitle=0.5mm,
    boxrule=0.2mm,
    colbacktitle=black!10!white, 
    coltitle=black, 
    fonttitle=\bfseries, 
    left=2.5mm,
    leftrule=0.2mm,
    right=2.5mm,
    title={#1},
    toptitle=0.75mm, 
    breakable
  }
}

\newtcolorbox{mainbox}[1]{ colframe=black, colback=white, base={#1} }

\newtcolorbox{subbox}[1]{
  colframe=white,
  base={#1}
}

\usepackage{comment}

\usepackage{hyperref}
\usepackage{url}
\usepackage{graphicx}
\usepackage[utf8]{inputenc} %
\usepackage[T1]{fontenc}    %
\usepackage{hyperref}       %
\usepackage{url}            %
\usepackage{makecell}      %
\usepackage{booktabs}       %
\usepackage{amsfonts}       %
\usepackage{nicefrac}       %
\usepackage{microtype}      %
\usepackage{booktabs}
\usepackage{amssymb}
\usepackage{multirow}
\usepackage{pifont}
\usepackage{array}
\usepackage{graphicx}  %
\usepackage[table]{xcolor}
\usepackage[utf8]{inputenc} %
\usepackage[T1]{fontenc}    %
\usepackage{hyperref}       %
\usepackage{url}            %
\usepackage{booktabs}       %
\usepackage{amsfonts}       %
\usepackage{nicefrac}       %
\usepackage{microtype}      %
\usepackage{amsmath}
\usepackage{amsthm}
\usepackage{amssymb}
\usepackage{changepage}
\usepackage{bbding}
\usepackage{longtable}
\usepackage{enumitem}
\usepackage{dashrule}
\usepackage{float}
\usepackage{graphicx}
\usepackage{subcaption}
\usepackage{caption}
\usepackage{bm}
\usepackage{todonotes}
\usepackage{enumitem}
\setlist[itemize]{noitemsep,topsep=0pt}
\newcommand{\blue}[1]{\textcolor{blue}{#1}}
\newcommand{\red}[1]{\textcolor{red}{#1}}
\definecolor{revisionpurple}{RGB}{126,47,142}

\newcommand{\mydashrule}{\\\hdashrule[0.5ex]{1.02\textwidth}{0.4pt}{1mm}\\}

\newcommand{\authorsep}[0]{\ \ }

\newcommand{\parahead}[1]{\vspace{1.5mm}\noindent\textbf{#1}.\ }

\title{PhysVidBench: Language-Grounded Evaluation of Physical Commonsense in Text-to-Video Models}

\author{Enes Sanli$^{1,2,*\dagger}$ \authorsep Baris Sarper Tezcan$^{3,*}$ \authorsep Erkut Erdem$^{2,4}$ \authorsep Aykut Erdem$^{1,2}$\\
$^1$Ko\c{c} University, Istanbul, Turkey \authorsep 
$^2$KUIS AI Center, Istanbul, Turkey \\
$^3$Purdue University, Indiana, USA \authorsep  $^4$Hacettepe University, Ankara, Turkey\\
$^*$\small{Equal contribution.}\\
$^\dagger$\small{\textbf{Correspondence:} \href{mailto:esanli25@ku.edu.tr}{\texttt{esanli25@ku.edu.tr}}}}

\begin{document}
\maketitle
\begin{abstract}
Text-to-video (T2V) models now produce striking visuals, yet they routinely violate everyday physics; objects float, tools are misused, and causal sequences break down. Existing benchmarks mostly probe isolated physical laws and rely on vision-language models to score videos directly, which entangles perception and reasoning in one judgment and correlates poorly with humans. We introduce PhysVidBench, a benchmark of 383 base prompts, expanded to 766 prompts with enriched variants and 4,123 manually reviewed QA items, together with a language-grounded evaluation framework that takes a different route: instead of asking a VLM ``Is this video physically correct?'', we caption the video, then ask a language model to answer prompt-derived yes/no physics questions using only the captions. This split between seeing and reasoning makes each score traceable to the captions that justify it and aligns more closely with human judgment than direct VLM scoring (Pearson $r=0.45$--$0.69$, compared with $0.21$--$0.47$ for the strongest direct VLM evaluator). To test the framework, we carefully curate a set of human-validated prompts spanning seven physical dimensions, with a focus on tool use and affordances -- areas largely absent from prior physics-focused benchmarks. Across 12 open and proprietary T2V systems, the best model reaches only 36.2\% average accuracy, and no model consistently handles everyday physical reasoning. The same pipeline can also guide iterative error-guided prompt refinement, improving CogVideoX-2B from 21.6 to 32.7 and CogVideoX-5B from 17.8 to 29.7 without retraining.
\end{abstract}

\begin{figure*}[!t]
    \centering
    \includegraphics[width=0.925\textwidth]{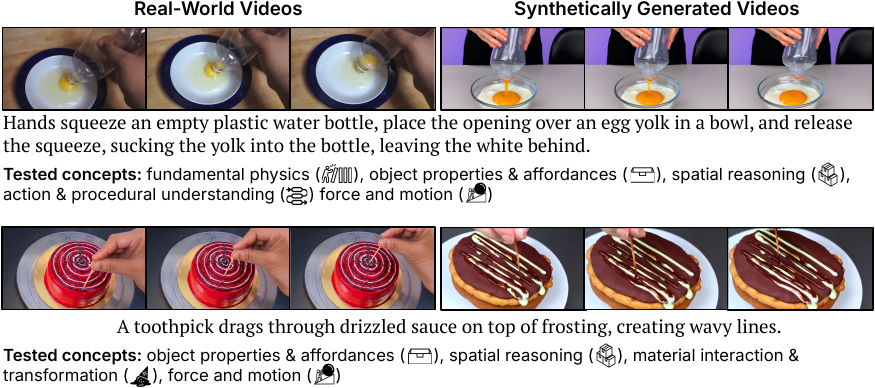}\vspace{-0.2cm}
    \caption{\textbf{Understanding physical commonsense in video generation.} Humans intuitively grasp how everyday interactions unfold: how objects respond to forces, how tools function, and how materials behave under manipulation. 
    \emph{PhysVidBench} tests whether T2V models share this commonsense, evaluating whether a generated video exhibits the physical consequences a prompt implies. Each row shows a real-world video of the correct behavior alongside a model's attempt at the same scenario. The generation looks plausible but misses the intended physical effect(s); the yolk is not drawn into the bottle, and the toothpick leaves no coherent pattern.
    }
    \label{fig:teaser}
\end{figure*}

\section{Introduction}
To separate an egg yolk from the white, one can squeeze an empty plastic bottle, place its mouth over the yolk, and release it so that suction lifts the yolk (Figure~\ref{fig:teaser}). This simple action requires commonsense about deformability, suction, spatial alignment, and temporal order. For T2V models, the challenge is not merely rendering a visually appealing scene, but realizing the physical interactions implied by language. Assessing whether they succeed is itself difficult. It requires checking whether the video evidence supports the prompt's causal and procedural expectations.

Recent T2V models produce increasingly realistic videos~\cite{agarwal2025cosmos, kong2024hunyuanvideo, yang2024cogvideox, chen2024videocrafter2, wang2025wan, kling2024, gen3_2024, veo2_2025} and are often discussed as possible world models for robotics, embodied AI, and simulation-based learning~\cite{he2025pre, hu2025simulating, wang2024worlddreamer, hu2023gaia, liu2024sora}. Yet visual realism does not imply physical understanding. Generated videos may preserve object appearance while failing to capture affordances, causal effects, material behavior, or the order of actions. For physical requirements expressible as discrete, visually verifiable propositions, this creates an evaluation problem that is naturally language-centered: the expected behavior is described linguistically, while the evidence must be extracted from video. 

Existing benchmarks provide important progress, but leave two gaps. First, many focus on isolated physical laws, motion plausibility, or predefined temporal changes rather than everyday, tool-mediated physical interactions~\cite{meng2025towards,bansal2025videophy,zheng2025vbench}. Second, direct VLM-based scoring can conflate perception, grounding, and reasoning, making it hard to inspect why a video was judged correct or incorrect. 

\begin{figure*}[!t]
    \centering
    \includegraphics[width=0.925\textwidth]{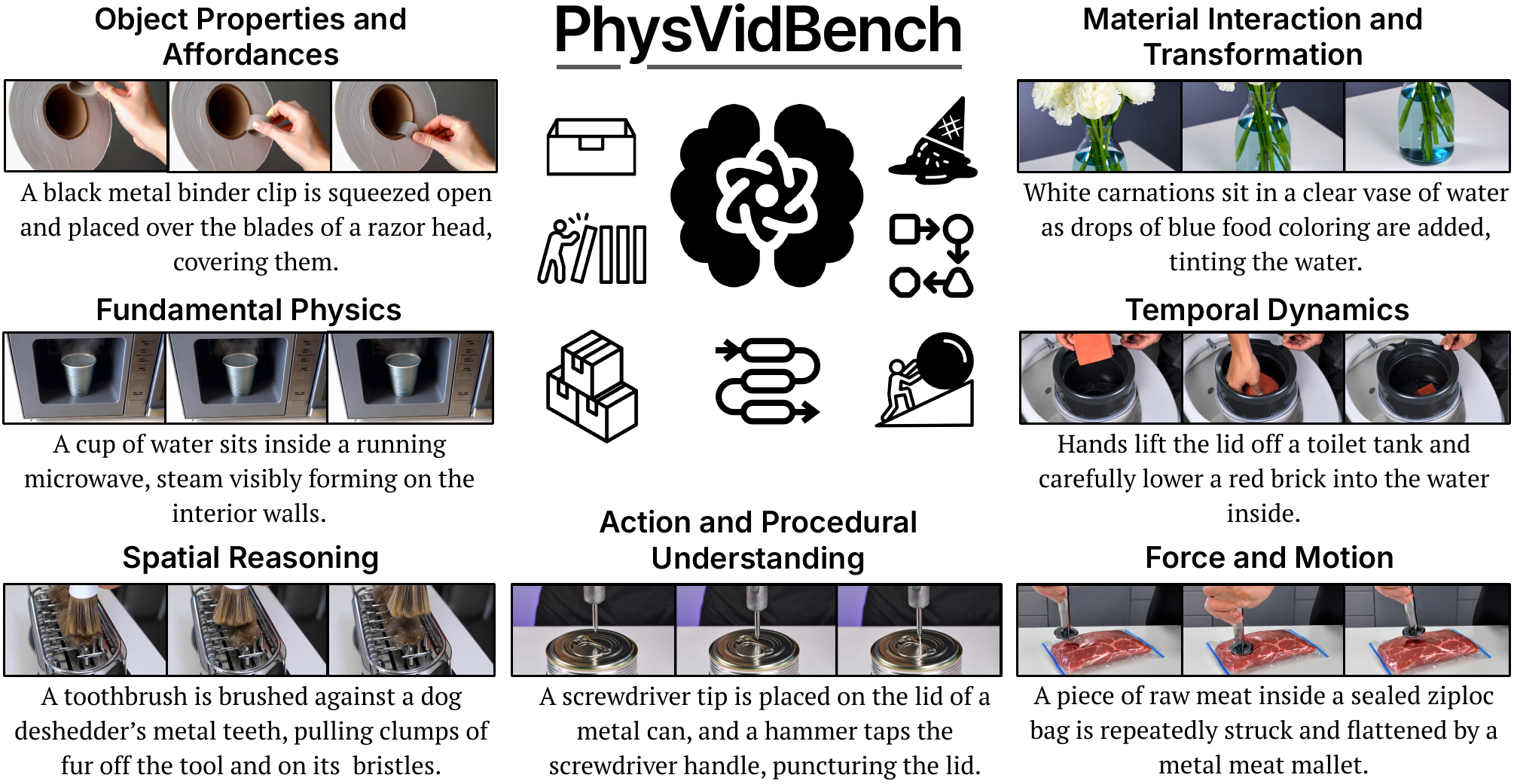}\vspace{-0.2cm}
    \caption{\textbf{Physical commonsense dimensions tested in PhysVidBench}, each illustrated with a video generated by Cosmos-14B. Prompts probe force and motion, object affordance, spatial containment, temporal progression, material interaction, and related requirements. The dimension shown for each example is illustrative rather than exclusive; prompts are annotated at the QA level and a single prompt typically contributes to multiple dimensions.}
    \label{fig:dimensions_overview}
\end{figure*}

We introduce \emph{PhysVidBench}, a benchmark and language-driven evaluation framework for physical commonsense in generated videos. It contains 383 prompts adapted from PIQA~\cite{bisk2020piqa}, covering everyday interactions such as tool use, container manipulation, material transfer, and causal object effects. Each prompt is paired with a detail-enriched variant that makes the expected physical behavior more explicit, and the prompts are organized around physical reasoning dimensions (Figure~\ref{fig:dimensions_overview}), from force and motion and object affordance to material interaction, temporal progression, and procedural understanding. These dimensions are multi-label rather than mutually exclusive: a prompt may belong to multiple dimensions through its reviewed QA items (see App.~\ref{app:e2e_example} for an end-to-end example).

The core of our approach is a \emph{caption-grounded question-answering evaluator}. Given a prompt, we derive yes/no questions about the expected physical behavior. A VLM captions the generated video, and an LLM answers using only this evidence. This decouples perception from reasoning: captions describe observable content, while the LLM checks whether they entail the required behavior. Unlike direct VLM judgments, the resulting decisions are traceable to explicit textual evidence, making evaluation more modular and interpretable.

\noindent We manually validate the generated questions and evaluate a diverse set of open- and closed-source T2V models. Across 12 systems, the best average accuracy is only 36.2\%. Our results show that current models still struggle with everyday physical commonsense, especially for spatial relations, temporal dynamics, material interactions, and tool-mediated causal effects. We further show that caption-grounded QA aligns more closely with human judgments than direct VLM scoring and generic automatic metrics. On the same 120 videos and human annotations, our evaluator achieves Pearson correlations of 0.45--0.69, compared with 0.21--0.47 for the strongest direct VLM evaluator.

\noindent Our contributions are listed as follows:
\begin{itemize}
    \item We introduce PhysVidBench, a PIQA-derived benchmark of 383 base prompts, 383 enriched variants, and 4,123 manually reviewed QA items that evaluates whether generated videos satisfy language-specified everyday physical commonsense.
    \item We formulate generated-video evaluation as caption-grounded QA, separating visual evidence extraction from language-based physical reasoning for physical requirements expressible as discrete, visually verifiable propositions.
    \item We benchmark multiple T2V models, both open and proprietary, and show persistent physical-reasoning failures, demonstrate stronger human alignment than direct VLM scoring, and show the same pipeline can guide training-free prompt refinement.
\end{itemize}
All 766 base and upsampled prompts, the 4,123 reviewed QA items, captioning, and judge prompts, scoring code, and documentation of the curation and evaluation procedures are available at \url{https://cyberiada.github.io/PhysVidBench/}.

\section{Related Work}
\parahead{Text-to-Video Generation and Its Evaluation} Text-to-video (T2V) models, such~as Cog-VideoX~\cite{yang2024cogvideox}, HunyuanVideo~\cite{kong2024hunyuanvideo}, VideoCrafter2~\cite{chen2024videocrafter2}, Cosmos~\cite{agarwal2025cosmos}, Wan~\cite{wang2025wan}, Sora~\cite{openai2024sora}, and Veo~\cite{veo2_2025}, now generate increasingly realistic and temporally coherent videos. Yet realism and prompt alignment do not guarantee that a video satisfies the physical interaction described by the prompt. Early metrics such as CLIP Score~\cite{hessel2021clipscore}, FVD~\cite{unterthiner2018towards}, and IS~\cite{salimans2016improved} capture coarse alignment or distributional quality, but are weak proxies for temporal consistency, motion plausibility, and physical validity. Recent benchmarks provide finer-grained evaluation~\cite{huang2024vbench,liu2024evalcrafter,liu2023fetv}, but remain limited when correctness hinges on tool use or physical effects.

\parahead{Vision-Language Models as Evaluators} An emerging trend is to use VLMs as automatic evaluators for generated content~\cite{huang2024vbench,zheng2025vbench}, and on the dimensions they target (e.g. visual quality, temporal smoothness, or object presence) VLM scores correlate well with human ratings. That said, physical commonsense is a harder regime. The evaluator must decide whether the video supports a language-implied causal event, not merely whether the relevant objects appear, and prior analysis places physical reasoning closer to hallucination-sensitive evaluation than to standard VLM benchmarks~\cite{chow2025physbench}. Folding perception, grounding, and reasoning into one model call then makes scores hard to interpret and prone to hallucination, superficial matching, or response collapse~\cite{huang2024visual}. In App.~\ref{app:videophy2_example}, we show such scoring can fail to produce meaningful judgments on physical prompts.

\begin{table*}[!t]
\footnotesize
\centering
\caption{\textbf{Physical commonsense video benchmarks.} PhysVidBench emphasizes tool use and affordances, compares base and detail-enriched prompts, and evaluates generated videos through caption-grounded Yes/No QA.\vspace{-0.2cm}}
\label{tab:physvidbench_comparison}
\renewcommand{\arraystretch}{1.15}
\begin{tabular}{@{}l@{$\;\;$}c@{$\;\;$}c@{$\;\;$}c@{$\;\;$}c@{}}
\toprule
\textbf{Feature} & \textbf{PhyGenBench} & \textbf{VBench-2.0} & \textbf{VideoPhy-2} & \textbf{PhysVidBench} (Ours)\\
\midrule
\# of Prompts & 160 & 1013 & 3940 & 383 \\
Tool \& Affordance Focus & {No} & {No} & {No} & {Yes} \\
Human Alignment Study & {Yes} & {Yes} & {Yes} & {Yes} \\
Upsampled Prompt Comparison & {Yes} & {No} & {Yes} & {Yes} \\
Automatic Evaluation & {Yes} & {Yes} & {Yes} & {Yes} \\
Human Feedback Type & Rating & Pairwise & Rating (1-5) & Yes/No QA \\
\bottomrule\vspace{-0.2cm}
\end{tabular}
\end{table*}

Prior work has decomposed generated-content evaluation into fine-grained questions or structured descriptions, including TIFA, DSG, Divide-Evaluate-Refine, LLMScore, T2VScore, and ETVA \citep{hu2023tifa,cho2024dsg,singh2023divide,lu2023llmscore,wu2024t2vscore,guan2025etva}. We do not claim question decomposition itself as novel; our key distinction is the intermediate caption-evidence layer, which separates visual evidence extraction from physical reasoning. Related caption-then-reason pipelines have also been effective in long-range video question answering \citep{zhang-etal-2024-simple}.

Instead, we decompose the judgment where, following the LLM-as-judge paradigm~\cite{zheng2023judging}, a captioning step sits between perception and reasoning, converting each video into caption evidence over which an LLM answers physics-grounded yes/no questions. Since the judge reasons over text rather than raw video, decisions remain traceable to explicit captions, reformulating T2V assessment as constrained caption-grounded QA and reducing reliance on opaque VLM judgments.

\parahead{Physical Commonsense Benchmarks and Reasoning} Physical commonsense has been studied in both language and vision-language models. PIQA tests everyday physical reasoning through goal-solution pairs involving tools, objects, and affordances, while PhysBench~\cite{chow2025physbench} and Impossible Videos~\cite{bai2025impossible} show that VLMs and video models still struggle with physical world knowledge and implausible events. Beyond understanding, several benchmarks probe physical commonsense in \emph{generated} video. VBench-2.0 evaluates dimensions such as state change, geometry, and motion rationality; VideoPhy and VideoPhy-2~\cite{bansal2024videophy,bansal2025videophy} focus on action-centric plausibility; PhyGenBench~\cite{meng2025towards} targets isolated laws such as buoyancy and friction; and Physics-IQ~\cite{motamed2025generativevideomodelsunderstand} and Cosmos-Reason1~\cite{nvidia2025cosmosreason1} evaluate broader physical and causal reasoning phenomena.

As summarized in Table~\ref{tab:physvidbench_comparison}, PhysVidBench departs from these benchmarks in three important ways. First, it focuses on everyday tool-mediated interactions and affordance reasoning, where physical correctness depends on how objects are used, transformed, and causally related. Second, it pairs each base prompt with a detail-enriched variant to study how explicit physical constraints affect generation, while using aggregate model performance to organize prompts into Medium, Hard, and Very Hard groups for analysis. Third, it casts evaluation as yes/no QA, treating physical commonsense not merely as a capability probed in video but as a structured, language-grounded evaluation problem. Please refer to App.~\ref{app:benchmark_comparison} for additional details.

\begin{table*}[!t]
\caption{\textbf{Physical commonsense ontology and prompt coverage in PhysVidBench.} The seven dimensions guide question generation and dimension-specific captioning. Coverage counts are non-exclusive: a prompt is counted toward every dimension for which it contains at least one manually reviewed QA item.\vspace{-0.2cm}}%
\renewcommand{\arraystretch}{1.25}
\centering
\small
\begin{tabular}{@{}>{\raggedright\arraybackslash}p{4cm}p{8.5cm}c@{}}
\toprule
\textbf{Commonsense Dimension} & \textbf{Definition} & {\textbf{\# Prompts Covered}} \\ \midrule
\textbf{Fundamental Physics} & Core physical principles such as energy conservation, causality, equilibrium, and state transitions in physical systems. & {349}\\
\textbf{Object Properties \& Affordances} & Inherent object attributes including material composition, rigidity, softness, and functional affordances like containment or support. & {371} \\

\textbf{Spatial Reasoning} & Spatial relations including position, geometry, occlusion, orientation, and fit between objects in a scene. & {363} \\

\textbf{Temporal Dynamics} & Timing, sequencing, and the causal structure of events over time, such as ordering, delays, and waiting. & {209} \\

\textbf{Action \& Procedural Understanding} & Structured, goal-driven sequences of actions involving procedural steps toward task completion. & {335} \\

\textbf{Material Interaction \& Transformation} & Responses of materials to external forces or processes including transformations like melting, freezing, breaking, or chemical change. & {288} \\

\textbf{Force and Motion} & Physical interactions governed by forces, including motion, pushing, pulling, lifting, and properties like inertia. & {370} \\
\bottomrule
\end{tabular}
\label{table:ontology}
\end{table*}

\section{Benchmark}
\label{sec:benchmark}
PhysVidBench targets everyday physical commonsense: whether objects are used according to their affordances, actions unfold in the correct order, and physical effects such as suction, cutting, folding, pouring, or material deformation are actually realized. This makes the benchmark both a video-generation task and a language-mediated evaluation problem. Prompts define the expected behavior, and generated videos are assessed through caption-grounded question answering.

We build PhysVidBench from PIQA~\cite{bisk2020piqa}, a dataset of everyday goal-solution pairs designed to test physical commonsense, mining it for the subset whose solutions involve tool use and non-obvious object affordances rather than abstract physical laws or synthetic scenes (App.~\ref{app:benchmark_design}). PIQA contains roughly 21K examples. Using the labeled train and validation splits yields more than 17K correct goal-solution pairs, from which we retain 383 final prompts. We organize these scenarios along a seven-dimensional physical commonsense ontology (Table~\ref{table:ontology}) spanning fundamental physics, object properties and affordances, spatial reasoning, temporal dynamics, action and procedural understanding, material interaction and transformation, and force and motion. The ontology serves two roles: it guides prompt selection and structures the evaluation pipeline, where questions and dimension-specific captions target the same physical reasoning dimensions.

\subsection{Prompt Construction}
\label{sec:prompt_construction}
\begin{figure*}[!t]
    \centering
    \includegraphics[width=0.89\textwidth]{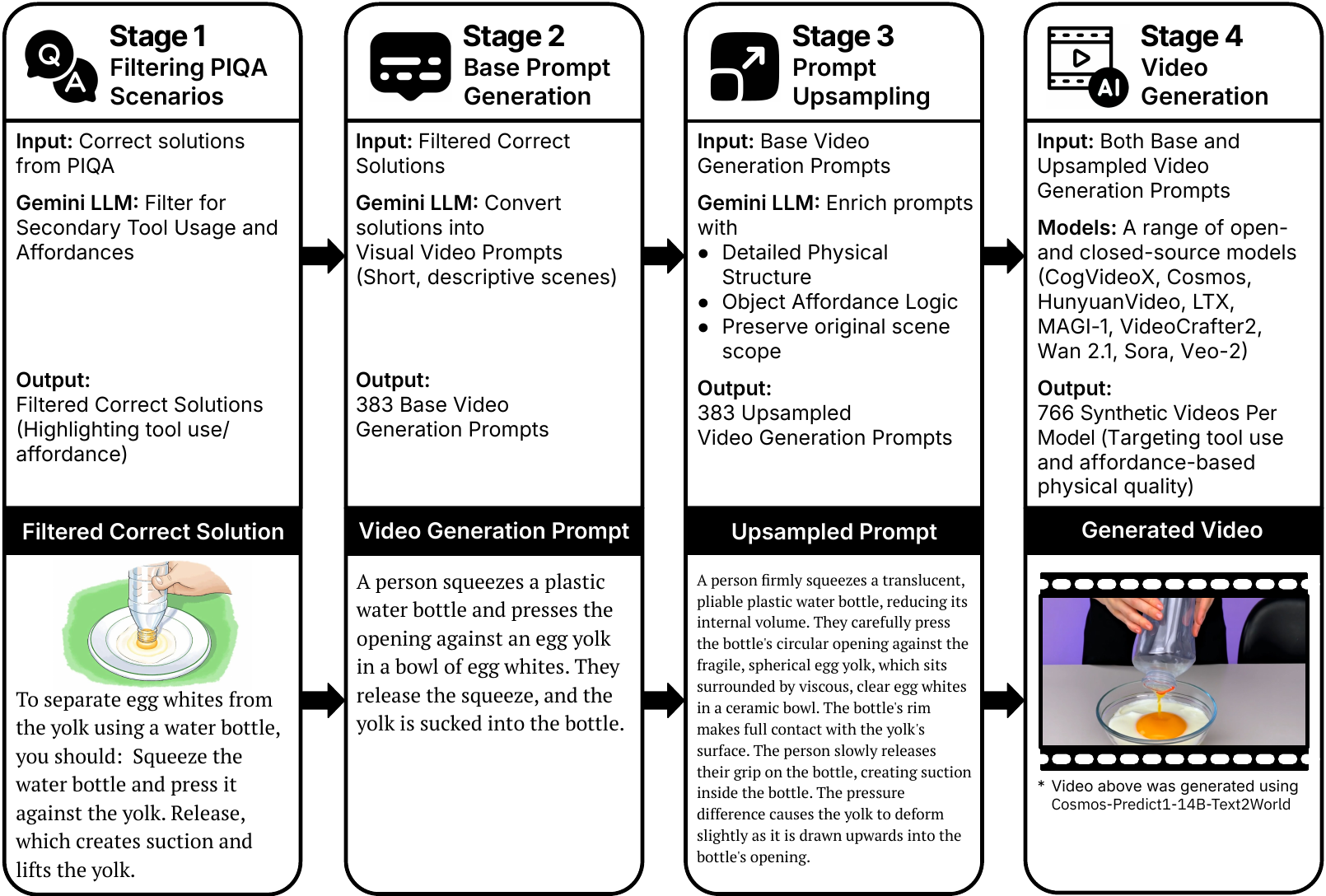}\vspace{-0.2cm}
    \caption{\textbf{Our prompt construction pipeline.} PIQA goal-solution pairs are filtered for tool use and non-obvious affordances, converted into video prompts, and expanded into detail-enriched variants before T2V generation.\vspace{-0.2cm}
    }
    \label{fig:vidgen_pipeline}
\end{figure*}

Each prompt describes a short, concrete action that can be directly visualized and reasoned about. Construction proceeds in three stages: selecting and filtering physically grounded goal-solution pairs, generating base video prompts, and enriching them to surface the underlying physical properties and causal structure (Figure~\ref{fig:vidgen_pipeline}).

\parahead{Stage 1: Filtering PIQA Scenarios} Using Gemini-2.5-Pro-Preview-05-06~\cite{gemini2025} (hereafter referred to as Gemini 2.5 Pro), we retain PIQA pairs whose labeled-correct solution involves secondary tool use or non-obvious object affordances, such as using a credit card to scrape ice or a paperclip to eject a SIM tray. This selects for interactions where physical correctness depends not on recognizing objects, but on understanding how their properties enable a particular action.

\begin{figure*}[!t]
    \centering   \includegraphics[width=0.95\textwidth]{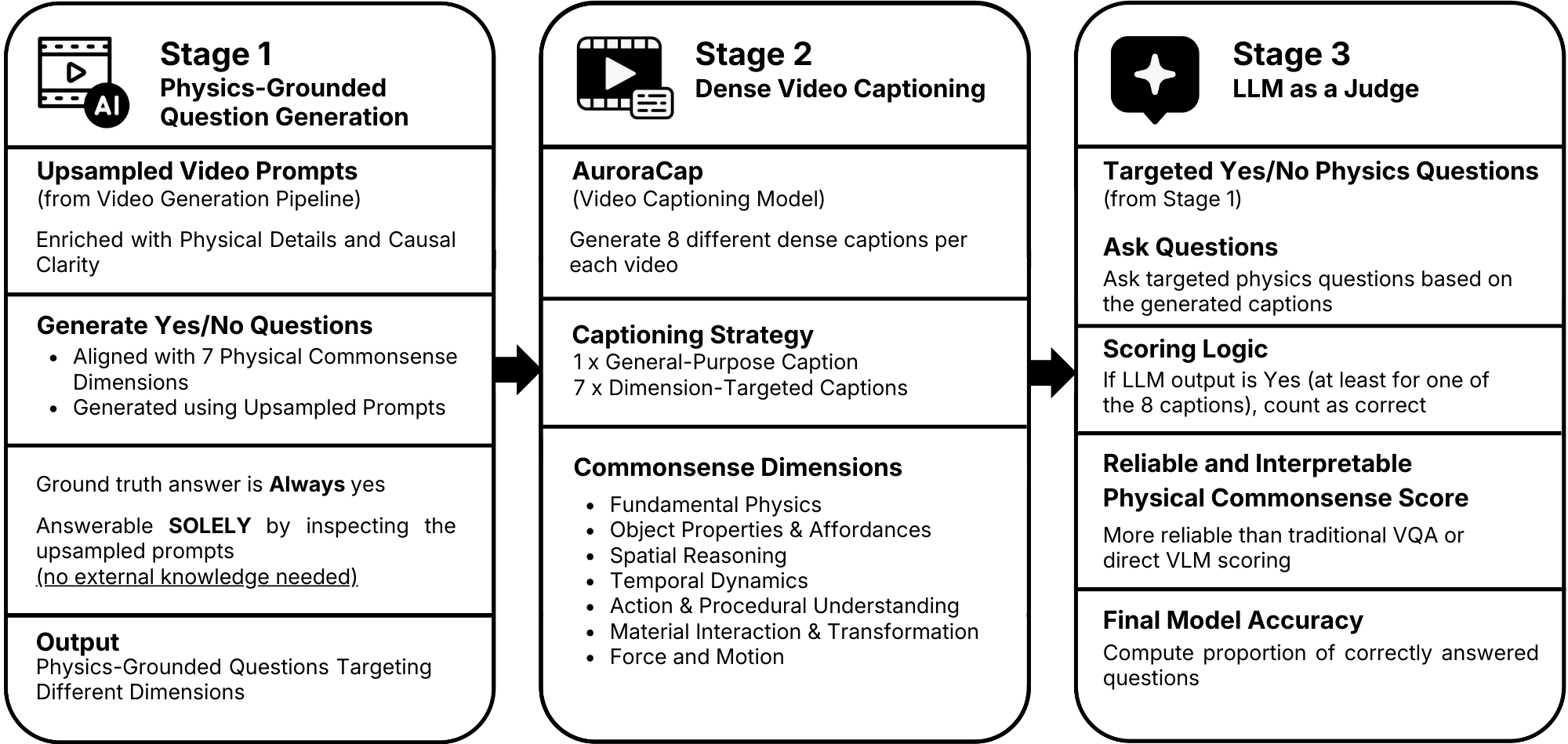}\vspace{-0.2cm}
    \caption{\textbf{Caption-grounded QA evaluation.} PhysVidBench converts each prompt into physical yes/no questions, captions each generated video with general and dimension-specific descriptions, and uses an LLM judge to answer using only this textual evidence.}
    \label{fig:eval_pipeline}
\end{figure*}

\noindent\textbf{Stage 2: Base Prompt Generation}.
Each retained pair is converted into a concise video prompt using Gemini 2.5 Pro. Prompts must describe realistic, self-contained, directly observable actions such as cutting, lifting, folding, or pouring; non-visual states, abstract intentions, and scenarios that cannot be shown in a short video are excluded. This yields 383 base prompts spanning more than 1,200 distinct objects and over 600 manipulation actions.\footnote{Object and action counts are obtained by parsing prompts with Gemini 2.5 Pro; see App.~\ref{app:dataset-diversity}.}

\noindent\textbf{Stage 3: Prompt Upsampling}. We pair each base prompt with a detail-enriched variant, following prompt upsampling methods~\cite{xue2024phyt2v}. Enrichment makes \emph{implied} physical properties and causal structure explicit, for instance naming the suction or deformation an action already entails, while preserving the original scene scope; the prompts include instructions to avoid new objects, events, or speculative consequences. The final set contains 383 base and 383 enriched prompts (766 total).\footnote{All prompts are manually reviewed to remove hallucinated, unsafe, inappropriate, or visually underspecified cases before T2V generation.}

The prompts were manually validated by four authors, and each item was independently reviewed by two authors. Further curation details are provided in App.~\ref{app:benchmark_design}.

\subsection{Evaluation Pipeline}
\label{sec:evaluation_pipeline}
PhysVidBench evaluates generated videos with the caption-grounded QA pipeline in Figure~\ref{fig:eval_pipeline}. Rather than scoring a video with a single VLM call, we separate visual grounding from textual reasoning. A video is first converted into caption evidence, and an LLM then answers physics-grounded yes/no questions using only that evidence (App.~\ref{app:captioning_qa}). This turns generated-video evaluation into a constrained language-reasoning problem whose decisions can be traced back to explicit captions.

\parahead{Stage 1: Physics-Grounded Question Generation} For each enriched prompt, we generate targeted yes/no questions about the expected physical behavior.  Each question is posed so that the correct answer is always ``yes''. It asks whether a required event, relation, affordance, or transformation occurs. This is a deliberate design choice. A physical interaction can fail in unboundedly many ways, so enumerating ``no'' cases yields ill-defined targets, whereas asking whether the intended effect is present gives each question a single, well-grounded positive answer.

If a prompt requires an object to fall, we ask whether the falling is supported by the video, rather than enumerating the many ways a generation might violate physics. Because the questions are derived from the enriched prompt, both base and enriched videos are evaluated against the same physical expectations, making their scores directly comparable. The process yields 4{,}123 questions, averaging 11 per video.\footnote{All questions are manually reviewed for clarity, answerability, and visual grounding; about 5\% are discarded because they refer to non-visual details or to information not shared by the base and enriched prompts.}

\begin{table*}[!t]
\caption{
\textbf{Performance by physical commonsense dimension.} 
{Accuracy is reported on upsampled prompts; parentheses show the change when switching to base prompts.} Dimensions follow Table~\ref{table:ontology}: \textbf{AU} = Action \& Procedural Understanding, \textbf{FM} = Force and Motion, \textbf{FP} = Fundamental Physics, \textbf{MT} = Material Interaction \& Transformation, \textbf{OP} = Object Properties \& Affordances, \textbf{SR} = Spatial Reasoning, and \textbf{TD} = Temporal Dynamics. {Bold and underline mark the best and second-best methods within each column.}}
\centering
\scriptsize
\resizebox{\textwidth}{!}{%
\begin{tabular}{@{}l@{$\;\;$}c@{$\;\;$}c@{$\;\;$}c@{$\;\;$}c@{$\;\;$}c@{$\;\;$}c@{$\;\;$}c@{$\;\;$}c@{}}
\toprule
\textbf{Model} 
& \textbf{AU} & \textbf{FM} & \textbf{FP} & \textbf{MT} & \textbf{OP} & \textbf{SR} & \textbf{TD} & \textbf{Average} \\
\midrule
LTX-Video             & 20.7 (-4.2) & 19.9 (-3.5) & 18.6 (-2.5) & 16.9 (-1.6) & 25.0 (-4.7) & 17.3 (-2.3) & 13.7 (-4.6)  & 21.0 (-3.0) \vspace{0.09cm}\\
VideoCrafter2         & 17.7 (-1.4) & 19.4 (-2.9) & 23.3 (-3.3) & 18.8 (-2.9) & 26.6 (-2.1) & 15.8 (-0.2) & 12.6 (-2.7)  & 22.1 (-1.7) \vspace{0.09cm}\\
CogVideoX (2B)        & 22.6 (-1.8) & 24.6 (-4.5) & 23.8 (-2.7) & 22.6 (-0.3) & 28.9 (-3.5) & 19.6 (-1.3) & 16.8 (-1.5)  & 25.6 (-3.1) \\
CogVideoX (5B)        & 18.9 (-0.8) & 19.6 (-3.3) & 21.3 (-3.2) & 18.1 (-1.4) & 25.6 (-1.3) & 17.9 (-0.5) & 12.6 (+0.8) & 21.0 (-1.5) \vspace{0.09cm}\\
Wan2.1 (1.3B)         & 30.1 (-3.5) & 29.0 (-5.0) & 28.5 (-2.8) & 28.9 (-1.9) & 35.2 (-3.6) & 27.4 (-3.1) & 18.7 (+0.0) & 30.6 (-3.1) \\
Wan2.1 (14B)          & 33.6 (-3.5) & 31.6 (-1.4) & \underline{32.6} (-1.3) & 32.9 (-3.5) & 39.0 (-3.9) & 29.6 (-1.3) & \textbf{23.7} (-1.5) & 34.1 (-1.9) \vspace{0.09cm}\\
MAGI-1                & 27.3 (-5.9) & 26.6 (-4.0) & 29.4 (-1.7) & 28.5 (-5.1) & 37.4 (-5.0) & 30.4 (-5.0) & 19.1 (-8.0) & 32.6 (-5.3) \vspace{0.09cm}\\
HunyuanVideo          & 26.7 (-0.5) & 26.2 (+0.2) & 28.1 (-0.1) & 32.3 (-4.8) & 36.1 (-1.9) & 23.9 (+2.1) & \underline{21.8} (-5.3) & 30.0 (-0.9) \vspace{0.09cm}\\
Cosmos (7B)           & 32.5 (-8.2) & 32.8 (-9.3) & \textbf{33.6} (-8.9) & \underline{36.1} (-10.6) & \underline{40.1} (-10.6) & \underline{31.6} (-10.0) & 21.4 (-8.4) & \underline{35.0} (-8.3) \\
Cosmos (14B)          & \textbf{35.6} (-5.9) & \underline{33.3} (-6.1) & 31.7 (-3.9) & \textbf{37.4} (-5.9) & \textbf{40.9} (-9.6) & \textbf{32.5} (-5.3) & 21.0 (-0.4) & \textbf{36.2} (-7.1) \vspace{0.09cm}\\
\midrule
Sora                  & 28.6 (+2.4) & 30.0 (-1.1) & 29.9 (+0.2) & 33.1 (+0.3) & 37.0 (-1.1) & 24.8 (+1.7) & 16.7 (+2.6) & 31.4 (+0.3) \\
Veo-2                 & \underline{34.9} (-0.7) & \textbf{33.8 }(-2.3) & 31.7 (-2.7) & 36.0 (-3.1) & 38.4 (-1.5) & 29.7 (+1.0) & 19.5 (+2.5) & 34.8 (-0.7) \\
\bottomrule
\end{tabular}
}
\label{tab:prompt_comparison}
\end{table*}

\parahead{Stage 2: Dense Video Captioning} Each generated video is captioned with AuroraCap~\cite{chai2024auroracap}. For every video, we produce one general-purpose caption and seven dimension-specific captions aligned with the ontology, each prompting the captioner to describe evidence relevant to one dimension, such as object affordances, temporal progression, material behavior, or force-based interactions. Together, these eight captions form the textual evidence available to the judge.

\parahead{Stage 3: LLM as a Judge} We use Gemini-2.5-Flash as the LLM judge. For each question, the judge receives all eight captions and the yes/no question; it does not see the original video or prompt -- the questions alone carry the prompt's physical expectations. Because the eight captions emphasize different physical aspects, we count a question correct when \emph{any} caption provides sufficient evidence for a ``yes'', so that a correct video is not penalized when a single caption happens to omit the relevant cue. Final performance is the proportion of questions answered correctly.

By decomposing evaluation into question generation, caption grounding, and LLM judging, our benchmark avoids conflating perception, grounding, and reasoning in a single VLM decision. The resulting scores measure whether generated videos contain captionable evidence for the physical behavior specified by language, making the benchmark scalable and its judgments interpretable.

For a complete end-to-end example, see App.~\ref{app:e2e_example}.

\section{Experimental Setup}
\parahead{Models} We evaluate recent open-source and proprietary text-to-video (T2V) models, VideoCrafter2 \cite{chen2024videocrafter2}, CogVideoX-2B/5B~\cite{yang2024cogvideox}, Wan2.1-1.3B/14B~\cite{wang2025wan}, Cosmos-7B/14B~\cite{agarwal2025cosmos}, HunyuanVideo~\cite{kong2024hunyuanvideo}, LTX-Video~\cite{HaCohen2024LTXVideo}, MAGI-1~\cite{magi1}, Sora~\cite{openai2024sora}, and Veo-2~\cite{veo2_2025}. 
Open-source inference runs on a single NVIDIA A40 GPU; proprietary outputs are generated through their public interfaces.

\begin{figure*}[!t]
    \centering
    \vspace{-0.4cm}  \includegraphics[width=0.975\textwidth]{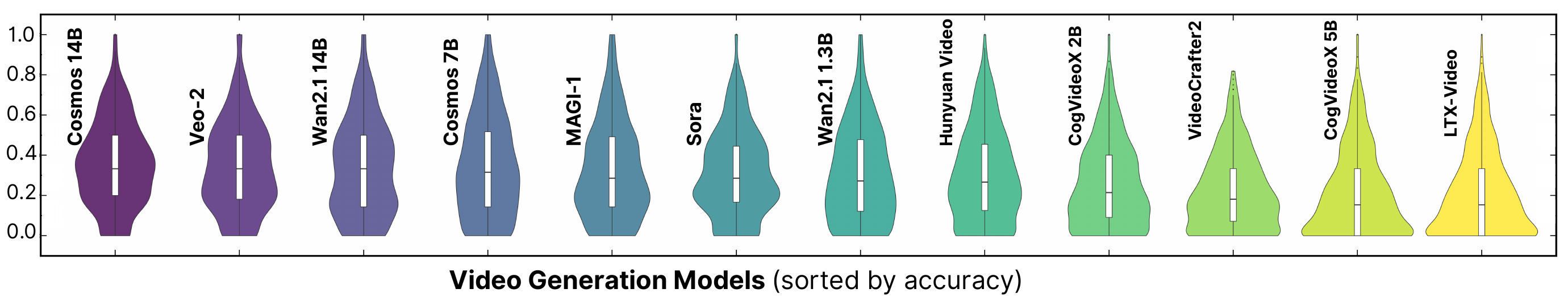}\vspace{-0.25cm}
    \caption{\textbf{Performance comparison of video generation models on PhysVidBench.} Each violin plot represents a distinct model and depicts the distribution of accuracy scores across all prompts.}
    \label{fig:perf_comparison}
\end{figure*}
\parahead{Model-based difficulty grouping}
We partition PhysVidBench into \emph{medium}, \emph{hard}, and \emph{very hard} subsets to analyze how performance varies with physical-reasoning difficulty. We omit an ``easy'' tier since all retained prompts require nontrivial physical commonsense. As a difficulty proxy, we use the mean caption-grounded QA accuracy of four high-performing open-source models (Cosmos, HunyuanVideo, MAGI-1, Wan2.1), so that the estimate reflects difficulty shared across systems rather than any single model's behavior. Prompts scoring above 40\%, 20--40\%, and below 20\% are assigned to the medium, hard, and very hard subsets (123, 160, and 100 prompts).  We use these performance-derived groups as an analysis tool for comparing model behavior across increasingly challenging prompts.

\section{Results and Discussion}
Table~\ref{tab:prompt_comparison} reports accuracy across the seven dimensions, and Figure~\ref{fig:perf_comparison} shows the distribution of per-prompt accuracies. The strongest open-source models, notably Cosmos-14B and Wan2.1-14B, obtain the highest averages, but performance is highly variable across prompts and dimensions. Proprietary models such as Sora and Veo-2 are competitive in some settings but do not consistently dominate the benchmark. No evaluated model reliably handles the full range of everyday physical commonsense in PhysVidBench. Accuracy also differs across the model-based difficulty groups (Table~\ref{tab:accuracy_comparison}, App.~\ref{app:failure_modes}).

\begin{table}[!t]
\caption{
\textbf{Accuracy across model-based difficulty groups.} {Values are caption-grounded QA accuracy percentages on upsampled prompts. The values inside the parentheses show the change when switching to base prompts.}
}
\centering
\footnotesize
\begin{tabular}{@{}lc@{$\;\;\;$}c@{$\;\;\;$}c@{}}
\toprule
\textbf{Model} & \textbf{Medium} & \textbf{Hard} & \textbf{Very Hard} \\
\midrule
LTX-Video           & 41.2 (-1.6) & 22.8 (-4.4)  & 9.1 (-1.2)  \vspace{0.09cm}\\
VideoCrafter2         & 44.5 (-4.3) & 23.1 (-1.1) & 10.7 (-1.6)   \vspace{0.09cm}\\
CogVideoX (2B)       & 49.8 (-7.8) & 28.2 (-3.2) & 11.6 (-0.6)   \\
CogVideoX (5B)       & 39.2 (-2.7) & 22.0 (-0.7)  & 11.3 (-1.1)  \vspace{0.09cm}\\
Wan2.1 (1.3B)        &  57.4 (-3.8) & 33.8 (-4.7)  & 14.6 (-1.3) \\
Wan2.1 (14B)         & 61.5 (-4.8) & 37.0 (-2.7) & 18.0 (+0.2)  \vspace{0.09cm}\\
MAGI-1               & 60.5 (-10.7) & 35.2 (-5.5) & 16.8 (-3.2)  \vspace{0.09cm}\\
HunyuanVideo          & 52.6 (-4.5) & 33.5 (-5.1) & 15.9 (-4.8)  \vspace{0.09cm}\\
Cosmos (7B)           & 60.5 (-10.7)& 38.8 (-10.3)& 18.6 (-5.0)   \\
Cosmos (14B)          & 59.2 (-13.1) &  39.2 (-8.7) & 22.2 (-2.2)   \vspace{0.09cm}\\
\hline
\end{tabular}
\label{tab:accuracy_comparison}
\end{table}

\parahead{Performance by physical dimension} Two trends recur. \emph{Spatial Reasoning} and \emph{Temporal Dynamics} are consistently among the hardest dimensions, even for Wan2.1-14B, Cosmos-14B, Sora, and \mbox{Veo-2}; these require preserving geometric relations, occlusion, event order, and temporal causality, which remain difficult despite high visual fidelity. \emph{Object Properties \& Affordances} yields the highest scores, which we read as evidence that current models depict recognizable objects and simple affordance cues more readily than they model relational, temporal, or causal structure. Visual plausibility alone is thus not sufficient for physical commonsense. Prompt-level bootstrap confidence intervals for the dimension-level results are reported in App.~\ref{app:robustness}. See App.~\ref{app:qualitative} for qualitative results.

\begin{table*}[!t]
\centering
\small
\setlength{\tabcolsep}{8pt}
\caption{\textbf{Comparison of automatic evaluation methods.} Pearson correlation with human judgments on 120 videos. N/A denotes VLM collapse, where a model answers ``yes'' to nearly all questions. Our pipeline aligns with humans more strongly than prompted or specially fine-tuned VLM evaluators.
}
\begin{tabular}{@{}lcccc@{}}
\toprule
\textbf{Evaluator} & \textbf{MAGI-1} & \textbf{Cosmos-14B} & \textbf{HunyuanVideo} & \textbf{Wan2.1-14B} \\
\midrule
Qwen2.5-VL-72B-Instruct-AWQ~\citep{Qwen2.5-VL} & 0.47 & 0.36 & 0.42 & 0.21 \\
Video-LLaVA~\citep{lin-etal-2024-video} &  N/A  &  N/A  & N/A  &  N/A  \\
InternVL3~\citep{zhu2025internvl3} & N/A  & N/A  & N/A  & N/A  \vspace{0.1cm}\\
VideoScore~\citep{he2024videoscore} & -0.26 & 0.24 & -0.09 & -0.08 \\
VideoPhy-2 AutoEval~\citep{bansal2025videophy} & 0.23 & -0.18 & 0.16 & -0.06 \vspace{0.1cm}\\
Ours & \textbf{0.69} & \textbf{0.60} & \textbf{0.61} & \textbf{0.45} \\

\bottomrule
\end{tabular}
\label{tab:human-corr-vlmqa}
\end{table*}

\parahead{Effect of prompt enrichment} We compare base and enriched prompts to test whether making physical constraints explicit improves generation. Prompt enrichment yields modest, model-dependent mean gains, with paired bootstrap 95\% confidence intervals excluding zero for six of the ten open-source models tested (Table~\ref{tab:bootstrap1}). The effect is not uniform. Sora declines under enrichment, possibly because its pipeline already performs internal prompt optimization, as discussed in VBench-2.0~\cite{zheng2025vbench}. Enrichment therefore provides model-dependent benefits but does not close the underlying physical-reasoning gap.

\parahead{Effect of model scale} Larger models often score higher, but scale alone does not solve physical commonsense, and the trend is not monotonic. Within families, Wan2.1-14B improves over Wan2.1-1.3B, yet CogVideoX-2B matches or exceeds CogVideoX-5B on most dimensions, and smaller systems such as MAGI-1 and Wan2.1-1.3B can outperform larger ones such as VideoCrafter2 and CogVideoX-5B. We take this as a signal that factors beyond parameter count, such as architecture, training data, and supervision, matter for physical reasoning, though our setup does not isolate their individual contributions.

\parahead{Human alignment of the evaluation pipeline}
 To validate the evaluator, we run a user study in which 15 participants watch 120 generated videos and answer the same yes/no questions used by the automatic judge, presented in randomized order. Participants see the video while the judge sees only captions, so the study measures how well caption-grounded reasoning recovers full-information human judgments. We report Pearson correlation with human responses separately for four generation models (Table~\ref{tab:human-corr-vlmqa}). Across all four, our pipeline correlates most strongly with humans (0.45–0.69), outperforming every alternative. The best VLM judge, Qwen2.5-VL-72B-Instruct-AWQ~\citep{Qwen2.5-VL}, trails throughout (0.21–0.47), and two video VLMs collapse entirely, answering ``yes'' to nearly all questions. Notably, VLMs fine-tuned specifically for video evaluation (VideoScore~\cite{he2024videoscore}, VideoPhy-2 AutoEval) do no better. {Krippendorff's $\alpha$ ranges from 0.71 to 0.77 for items with at least three responses and from 0.75 to 0.85 for items with at least four responses.} Our pipeline that separates evidence extraction from reasoning produces a more reliable evaluator of physical commonsense than prompted and fine-tuned VLMs. {As complementary robustness checks, a manual audit of the same 120 human-study videos yields 98.5\% event-level caption recall, while mismatching question sets and videos for Wan2.1-14B reduces the positive-response rate from approximately 34\% to 4\%. For more detailed analyses, see App.~\ref{app:human_eval} and~\ref{app:robustness}.}

\parahead{Error-guided prompt refinement} Finally, we ask whether PhysVidBench can provide actionable feedback, not only aggregate scores. Following iterative self-refinement~\cite{xue2024phyt2v}, we treat failed QA items as constraints for revising prompts without updating model parameters. Each round generates a video, the pipeline identifies violated physical requirements, and the next prompt adds targeted constraints (object attributes, interaction verbs, temporal cues, or explicit negatives). {Unlike the fixed enriched prompts evaluated above, this procedure uses evaluation feedback to iteratively revise the prompt after generation.} On 100 prompts, CogVideoX–2B and CogVideoX–5B improved from $21.6{\to}32.7$ (+11.1 percentage points) and $17.8{\to}29.7$ (+11.9 percentage points), respectively. This shows that our PhysVidBench evaluator can expose interpretable physical failures and guide prompt-level refinement.

\section{Conclusion}
We introduced PhysVidBench, a benchmark and language-grounded evaluation framework for physical commonsense in text-to-video generation, with an emphasis on tool use, object affordances, and causal interactions largely overlooked by prior benchmarks. By decomposing evaluation into captioning and caption-grounded question answering, it separates perception from reasoning and {consistently aligns more strongly with human judgment than the prompted or fine-tuned VLM evaluators across all four evaluated generators}. Across open and proprietary models, none reliably handles everyday physics, prompt enrichment helps only modestly, and scale does not close the gap. Our evaluator can also guide training-free prompt refinement.

\section*{Limitations}
By design, our evaluator reasons over captions rather than raw video. This enables the perception--reasoning separation and traceability central to our approach, but makes evaluation dependent on caption quality: omitted information cannot be recovered by the judge. We mitigate this dependence through multi-perspective captioning (App.~\ref{sec:eval-captioning}) and robustness analyses. Replacing AuroraCap with Qwen2.5-VL-72B-Instruct-AWQ~\citep{Qwen2.5-VL} or Qwen3-VL-30B-A3B-Instruct~\citep{qwen3vl} yields highly consistent scores (App.~\ref{sec:robustness:captioner}), while replacing the Gemini-Flash judge with DeepSeek nearly preserves model rankings (Spearman $\rho{=}0.98$; Sec.~\ref{sec:robustness:judge}). The protocol is also not intended for correctness that depends on continuous visual dynamics, such as wave profiles, flame motion, or turbulent mixing. Finally, PhysVidBench is smaller than some concurrent benchmarks, reflecting our emphasis on human-validated prompts and tool-mediated interactions. PIQA predominantly reflects Western household scenarios, leaving cross-cultural generalization as an open question.

\clearpage
\bibliography{custom}
\clearpage
\appendix

\renewcommand{\thefigure}{A.\arabic{figure}}
\renewcommand{\thetable}{A.\arabic{table}}
\setcounter{section}{0}
\setcounter{figure}{0}
\setcounter{table}{0}

\section*{Appendix}
In this supplementary material, we provide:

\begin{itemize}[left=0cm]
  \item A comparison of PhysVidBench with prior physical commonsense video benchmarks, namely PhyGenBench, \mbox{VBench~2.0}, \mbox{VideoPhy-2}, and Impossible Videos, with representative prompts and diversity statistics (App.~\ref{app:benchmark_comparison}).

  \item A worked example showing that a strong off-the-shelf video metric fails to separate correct from incorrect generations of the same prompt (App.~\ref{app:videophy2_example}).
    
  \item Our benchmark design, including prompt generation and upsampling templates, and the expanded seven-dimension physical commonsense ontology, with per-dimension definitions and example base vs. upsampled prompts (App.~\ref{app:benchmark_design}).
 
  \item Our dense captioning and QA workflow, including dimension-targeted AuroraCap prompts, sample captions, and the full judge prompt (App.~\ref{app:captioning_qa}).

  \item A taxonomy of physically grounded failure modes along three axes, Action Complexity, Scientific Principle, and Object Functionality, explaining \emph{why} certain prompts are hard (App.~\ref{app:failure_modes}).

  \item A full end-to-end example evaluation on Phys-VidBench (App.~\ref{app:e2e_example}).
 
  \item Qualitative examples contrasting successful and failed generations across models and dimensions, with key frames and QA verdicts, and the full specifications of the tested models (App.~\ref{app:qualitative}).
  
  \item Details of our human evaluation study, including interface screenshots and human–LLM agreement (App.~\ref{app:human_eval}).
 
  \item Robustness analysis of the evaluation pipeline, including cross-judge stability, captioner sensitivity, caption recall, a mismatched-video control, and prompt-level confidence intervals for dimension-level results, error-guided prompt refinement, and statistical significance analysis (App.~\ref{app:robustness}).

\end{itemize}

\section{Expanded Comparison with Existing Benchmarks}
\label{app:benchmark_comparison}

\subsection{Representative Prompts from Each Benchmark}
\label{app:priorbench_prompts}
While prior video‐generation benchmarks probe particular aspects of physical reasoning, none offer the breadth and task‐oriented focus of PhysVidBench. 

\begin{itemize}[left=0cm]
  \item \textbf{PhyGenBench}~\citep{meng2025towards} targets isolated physical laws like buoyancy, stress, elasticity through scenarios such as a sponge being squeezed or a ball bouncing.  Its prompts highlight single phenomena in passive settings and omit any agent‐driven tool use or multi‐step manipulations.
  \item \textbf{VBench 2.0}~\citep{zheng2025vbench} introduces a wide variety of scene types and evaluates models on spatial consistency, temporal coherence, and visual realism.  However, it devotes limited attention to object affordances or the mechanics of purposeful interactions, testing where objects ``are” rather than what agents can “do” with them.
  \item \textbf{VideoPhy-2}~\citep{bansal2025videophy} emphasizes motion plausibility and physical violations, e.g. a Segway traversing speed bumps or a gymnast back‐flipping near a wall. Yet, these scenarios rarely involve purposeful tool handling or causal sequences of actions.
  \item \textbf{Impossible Videos}~\citep{bai2025impossible} leverages visually striking, intentionally implausible phenomena such as endless pouring of liquid, and self‐moving books to expose hallucinations in generation models.  While effective at uncovering obvious physics failures, its surreal setups lack grounding in real‐world, goal‐driven tasks.
\end{itemize}

In contrast, \textbf{PhysVidBench} is built around grounded, goal‐oriented prompts derived from the PIQA dataset.  Each scenario requires agents to use or repurpose everyday objects in deliberate sequences that simultaneously engage multiple commonsense dimensions (force application, spatial fit, material properties, containment, and procedural causality).  This affordance‐centric design makes PhysVidBench uniquely suited for assessing whether text‐to‐video models truly ``understand” how tools and materials behave in practical, manipulable tasks.

In the following, we list representative prompts from each benchmark to illustrate these contrasts.

\begin{mainbox}{PhyGenBench}
\textbf{Example Prompt 1:} \textsf{\small A cup of water is slowly poured out in the space station, releasing the liquid into the surrounding area.}\vspace{0.4em}\\
\textit{Tests gravity absence in a synthetic environment; lacks real-world, goal-directed action or interaction with tools.}
\mydashrule
\textbf{Example Prompt 2:} \textsf{\small A timelapse of a water-filled sponge being forcefully squeezed by hand, with the pressure intensifying rapidly over time.}\vspace{0.4em} \\
\textit{Shows material deformation under stress, but the action is passive. It does not involve affordance-based manipulation or multi-step use of the sponge.}
\mydashrule
\textbf{Example Prompt 3:} \textsf{\small A vibrant, elastic rubber ball is thrown forcefully towards the ground, capturing its dynamic interaction with the surface upon impact.}\vspace{0.4em} \\
\textit{Focuses on elasticity and rebound, but lacks physical goal structure, repurposed tool logic, or causal task sequences.}
\end{mainbox}

\begin{mainbox}{VBench 2.0}
\textbf{Example Prompt 1:} \textsf{\small A brown dog is on the left of an apple, then the dog moves to the right of the apple.} \vspace{0.4em}\\
\textit{Tests spatial consistency from different perspectives but lacks physical interaction, causality, or manipulation involving materials or tools.}
\mydashrule
\textbf{Example Prompt 2:} \textsf{\small A jar of peanut butter is opened in the space station, with the viscous liquid slowly dispersing.} \vspace{0.4em} \\
\textit{Depicts fluid behavior in zero gravity but is passive and context-specific. There is no goal-directed manipulation or tool-based interaction.}
\mydashrule
\textbf{Example Prompt 3:} \textsf{\small A person is drinking a glass of water, then they suddenly start cleaning the windows.} \vspace{0.4em} \\
\textit{Shows a behavioral transition but omits the physical process involved in cleaning. The objects used and the steps required are unspecified and ungrounded.}
\end{mainbox}

\begin{table*}[!h]
\caption{\textbf{Lexical diversity comparison across benchmarks.} PhysVidBench (base and upsampled variants) exhibits high diversity in verb lemmas, nouns, and action--object pairs relative to prior physics-oriented video benchmarks. {These statistics provide a lexical proxy for the variety of physical interactions represented in the prompts.}}
\centering
\small
\setlength{\tabcolsep}{5pt}
\renewcommand{\arraystretch}{1.6}
\resizebox{\textwidth}{!}{
\begin{tabular}{lccccccc}
\toprule
{\textbf{Benchmark}} &
{\makecell{\textbf{\#}\\\textbf{Prompt}}} &
{\makecell{\textbf{Unique}\\\textbf{Verbs}}} &
{\makecell{\textbf{Verbs}\\\textbf{per Prompt}}} &
{\makecell{\textbf{Unique}\\\textbf{Nouns}}} &
{\makecell{\textbf{Nouns}\\\textbf{per Prompt}}} &
{\makecell{\textbf{Action--Object}\\\textbf{Pairs per Prompt}}} &
{\makecell{\textbf{Lexical}\\\textbf{Density}}} \\
\midrule
{\makecell[l]{PhyGenBench}} & {160} & {{78}} & {{0.488}} & {239} & {1.494} & {0.79} & {0.55} \\
{\makecell[l]{PhyGenBench\\(explicit prompt)}} & {160} & {232} & {1.45} & {408} & {2.55} & {2.91} & {0.57} \\
{\makecell[l]{VideoPhy-2\\(train set)}} & {3343} & {773} & {0.231} & {2111} & {0.631} & {1.43} & {0.59} \\
{\makecell[l]{VBench 2.0}} & {1013} & {593} & {0.585} & {1282} & {1.266} & {1.75} & {0.55} \\
{\makecell[l]{VBench 2.0\\(augmented)}} & {1013} & {1195} & {1.18} & {2979} & {2.941} & {6.37} & {0.61} \\
{\makecell[l]{PhysVidBench\\(base)}} & {383} & {262} & {0.684} & {708} & {1.849} & {2.03} & {0.60} \\
{\makecell[l]{PhysVidBench\\(upsampled)}} & {383} & {{625}} & {{1.632}} & {1233} & {3.219} & {6.68} & {0.64} \\
\bottomrule
\end{tabular}}
\label{tab:diversity}
\end{table*}

\begin{mainbox}{VideoPhy-2}

\textbf{Example Prompt 1:} \textsf{\small A Segway bumps over a series of small speed bumps, the rider maintaining balance with slight adjustments.} \vspace{0.4em} \\
\textit{Evaluates motion plausibility and equilibrium but lacks goal-directed object use, sequential interaction steps, or any reasoning about tool affordances.}
\mydashrule
\textbf{Example Prompt 2:} \textsf{\small A backflip is performed close to a wall. The person carefully ensures their momentum keeps them clear from hitting the wall.} \vspace{0.4em} \\
\textit{Involves physical trajectory reasoning but does not require interaction with any objects or manipulation of the environment for task completion.}
\mydashrule
\textbf{Example Prompt 3:} \textsf{\small A backhoe digs a large hole in the ground, throwing up a mound of dirt.} \vspace{0.4em} \\
\textit{Depicts mechanical tool usage, yet the prompt remains high-level and underspecified. It does not describe the procedural steps or physical constraints involved in the interaction.}
\end{mainbox}

\begin{mainbox}{Impossible Videos}

\textbf{Example Prompt 1:} \textsf{\small A person continuously pours water from a glass pot, yet remarkably the water level inside remains constant. This photo-realistic illusion defies natural physics as the endless stream flows without depleting the pot's contents. The scene takes place against a plain background, creating a focused view of this seemingly impossible phenomenon.} \vspace{0.4em} \\
\textit{Intentionally violates conservation of mass for visual paradox, but lacks grounded, goal-driven interaction or affordance-sensitive manipulation.}
\mydashrule
\textbf{Example Prompt 2:} \textsf{\small A metallic iron ball bounces off a polished marble floor in a physically unusual manner. Defying normal physics, each consecutive bounce reaches a greater height than the last, creating an eerily unnatural progression of increasing rebounds. The glossy surface of the floor reflects the ball's motion in this photo-realistic scene, set against what appears to be an indoor space.} \vspace{0.4em} \\
\textit{Targets energy conservation violation in isolation. However, it contains no agent, object use, or causal physical interaction sequences grounded in real-world affordances.}
\mydashrule
\textbf{Example Prompt 3:} \textsf{\small A book mysteriously opens by itself on a plain wooden desk surface, its pages flipping autonomously without any visible human intervention or external force. The photo-realistic footage captures this unexplainable movement in a well-lit indoor setting, defying natural physics as the book cover and pages lift and turn on their own.} \vspace{0.4em} \\
\textit{Demonstrates spontaneous motion but offers no agent–object interaction, mechanical force reasoning, or procedural physicality that would be necessary in task-driven video generation.}
\end{mainbox}

\begin{figure*}[!h]
\centering
\includegraphics[width=\textwidth]{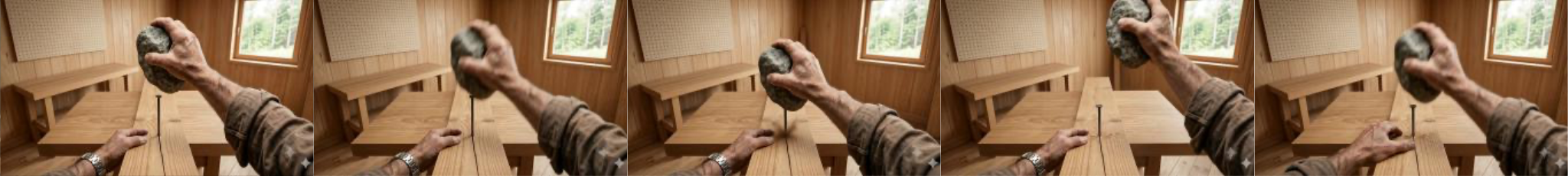}\vspace{-0.3em}\\
{\small (a) Physically correct (SA:3, PC:3)}\\
\vspace{0.5em}
\includegraphics[width=\textwidth]{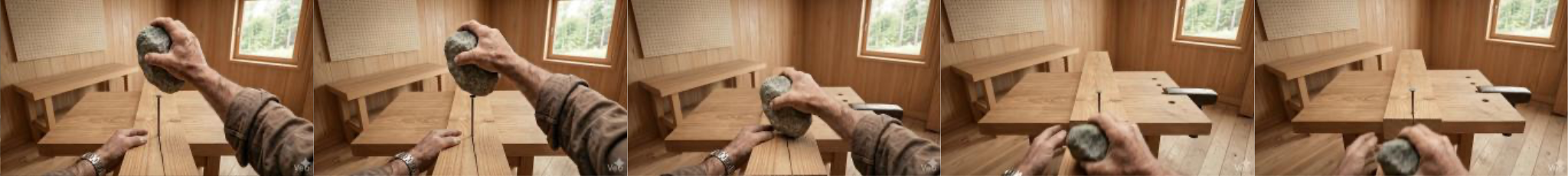}\vspace{-0.3em}\\
{\small (b) Physically wrong (SA:3, PC:3)}\\
\vspace{0.5em} \includegraphics[width=\textwidth]{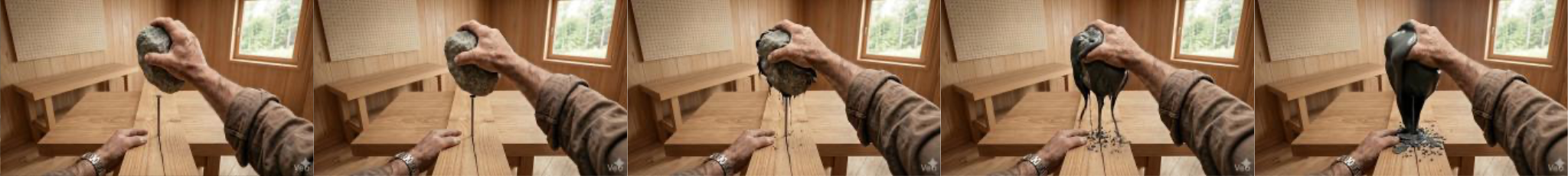}\vspace{-0.3em}\\
{\small (c) Physically wrong (SA:3, PC:3)}
\caption{\textbf{Limitations of direct VLM-based evaluation for physical commonsense assessment.} 
    While Semantic Adherence (SA) measures whether the generated video matches the input text prompt semantically, Physical Commonsense (PC) evaluates whether the generated content follows intuitive real-world physical laws. 
    Existing direct scoring approaches often fail to distinguish semantically correct but physically implausible generations, assigning similarly high scores even when severe physical violations occur.}
    \label{fig:three_figures}
\end{figure*}

\begin{mainbox}{PhysVidBench (Ours)}

\textbf{Example Prompt 1:} \textsf{\small A slice of bread is pressed onto small glass fragments scattered on a surface, picking them up.} \vspace{0.4em} \\
\textit{Involves soft material usage for delicate cleanup. Engages reasoning about texture, surface adhesion, and containment, grounded in everyday physical affordances.}
\mydashrule
\textbf{Example Prompt 2:} \textsf{\small Duct tape is wrapped around a stuck light bulb in a socket, and a hand uses the tape to twist the bulb counter-clockwise, loosening it.} \vspace{0.4em} \\
\textit{Demonstrates tool repurposing and torque application. Highlights grip mechanics, flexibility, and rotational force in a functional task context.}
\mydashrule
\textbf{Example Prompt 3:} \textsf{\small A drink can is placed upright into the opening of a shoe resting on the floor of a car's passenger side.} \vspace{0.4em} \\
\textit{Tests reasoning about shape compatibility, friction, and containment. Showcases improvised affordance in a constrained physical setup.}
\end{mainbox}

{Complementing the lexical diversity statistics in Table~\ref{tab:diversity}, PhysVidBench contains 62 of PhyGenBench's 78 unique verb lemmas (79.5\%) and covers 75.5\% of its verb occurrences. These lexical statistics provide an additional proxy for the variety of physical interactions represented in the prompts.}

\parahead{Cross-benchmark dimension coverage}
{Within PhysVidBench, the seven physical dimensions are assigned based on the manually reviewed QA items, with each prompt contributing to every dimension represented by at least one of its questions. For cross-benchmark comparison, we instead apply a common GPT-4o-mini-based multi-label classifier uniformly to the prompts of all benchmarks.}

{Under this common classifier, 97.7\% of PhysVidBench prompts cover Object Properties \& Affordances, compared with 27.5--59.1\% in the comparison benchmarks. Coverage is 89.3\% for Force and Motion, 71.0\% for Material Interaction \& Transformation, and 39.7\% for Action \& Procedural Understanding. Prompts jointly involving Object Properties \& Affordances, Force and Motion, and Material Interaction \& Transformation account for 14.4\% of PhysVidBench, compared with at most 2.0\% elsewhere.}

\subsection{Dataset Diversity and Scale}
\label{app:dataset-diversity}

PhysVidBench achieves broad conceptual coverage despite its compact size. Each prompt simultaneously probes several aspects of everyday physical reasoning such as affordances, causality, material response, and procedural manipulation, yielding higher supervision density than benchmarks built around isolated physical laws. {As Table~\ref{tab:diversity} shows, the base and upsampled prompts exhibit high lexical diversity in verbs, nouns, and action--object pairs; we use this only as a proxy for the variety of real-world interactions, alongside the separate classifier-based coverage analysis above.} Although the benchmark contains 383 base prompts, its effective evaluation scale is large: expanded across models, captions, and physics-grounded questions, it produces more than 42{,}000 QA instances per split, offering fine-grained diagnostic resolution in practice. This scale is consistent with established physics-oriented diagnostic datasets (PhyGenBench: 160 prompts; VideoPhy: 688 prompts), which likewise prioritize conceptual precision and interpretability over raw quantity. PhysVidBench follows the same design philosophy, remaining computationally feasible for the community while still rich enough to meaningfully evaluate physical commonsense in T2V models.

\section{Limitations of Direct VLM Scoring for Physical Commonsense}
\label{app:videophy2_example}
Direct VLM-based scoring is appealing because it is simple: a video and a prompt are passed to a vision-language model, which returns scalar scores for semantic adherence and physical correctness. Physical commonsense, however, is not merely a visual recognition problem. It requires verifying whether the video actually supports the causal and procedural consequences the prompt implies. Direct scoring is therefore vulnerable to shortcut judgments grounded in surface-level visual plausibility, object presence, or general prompt alignment.

Figure~\ref{fig:three_figures} illustrates this failure mode in a controlled setting. We compare generations that share the same semantic context: a person uses a stone to hit a nail on a wooden surface. In the physically plausible case, the stone contacts the nail and the nail is gradually driven into the wood. In the physically implausible cases, the semantic scene is still recognizable, but the physical interaction breaks down: the stone either passes through the nail in an impossible manner or melts before producing the expected mechanical effect. Despite these severe physical violations, the direct VLM scorer (VideoPhy 2 Autoeval) assigns the same high scores to all cases, including both the plausible and impossible generations (SA:3, PC:3). This suggests that the scorer recognizes the objects, action context, and overall visual scene, but fails to distinguish whether the required physical interaction actually occurs.

This behavior is especially problematic when direct VLM scores are used as training targets or reward signals. If a model is trained only to match scalar human or VLM scores, it may learn shortcuts that correlate with physical correctness without truly evaluating physical causality. For example, stronger video generation models often produce sharper, more coherent, and more realistic-looking videos; these same models are also less likely to contain obvious physical artifacts. As a result, visual quality can become spuriously correlated with physical quality. A direct evaluator trained on such scores may therefore learn to reward high visual fidelity, temporal smoothness, or object recognizability, rather than the actual satisfaction of the physical constraints implied by the prompt. In this case, the evaluator may appear correlated with human physical-quality judgments at the dataset level, while still failing on counterfactual examples where semantic alignment and visual plausibility are held fixed but physical correctness differs.

Our method is designed to reduce this shortcut risk. Instead of asking a VLM to directly assign a single physical correctness score, PhysVidBench decomposes evaluation into captioning and caption-grounded question answering. The video is first converted into explicit textual evidence, and then an LLM answers targeted yes/no questions about the expected physical consequences. In the example in Figure~\ref{fig:three_figures}, the evaluator does not simply ask whether the scene shows a person, a stone, and a nail. It checks whether the stone remains solid, whether it contacts the nail, whether the nail moves into the wood after impact, and whether the observed sequence supports the intended causal effect. If the caption evidence does not entail these physical events, the corresponding questions are answered negatively.

This decomposition makes evaluation more interpretable and less reliant on a single opaque judgment. Direct scoring can collapse a physically correct video and a physically impossible one onto the same score; caption-grounded QA instead pinpoints which physical requirement is missing. PhysVidBench thus provides a more reliable signal for physical commonsense, especially when visually realistic generations still violate basic causal, material, or interaction-level constraints.

\section{Benchmark Design and Prompt Construction}
\label{app:benchmark_design}

\subsection{Prompt Generation and Upsampling}
\label{sec:prompt-generation}
Our benchmark construction pipeline begins with the PIQA dataset, utilizing its extensive repository of physically grounded scenarios. {PIQA contains roughly 21K total examples. Because the test split lacks public ground-truth labels, our processing pipeline uses the labeled train and validation splits, yielding more than 17K correct goal-solution pairs before filtering, from which 383 final prompts are retained.} Our pipeline then proceeds through three stages: (1) \textit{Initial filtering}; (2) \textit{Video prompt generation}; and (3) \textit{Prompt upsampling}.

\parahead{Stage 1: PIQA Dataset and Initial Filtering} We use Gemini 2.5 Pro model to select only those PIQA scenarios that involve secondary or repurposed tool use, those that inherently require physical-commonsense reasoning. This filtering yields a compact set of complex, interaction-centric tasks ideal for text-to-video evaluation. Below is an example prompt illustrating the instructions provided to Gemini 2.5 Pro in this stage.

\begin{mainbox}{Video Generation Pipeline -- Stage 1: PIQA Dataset and Initial Filtering}
\textsf{\small I will provide a series of examples from the PIQA dataset where common household objects or tools are used either in their primary/intended way or in repurposed/alternative ways (i.e., “life hacks”). Your task is to determine whether a given PIQA goal–solution pair reflects the primary usage of an object or not.} \\

\textsf{\small \textbf{Definition}: If the object is being used in its intended, standard, or conventional way, respond with: No \\
If the object is being used in an alternative or repurposed way (i.e., not its typical function), respond with: Yes }\\

\textsf{\small You are given a list of goal–action pairs (e.g., “Water Bottle → Egg Separator”, “Credit Card → Ice Scraper”) which illustrate how PIQA questions often involve creative or secondary uses of common items. These examples represent secondary usage. This list can guide your understanding of what constitutes repurposing. You will now be given a PIQA goal and two solutions. Based on your understanding of object affordances and typical usage. Determine only whether the goal/solution pair represents a secondary use of an object. Write only the answer, Yes or No. Do not explain or describe the reasoning. }\\

\textsf{\small Here are examples from the PIQA dataset where common objects or tools are used for a purpose other than their primary, intended function (i.e., alternative uses or "life hacks"):} \\

\textsf{\small \textbf{Water Bottle → Egg Separator:}}\\
\textsf{\small \textbf{Goal:} To separate egg whites from the yolk using a water bottle...}\\
\textsf{\small \textbf{Action:} Squeeze the water bottle and press it against the yolk. Release, which creates suction and lifts the yolk.}\\

\textsf{\small \textbf{Hair Net → Vacuum Filter:}}\\
\textsf{\small \textbf{Goal:} How do I find something I lost on the carpet?}\\
\textsf{\small \textbf{Action:} Put a hair net on the end of your vacuum and turn it on. (To catch small items before they go into the vacuum bag).}\\

\textsf{\small These examples demonstrate the kind of creative repurposing and understanding of object affordances beyond their primary function that the PIQA dataset aims to capture.}

\textsf{\small Write only the answer for each goal with comma separator: Yes or No. Do not explain or describe the reasoning.}\\

\textsf{\small Give the answer for this Goal Solution pair:}\\
\textsf{\small \textbf{Goal:} How can I paint stars on a black background? }\\
\textsf{\small \textbf{Solution:} Use an old toothbrush covered in white paint and flick at the background. }\\
\end{mainbox}

\parahead{Stage 2: Video Prompt Generation} We use the Gemini 2.5 Pro model to transform each filtered goal–solution pair into a concise, visually explicit video generation prompt. This process produces 383 distinct \emph{base} prompts, each outlining a self-contained scenario with clear, physically grounded actions. Below is an example prompt illustrating the instructions provided to Gemini 2.5 Pro in this stage.

\begin{mainbox}{Video Generation Pipeline -- Stage 2: Video Prompt Generation}
\textsf{\small You are given question–answer pairs from the PIQA dataset. Each pair consists of a physical goal (what someone wants to achieve) and a solution (how to achieve it). Your task is to write a \textbf{realistic and physically plausible video generation prompt} that visually demonstrates the provided solution.}\\

\textsf{\small \textbf{Important Instructions:}}
\begin{itemize}[left=0pt]
\item \textsf{\small Only generate a short, visualizable \textbf{video prompt} that clearly shows the solution being carried out.}
\item \textsf{\small The prompt should describe \textbf{a realistic short video clip} (under 10 seconds).}
\item \textsf{\small You do \textbf{not} have to begin each prompt with “A person”. Be natural — use whatever fits best (e.g., “A dog…”, “Hands…”, “Someone…”, “Two people…”, “The object…”).}
\item \textsf{\small If the solution is not visualizable or does not reflect meaningful physical interaction or common-sense action, \textbf{skip it} (do not generate a prompt).}
\item \textsf{\small Be \textbf{specific} and \textbf{physically grounded} — actions should be directly observable (e.g., cutting, pouring, lifting, placing).}
\item \textsf{\small Do not repeat the goal or the original solution text.}
\end{itemize}
$\;$\\
\textsf{\small \textbf{Example Conversions:}}\\

\textsf{\small \textbf{Goal:} You want to dry your wet hands.}\\
\textsf{\small \textbf{Solution:} Use a towel  }\\
\textsf{\small \textbf{Video Prompt for Solution:} Wet hands are rubbed against a cotton towel.}\\

\textsf{\small \textbf{Goal:} You want to keep a door open.}\\  
\textsf{\small \textbf{Solution:} Use a doorstop}  \\
\textsf{\small \textbf{Video Prompt for Solution:} A door is held open by sliding a rubber doorstop underneath.}\\

\textsf{\small \textbf{Goal:} You want to get rid of wrinkles in a shirt. } \\
\textsf{\small \textbf{Solution:} Use an iron  }\\
\textsf{\small \textbf{Video Prompt for Solution:} An iron glides over a wrinkled shirt on an ironing board, flattening it.}\\

\textsf{\small Now, for each of the following goal–solution pairs from the PIQA dataset, return a single line in the format below \textbf{only if it qualifies}:}\\

\textsf{\small PIQA Input:}\\
\textsf{\small \textbf{Goal:} How can I paint stars on a black background? }\\
\textsf{\small \textbf{Solution:} Use an old toothbrush covered in white paint and flick at the background.} \\
\end{mainbox}

\parahead{Stage 3: Prompt Upsampling} We use the Gemini 2.5 Pro model to enrich each \emph{base} prompt with fine-grained physical details such as explicit force application, material behaviors, and spatial constraints, while preserving the original scene scope. This controlled upsampling yields 383 \emph{upsampled} prompts that surface deeper commonsense cues without introducing new objects or outcomes. Below is an example prompt illustrating the instructions provided to Gemini 2.5 Pro in this stage.

\begin{mainbox}{Video Generation Pipeline -- Stage 3: Prompt Upsampling}
\textsf{\small You are a reasoning-augmented AI model tasked with upsampling video generation prompts using knowledge of physical commonsense and object affordances, while strictly preserving the original scene boundaries. Given a simple prompt, you must rewrite it into a richer, more physically grounded version by:}
\begin{itemize}[left=0pt]
\item \textsf{\small Elaborating on object properties (e.g., material, size, rigidity, weight) that affect how actions are performed}
\item \textsf{\small Clarifying affordances (e.g., what can be held, moved, poured, tied) based on those properties}
\item \textsf{\small Expanding the action description in a physically realistic, causally plausible sequence}
\item \textsf{\small Including any relevant environmental or contextual conditions (e.g., gravity, surface type)}
\item \textsf{\small Do not introduce any new objects or tools not mentioned in the original prompt}
\item \textsf{\small Do not describe outcomes or events that happen after the main action (no result, consequence, or future speculation)}
\item \textsf{\small Your goal is to make the prompt maximally useful for video generation models that rely on physical reasoning, without hallucinating or extending beyond the original scope.}
\end{itemize}$\;$\\
\textsf{\small \textbf{Example Input:}}\\
\textsf{\small "A person places coffee filters inside the cup holders of a car."}\\

\textsf{\small \textbf{Upsampled Output:}}\\
\textsf{\small "A person carefully pushes white paper coffee filters, each lightweight and slightly crinkled, into the round plastic cup holders on the car’s center console. The filters conform to the shape of the holders due to their flexible, foldable structure. The car interior is clean and the cup holders are empty, allowing the filters to settle flush against the bottom surface."}\\

\textsf{\small Only return the upsampled prompt as output. Do not explain your reasoning or include anything else in your answer.}\\

\textsf{\small \textbf{Prompt:}
A person dips an old toothbrush in white paint and flicks it onto a black background, creating stars.}
\end{mainbox}

{Manual validation was performed by four authors using a shared review form and written criteria covering visualizability, tool use, affordance relevance, and physical commonsense. Each candidate was independently reviewed by two authors, and disagreements were resolved through discussion before inclusion.}

\noindent Altogether, PhysVidBench comprises 766 unique prompts (383 base + 383 upsampled), enabling a comprehensive evaluation of physical commonsense reasoning in text-to-video models. Representative examples illustrating the outcome of each stage of our pipeline are shown below.

\begin{mainbox}{Example 1}
\textsf{\small \textbf{Stage 1: PIQA Dataset and Initial Filtering}}\\
\textsf{\small {Goal:} How can I paint stars on a black background?} \\
\textsf{\small {Solution:} Use an old toothbrush covered in white paint and flick at the background.}

\hdashrule[0.5ex]{\linewidth}{0.5pt}{2pt 2pt}  %

\textsf{\small \textbf{Stage 2: Video Prompt Generation}}\\
\textsf{\small A person dips an old toothbrush in white paint and flicks it onto a black background, creating stars.}
\hdashrule[0.5ex]{\linewidth}{0.5pt}{2pt 2pt}  %

\textsf{\small \textbf{Stage 3: Prompt Upsampling}}\\
\textsf{\small A person firmly grips the handle of an old, worn toothbrush, its plastic slightly yellowed with age. They dip the frayed bristles into a can of thick, white paint. The paint clings to the bristles due to its viscosity. With a controlled wrist motion, they flick the paint-laden toothbrush towards a large, matte black canvas backdrop. Tiny droplets of white paint detach from the bristles and spray outwards, propelled by the force of the flick. The droplets travel through the air, influenced by gravity, before adhering to the vertical surface of the canvas.}
\end{mainbox}

\begin{mainbox}{Example 2}
\textsf{\small \textbf{Stage 1: PIQA Dataset and Initial Filtering}}\\
\textsf{\small {Goal:} When I am driving an older car with no cupholders where can I place my beverage?} \\
\textsf{\small {Solution:} Use an old shoe, place the drink container in the place where your foot enters the shoe and it will help stabilize it.}

\hdashrule[0.5ex]{\linewidth}{0.5pt}{2pt 2pt}  %

\textsf{\small \textbf{Stage 2: Video Prompt Generation}}\\
\textsf{\small A drink can is placed upright into the opening of a shoe resting on the floor of a car's passenger side.}

\hdashrule[0.5ex]{\linewidth}{0.5pt}{2pt 2pt}  %

\textsf{\small \textbf{Stage 3: Prompt Upsampling}}\\
\textsf{\small A cylindrical aluminum drink can, rigid due to its metal construction and filled with liquid giving it noticeable weight, is carefully maneuvered vertically downwards. It is inserted, base-first, into the flexible fabric opening of a casual shoe which is resting flat on its firm sole upon the textured floor surface of the car's passenger side. The can maintains its upright orientation under gravity as its smooth, flat circular base makes stable contact with the interior insole surface deep within the shoe's cavity.}
\end{mainbox}

\begin{mainbox}{Example 3}
\textsf{\small \textbf{Stage 1: PIQA Dataset and Initial Filtering}}\\
\textsf{\small {Goal:} How do I find something I lost on the carpet?} \\
\textsf{\small {Solution:} Put a hair net on the end of your vacuum and turn it on.}

\hdashrule[0.5ex]{\linewidth}{0.5pt}{2pt 2pt}  %

\textsf{\small \textbf{Stage 2: Video Prompt Generation}}\\
\textsf{\small A person attaches a hair net to the end of a vacuum cleaner and uses it to search for a small object on a carpet.}

\hdashrule[0.5ex]{\linewidth}{0.5pt}{2pt 2pt}  %

\textsf{\small \textbf{Stage 3: Prompt Upsampling}}\\
\textsf{\small A person stretches the elastic rim of a fine, white nylon hair net, expanding its opening, and carefully fits it over the circular nozzle of a dark gray plastic vacuum cleaner hose. The net conforms to the hose's shape, held in place by its own tension. The vacuum cleaner is then used to gently hover over a low-pile carpet, its fibers slightly compressed by the vacuum's suction. The person guides the nozzle in slow, overlapping passes, the hair net acting as a barrier to prevent small objects from being sucked into the machine.}
\end{mainbox}

\subsection{Physical Commonsense Dimensions and Sample Prompts}
\label{sec:dim-prompts}
We present each of the seven physical commonsense dimensions defined in our benchmark, with representative base and upsampled prompts.The prompts shown under each category were chosen because they most strongly emphasize that form of reasoning, but PhysVidBench prompts are not confined to a single dimension. In practice, most scenarios engage several at once. A prompt may combine force application with spatial configuration, or material properties with temporal sequencing. This multi-dimensional grounding mirrors the complexity of real physical tasks and distinguishes PhysVidBench from benchmarks that test physical concepts in isolation.\vspace{-0.3cm}

\begin{mainbox}{Fundamental Physics: \textit{The foundational laws governing energy, causality, equilibrium, and changes of physical state}}
\textsf{\small \textbf{Base Prompt:} A flashlight lying on a dark floor illuminates tiny shards of broken glass by casting long shadows from them.}\\
\textsf{\small \textbf{Upsampled Prompt:} A cylindrical metal flashlight rests horizontally on a dark, flat floor, its switch in the 'on' position. A focused beam of bright light emits from its lens, traveling parallel and close to the floor surface. This low-angle light strikes multiple tiny, sharp-edged, transparent shards of broken glass scattered across the floor. Because the light source is nearly level with the floor, even the small vertical profile of each glass shard is sufficient to block the light path, causing disproportionately long, thin shadows to stretch out behind each shard, starkly visible against the surrounding darkness of the floor.
}
\mydashrule
\textsf{\small \textbf{Base Prompt:} A hand pours a small amount of white flour into a larger bowl of water and begins stirring with a spoon.
}\\
\textsf{\small \textbf{Upsampled Prompt:} A human hand directs a small stream of fine, white, lightweight flour powder downwards due to gravity, letting it fall onto the surface of still water held within a larger, rigid bowl resting on a flat surface. The flour particles make contact with the liquid. The hand then grips the rigid handle of a spoon, submerges the spoon's end into the water where the flour landed, and initiates a stirring motion, moving the spoon through the water and the suspended flour particles.
}
\mydashrule
\textsf{\small \textbf{Base Prompt:} A plastic bag containing meat sits on an upside-down aluminum pot, with another pot filled with water placed directly on top of the bagged meat.
}\\
\textsf{\small \textbf{Upsampled Prompt:} A flexible, translucent plastic bag, visibly holding a weighty piece of raw meat which causes the thin material to bulge slightly, is positioned on the flat, circular, upturned bottom of a lightweight yet rigid aluminum cooking pot resting stably upside-down on its rim. Placed directly centered on top of this bagged meat rests another rigid cooking pot, its flat bottom making full contact; this second pot contains a significant volume of water, adding substantial downward weight due to gravity. The pliable plastic bag is compressed between the base of the top, water-filled pot and the upturned base of the bottom pot, conforming slightly to the shapes above and below it while transferring the load.
}
\end{mainbox}

\begin{mainbox}{{Object Properties \& Affordances:} \textit{The study of inherent object attributes, including material composition, rigidity, softness, and functional affordances like containment or support}}
\textsf{\small \textbf{Base Prompt:} A stainless steel bucket is placed upside down over a small ant crawling on the ground.
}\\
\textsf{\small \textbf{Upsampled Prompt:} A rigid, opaque stainless steel bucket, positioned upside down with its open circular rim facing downwards, is carefully lowered vertically towards the ground. The bucket descends directly over a small, dark ant actively crawling across the relatively flat earth surface. The bucket's descent continues until its metal rim makes firm contact with the ground, encompassing the area occupied by the crawling ant.
}
\mydashrule
\textsf{\small \textbf{Base Prompt:} A hand rubs a piece of bread back and forth over pencil writing on paper.}\\
\textsf{\small \textbf{Upsampled Prompt:} A human hand grips a piece of soft, somewhat pliable white bread, likely without crusts, between its fingers. Positioned over a sheet of standard writing paper lying flat on a hard surface, the hand methodically rubs the bread's yielding, slightly textured underside back and forth. This motion occurs directly over an area marked with fine, gray pencil writing, causing the bread's surface to make repeated contact with the graphite lines on the paper under consistent, light pressure.
}
\mydashrule
\textsf{\small \textbf{Base Prompt:} Hands fold one short end of a credit card up and the other short end down, then place a smartphone onto the upward fold.
}\\
\textsf{\small \textbf{Upsampled Prompt:} A pair of hands holds a standard-sized, thin, rectangular plastic credit card. Applying pressure with fingers, the hands bend one short edge of the rigid-yet-flexible card sharply upwards along its width, creating a defined crease. Then, the hands bend the opposite short edge sharply downwards along its width, forming a second parallel crease, configuring the card into a small S-shaped structure. Next, the hands carefully lift a smooth, relatively heavy, rigid rectangular smartphone. Aligning the bottom edge of the phone with the small, upward-facing plastic ledge created by the first fold on the credit card, the hands gently lower the weight of the smartphone onto this narrow support.
}
\end{mainbox}

\begin{mainbox}{Spatial Reasoning: \textit{The ability to interpret and infer spatial relations, including position, geometry, occlusion, orientation, and fit between objects in a scene}}
\textsf{\small \textbf{Base Prompt:} The tip of a spoon presses a square outline into a slice of bread, then the back of the spoon pushes down the center.
}\\
\textsf{\small \textbf{Upsampled Prompt:} The hard, pointed metal tip of a rigid spoon presses sequentially into the soft, porous surface of a slice of bread, tracing a square outline by creating shallow indentations. The bread yields slightly under the focused pressure of the tip. Subsequently, the smooth, curved underside (back) of the spoon's bowl is centered within the traced square. Applying downward force, the broader surface of the spoon back pushes into the bread, compressing the central area significantly more deeply than the initial outline, creating a distinct depression within the square boundary on the yielding bread slice.
}
\mydashrule
\textsf{\small \textbf{Base Prompt:} Crushed egg shells and water are shaken vigorously inside a clear bottle.
}\\
\textsf{\small \textbf{Upsampled Prompt:} Inside a clear, rigid bottle containing clear liquid water, numerous small, lightweight, brittle fragments of crushed egg shells are forcefully agitated. The bottle undergoes vigorous, rapid shaking motions, causing the water to slosh violently and carry the sharp-edged shell fragments in turbulent, swirling patterns within the confined space. The mixture of water and shell fragments repeatedly impacts the inner surfaces of the bottle due to the applied force.
}
\mydashrule
\textsf{\small \textbf{Base Prompt:} A hand pushes a plastic drinking straw upwards through the bottom of a strawberry, popping the green stem out the top.
}\\
\textsf{\small \textbf{Upsampled Prompt:} A human hand firmly grips the lower end of a thin, cylindrical, rigid plastic drinking straw. The straw's end is pressed against the pointed bottom tip of a ripe, red strawberry, which is soft enough to be pierced but firm enough to maintain its shape. The hand applies a continuous, steady upward force along the axis of the straw. Due to its rigidity and the applied pressure, the straw penetrates the yielding flesh of the strawberry, moving vertically upwards through its center. As the leading end of the straw travels inside the fruit towards the wider top where the green, leafy stem is attached, it pushes against the stem's base from below. This focused upward force exerted by the straw dislodges the relatively light stem structure, causing it to pop out from the top surface of the strawberry.
}
\end{mainbox}

\begin{mainbox}{Temporal Dynamics: \textit{The domain addressing timing, sequencing, and the causal structure of events over time, such as ordering, delays, and waiting}}
\textsf{\small \textbf{Base Prompt:} A plastic bucket is placed upside down on someone's head like a hat.
}\\
\textsf{\small \textbf{Upsampled Prompt:} A lightweight, rigid plastic bucket, oriented upside down so its solid bottom faces upwards, is carefully lowered onto a person's head. The open circular rim of the bucket makes contact and rests stably upon the crown of the head, balanced there by gravity. The bucket's inherent rigidity allows it to maintain its shape, covering the top of the head like a large, hollow hat.
}
\mydashrule
\textsf{\small \textbf{Base Prompt:} A child places their hand on a large circle sticker affixed to the side of a car in a parking lot.
}\\
\textsf{\small \textbf{Upsampled Prompt:} A small child extends their arm and carefully places their soft, open hand flat onto the smooth surface of a large, circular sticker. The thin, flexible sticker is firmly adhered by its backing to the hard, vertical metal side panel of a stationary car. The car is situated on the level asphalt ground of an outdoor parking lot. The child's palm and fingers press lightly against the sticker's surface, conforming slightly to the car's panel curvature beneath it.
}
\mydashrule
\textsf{\small \textbf{Base Prompt:} A hand squeezes toothpaste onto a cloth and rubs it in circles over a hazy car headlight, leaving a thin white coating.
}\\
\textsf{\small \textbf{Upsampled Prompt:} A human hand squeezes a dollop of viscous, opaque white toothpaste directly onto a section of a soft, flexible fabric cloth. The paste adheres to the cloth's surface. The hand then presses this part of the cloth firmly against the hard, smooth, but visibly hazy surface of a car's plastic headlight cover. Maintaining steady pressure, the hand rubs the cloth in continuous circular motions over the hazy area. The pliable cloth conforms to the headlight's curvature, spreading the toothpaste evenly as it's rubbed, leaving behind a thin, uniform white coating on the headlight surface.
}
\end{mainbox}

\begin{mainbox}{Action \& Procedural Understanding: \textit{The comprehension of structured, goal-driven sequences of actions involving procedural steps toward task completion}}
\textsf{\small \textbf{Base Prompt:} Hands press a piece of clay over the joined edges of pieces on top of a bowl.
}\\
\textsf{\small \textbf{Upsampled Prompt:} Human hands hold a separate lump of soft, malleable, slightly damp clay. The hands carefully position this clay lump directly over the visible seam where the edges of several harder, pre-joined clay pieces meet. These pieces rest securely on the upward-facing surface of a rigid ceramic bowl. Fingers then apply firm, steady downward pressure onto the soft clay lump, causing it to flatten and spread smoothly across the underlying seam, adhering to the surfaces of the joined pieces due to its pliable and slightly sticky nature.
}
\mydashrule
\textsf{\small \textbf{Base Prompt:} A cloth damp with olive oil is wiped over a metal key. The key is then inserted smoothly into a keyhole.
}\\
\textsf{\small \textbf{Upsampled Prompt:} A soft, absorbent cloth, damp and glistening slightly with viscous olive oil, is wiped firmly across the surfaces of a rigid, metallic key. The pliable cloth presses into the key's teeth and grooves, coating the hard metal with a thin, lubricating film of oil. Immediately after, the oil-coated metal key is precisely aligned with the opening of a metal keyhole and inserted with a steady, linear motion, sliding effortlessly into the lock mechanism due to the oil reducing friction between the contacting metal surfaces.
}
\mydashrule
\textsf{\small \textbf{Base Prompt:} Hands carefully pour a green liquid over the back of a spoon into a shot glass containing a brown liquid, forming a distinct layer. Then, a cream liquid is poured similarly on top.
}\\
\textsf{\small \textbf{Upsampled Prompt:} Hands carefully position a rigid metal spoon, its convex back facing upwards, just above the surface of a dark brown, relatively dense liquid resting at the bottom of a small, clear glass shot glass. With precise control, a vibrant green liquid, less dense than the brown liquid, is steadily poured onto the spoon's curved back. The spoon's smooth surface guides the green liquid's flow, allowing it to gently cascade onto the brown liquid's surface, minimizing turbulence and forming a distinct, level layer without significant mixing. Following this, the hands adjust the spoon's position slightly upwards, holding it just over the newly formed green layer. Then, an opaque, cream-colored liquid, possessing an even lower density than the green liquid, is poured using the same technique over the back of the spoon, flowing down gently to create a third, clearly separated layer resting atop the green one within the confines of the shot glass's rigid walls.
}
\end{mainbox}

\begin{mainbox}{Material Interaction \& Transformation: \textit{The study of how materials respond to external forces or processes, including transformations like melting, freezing, breaking, or chemical change}}
\textsf{\small \textbf{Base Prompt:} Hands hold a smartphone inside a clear ziploc bag, tapping the screen while raindrops fall on the bag.
}\\
\textsf{\small \textbf{Upsampled Prompt:} Human hands firmly grip the sides of a clear, flexible plastic ziploc bag, holding it steady. Inside the sealed bag rests a rigid, rectangular smartphone with its glass screen visible. A finger repeatedly taps onto the thin, pliable plastic surface directly overlying the phone's touch-sensitive screen, the pressure transmitting through the material. Simultaneously, small water droplets, falling under gravity, land and bead up on the exterior surface of the waterproof bag.
}
\mydashrule
\textsf{\small \textbf{Base Prompt:} Water flows from a sink faucet into the scoop of a dustpan, travels down its hollow handle, and pours into a bucket sitting beside the sink.
}\\
\textsf{\small \textbf{Upsampled Prompt:} Clear water flows downwards in a steady stream from the metal nozzle of a sink faucet, driven by gravity. The stream lands directly onto the rigid, concave surface of a plastic dustpan's scoop, which is positioned underneath the faucet. Due to the angle of the dustpan, the accumulating water is channeled towards the opening at the base of the scoop leading into its hollow handle. The water then travels downwards through the enclosed tubular space within the handle. The open end of the handle is positioned directly over the wide opening of a sturdy plastic bucket resting stationary on the floor beside the sink, allowing the water exiting the handle to fall vertically into the bucket.
}
\mydashrule
\textsf{\small \textbf{Base Prompt:} A golf ball is placed on a countertop; it rolls slightly, then the counter is subtly adjusted, and the ball comes to a stop.
}\\
\textsf{\small \textbf{Upsampled Prompt:} A small, dense, hard golf ball with a characteristic dimpled texture is gently placed onto a large, flat, smooth countertop surface. Because of its perfectly spherical shape and the countertop potentially being minutely uneven, the heavy ball starts to roll slowly across the rigid surface. Subsequently, the countertop experiences a subtle, barely perceptible shift in its orientation, adjusting its levelness. This slight change counteracts the initial impetus for rolling, altering the gravitational pull along the surface, and the ball, influenced by friction and the new level plane, decelerates smoothly until it comes to a complete standstill upon the countertop.
}
\end{mainbox}

\begin{mainbox}{Force and Motion: \textit{The domain focused on physical interactions governed by forces, including motion, pushing, pulling, lifting, and properties like inertia}}
\textsf{\small \textbf{Base Prompt:} A strong magnet is slowly moved across a drywall surface, suddenly snapping and sticking to the wall over a stud location.}\\
\textsf{\small \textbf{Upsampled Prompt:} A dense, rigid metallic magnet, possessing a strong magnetic field, is held close to a vertical, painted drywall surface. It is guided slowly and steadily in a continuous motion across the slightly textured face of the wall. As the path of the strong magnet crosses over the position of a hidden ferromagnetic metal stud located just behind the thin drywall layer, the attractive magnetic force increases sharply and suddenly. This abruptly intensified pull overcomes the controlled movement, causing the heavy magnet to accelerate rapidly towards the wall, making firm contact and sticking securely against the drywall surface directly above the unseen stud's location due to the powerful magnetic attraction.}
\mydashrule
\textsf{\small \textbf{Base Prompt:} Soft tortillas are draped and wedged between the bumps of an upside-down muffin pan that has been sprayed with oil.}\\
\textsf{\small \textbf{Upsampled Prompt:} Soft, pliable, circular tortillas, thin and flexible due to their material properties, are draped over the rigid structure formed by an upside-down metal muffin pan. Gravity pulls the lightweight tortillas downwards. They are then carefully pressed and wedged into the concave spaces between the protruding, rounded bumps of the pan. The tortillas bend and conform to the curved shape of these spaces. The pan's surface, both bumps and crevices, is coated in a visible layer of slick cooking oil, facilitating the placement and slight sliding of the tortillas as they are settled into position.}
\mydashrule
\textsf{\small \textbf{Base Prompt:} A hand places a colorful silicone cupcake liner into the empty cup holder of a car's center console.
}\\
\textsf{\small \textbf{Upsampled Prompt:} A hand gently pinches a colorful, flexible silicone cupcake liner between thumb and forefinger. The liner, lightweight and pliable with distinct fluted sides, is carefully lowered vertically into an empty, cylindrical cup holder recessed into the car's rigid plastic center console. The liner slides smoothly against the inner wall and comes to rest flat against the bottom surface of the holder, its flexible structure allowing it to fit snugly within the defined space.}
\end{mainbox}

\section{Evaluation Details}
\label{app:captioning_qa}
To evaluate physical commonsense on PhysVidBench, we employ a three‐stage pipeline: (1) generate targeted yes/no questions from each upsampled prompt, (2) produce multi‐perspective dense captions for the corresponding video, and (3) use an LLM to answer those questions based solely on the captions.

\parahead{Stage 1: Physics‐Grounded Question Generation}
For each upsampled prompt, we prompt Gemini 2.5 Pro to craft precise yes/no questions that probe our seven physical‐commonsense dimensions (e.g., “Did the person apply sufficient force to separate the objects?”). Every question is guaranteed to have a “Yes” answer and is answerable by inspecting the prompt alone. We fix and store this question set for downstream evaluation of each model’s video output. Below is an example prompt used in this stage.

\begin{mainbox}{Sample Prompt for Evaluation -- Stage 1: Physics-Grounded Question Generation}

\textsf{\small Generate as many diverse, concrete yes/no questions as possible based only on the provided prompt. Each question must:}
\begin{itemize}[left=0pt]
    \item \textsf{\small Be specific and answerable solely from the prompt (no outside assumptions)}
    \item \textsf{\small Focus on physical or procedural understanding, not abstract or emotional reasoning}
    \item \textsf{\small Be non-redundant — do not repeat similar ideas or reword the same question}
    \item \textsf{\small Cover a wide range of physical reasoning categories}
    {\item \textsf{\small Be phrased so that the physically correct answer is Yes}}
    \item \textsf{\small  If a question relates to multiple reasoning categories, include all applicable types}
\end{itemize}
\vspace{1em}
\textsf{\small \textbf{Physical Reasoning Categories:} (Choose these types, write only the category like Fundamental Physics, Object Properties etc. )}
\begin{itemize}[left=0pt]
    \item \textsf{\small \textbf{Fundamental Physics:} energy, causality, equilibrium, state change}
    \item \textsf{\small \textbf{Object Properties \& Affordances:} material type, rigidity, softness, containment}
    \item \textsf{\small \textbf{Spatial Reasoning:} fit, position, occlusion, geometry, orientation}
    \item \textsf{\small \textbf{Temporal Dynamics:} ordering, timing, waiting, delays}
    \item \textsf{\small \textbf{Action \& Procedural Understanding:} goal-directed behavior, methodical steps}
    \item \textsf{\small \textbf{Material Interaction \& Transformation:} melting, freezing, breaking, chemical change}
    \item \textsf{\small \textbf{Force and Motion:} pushing, pulling, lifting, inertia}
\end{itemize}
\vspace{1em}

\textsf{\small \textbf{Output Format:}}
\begin{itemize}[left=0pt]
    \item \textsf{\small Questions and Answers:}
    \item \textsf{\small Q: <yes/no question>?}
    \item \textsf{\small A: {Yes}}
    \item \textsf{\small Type: <Reasoning Category 1, Reasoning Category 2, ...>}
\end{itemize}
$\;$\\
\textsf{\small \textbf{You must return:}}
\begin{itemize}[left=0pt]
    \item \textsf{\small A rich, diverse set of yes/no questions}
    \item \textsf{\small {Each formulated so that its correct answer is Yes}}
    \item \textsf{\small  Each labeled with all relevant reasoning categories, separated by commas}
\end{itemize}

\vspace{1em}
\textsf{\small Start with the following prompt:}\\ 
\textsf{\small \textbf{Prompt:} Person firmly grips the handle of an old, worn toothbrush, its plastic slightly yellowed with age. They dip the frayed bristles into a can of thick, white paint. The paint clings to the bristles due to its viscosity. With a controlled wrist motion, they flick the paint-laden toothbrush towards a large, matte black canvas backdrop. Tiny droplets of white paint detach from the bristles and spray outwards, propelled by the force of the flick. The droplets travel through the air, influenced by gravity, before adhering to the vertical surface of the canvas.}
\end{mainbox}

\parahead{Stage 2: Dense Video Captioning} \label{sec:eval-captioning} For each generated video, we use the AuroraCap model \citep{chai2024auroracap} to generate eight distinct captions: one general‐purpose caption and seven dimension‐specific captions (one per commonsense dimension).  For instance, to highlight spatial relationships, we use the prompt ``Describe the spatial layout and object positions…”; to surface force interactions we use ``Describe how forces are applied and resisted $\dots$” {Each prompt ends with an instruction to report only directly observed content, which is intended to reduce unsupported inference while eliciting complementary perspectives.} Below is the full template we provide to AuroraCap in this stage.

\begin{mainbox}{Prompt for Evaluation -- Stage 2: Dense Video Captioning}

\begin{itemize}[left=0pt]
    \item \textsf{\small Describe the video in detail.}
    \item \textsf{\small Describe the physical principles at work in the video, such as energy transfer, causal relationships, balance or imbalance, and any visible changes in physical state and etc. (e.g., solid to liquid).}
    \item \textsf{\small Describe the main objects in the video focusing on their material properties (e.g., rigid, soft, metallic), and what actions they allow or prevent and etc. (e.g., can be squeezed, can contain something).}
    \item \textsf{\small Describe the spatial layout and relationships in the video: object positions, orientations, fit between shapes, occlusions, and how geometry affects interactions and etc..}
    \item \textsf{\small Describe the sequence and timing of events in the video, including any delays, waiting periods, or causal orderings between actions or state changes and etc..}
    \item \textsf{\small Describe the actions performed in the video, focusing on the goals of the agent, the order of steps taken, and whether the actions appear intentional or methodical and etc..}
    \item \textsf{\small Describe how materials change or interact in the video: melting, freezing, breaking, mixing, or undergoing chemical or physical transformations and etc..}
    \item \textsf{\small Describe how forces are applied in the video (e.g., pushing, pulling, lifting), how objects respond (e.g., acceleration, resistance), and any indications of inertia or physical resistance and etc..}
\end{itemize}

    \textsf{\small Only describe what can be directly observed in the video. Do not make assumptions or include external knowledge that is not visually confirmed.}
\end{mainbox}

\parahead{Stage 3: LLM as a Judge}  We assemble a single text prompt that includes: (a) all yes/no questions generated in Stage 1, and (b) the eight dense captions produced in Stage 2. We pass this prompt to Gemini-2.5-Flash-Preview-04-17~\citep{geminiflash}, instructing it to respond only with “Yes” or “No” for each question based solely on the provided captions. Below, we present the exact Stage 3 prompt used to evaluate the sample video generated by the Wan2.1 (14B) model from the Stage 1 prompt. This prompt combines the eight dense captions produced in Stage 2 with the corresponding physical‐commonsense questions from Stage 1, illustrating how the LLM integrates multi‐perspective textual evidence and targeted queries during evaluation.

{\hspace{-0.4cm}\includegraphics[width=\columnwidth]{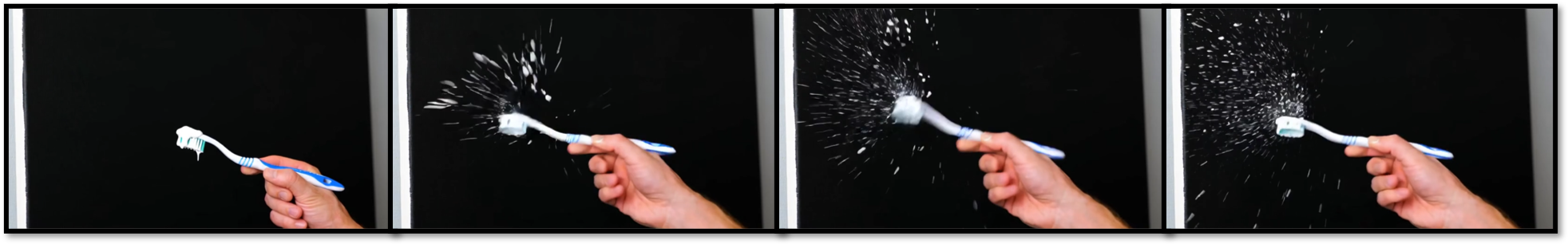}}

\begin{mainbox}{Example Prompt for Evaluation -- Stage 3: LLM as a Judge}
\textsf{\small You are given 8 captions describing different aspects of the same video.  
Answer “Yes” ONLY if at least one caption supports it, otherwise answer “No”.}

\vspace{1em}

\textsf{\small \textbf{Caption 1: }The video captures a moment where a person is using a toothbrush to apply a white substance, likely toothpaste, onto a dark surface. The toothbrush is held in a way that suggests the person is in the process of brushing, with the bristles making contact with the surface. The substance is being applied in a way that creates a textured effect, with some areas appearing more saturated than others. The background is a solid, dark color, providing a stark contrast to the bright white of the toothpaste. The lighting is even, allowing for clear visibility of the toothbrush and the substance being applied. The focus is on the action of brushing, with the background remaining unchanged, emphasizing the simplicity of the scene.}\\

\textsf{\small \textbf{Caption 2: }The video captures a moment where a toothbrush is being used to apply a white substance, likely toothpaste, onto a dark surface. The energy transfer is evident as the bristles of the toothbrush create a splatter effect, with the white toothpaste dispersing in a way that suggests a mix of solid and liquid components. The balance of the scene is slightly tilted, with the toothbrush being the focal point, while the dark surface provides a contrast that highlights the action. The physical state of the toothbrush and toothpaste is clearly observable; the toothbrush is in motion, and the toothpaste is in a semi-liquid state, transitioning from a solid to a liquid form as it spreads.}\\

\textsf{\small \textbf{Caption 3: }The main objects in the video are a toothbrush and a paintbrush. The toothbrush, with its soft bristles, is designed for cleaning teeth, allowing it to gently scrub surfaces. The paintbrush, with its metallic handle, is used for applying paint, preventing the paint from dripping and allowing for controlled application. The actions they allow are brushing and painting, respectively, while the toothbrush's soft bristles prevent abrasion and the paintbrush's handle allows for a firm grip.}\\

\textsf{\small \textbf{Caption 4: }The video captures a moment where a hand is applying a white substance, likely paint, to a dark surface using a toothbrush. The toothbrush is positioned diagonally across the frame, with the bristles facing the camera. The paint is being applied in a way that creates a textured effect, with some areas showing a smooth application and others showing a more uneven, brush-like texture. The hand is in the process of painting, with the fingers gripping the toothbrush and the thumb supporting it. The background is a solid, dark color, providing a stark contrast to the white paint, which enhances the visibility of the brush strokes. The overall layout is simple, focusing on the action of painting without any distractions.}\\

\textsf{\small \textbf{Caption 5: }The video captures a moment where a person is applying a white substance, likely paint, to a dark surface using a toothbrush. The action is deliberate and focused, with the bristles of the toothbrush making contact with the surface, creating a small, white splatter. The hand holding the toothbrush is steady, with fingers gripping the handle firmly. The background is a solid, dark color, providing a stark contrast to the white paint, which enhances the visibility of the splatter. The lighting is bright, casting a soft shadow of the toothbrush on the surface. The overall scene is static, with no movement or change in the environment, allowing the viewer to concentrate on the application process.}\\

\textsf{\small \textbf{Caption 6: }The video captures a moment where a person is applying a white substance, likely paint, to a dark surface using a toothbrush. The actions are deliberate and show a methodical approach, with the substance being carefully squeezed from a tube and then applied in a controlled manner. The agent's hand is steady, with fingers positioned to hold the tube and the toothbrush, indicating a focus on precision. The substance is being applied in a way that suggests a deliberate effort to create a specific effect, possibly for artistic or decorative purposes. The overall scene is calm and focused, with the agent's hand being the main point of interest.}\\

\textsf{\small \textbf{Caption 7: }The video captures a moment where a toothbrush is being used to apply a white substance, likely a cleaning agent, to a dark surface. The substance appears to be in a liquid state, as evidenced by the way it drips and pools on the surface. The toothbrush is held in a way that suggests it is being used to spread the substance, with the bristles making contact with the surface. The fingers of the hand holding the toothbrush are visible, with one finger slightly raised, indicating the action of applying pressure. The background is a solid, dark color, providing a stark contrast to the white substance, which emphasizes the cleaning process.}\\

\textsf{\small \textbf{Caption 8: }The video captures a moment where a toothbrush is being used to apply a white substance, likely toothpaste, to a dark surface. The bristles of the toothbrush are in contact with the surface, and the substance is being squeezed out, creating a small puddle. The hand holding the toothbrush is steady, with fingers gripping the handle, indicating a controlled application. The surface appears smooth, with no visible resistance, allowing the substance to spread easily. The lighting is bright, casting a soft shadow of the toothbrush on the surface, enhancing the visibility of the application process.}

\vspace{1em}

\textsf{\small \textbf{Q1:} Is the toothbrush handle described as plastic?}\\  
\textsf{\small \textbf{Q2:} Did the paint cling to the bristles?}\\  
\textsf{\small \textbf{Q3:} Was the paint stored in a can?}\\  
\textsf{\small \textbf{Q4:} Was the canvas described as black?}\\
\textsf{\small \textbf{Q5:} Was the canvas backdrop described as vertical?} \\
\textsf{\small \textbf{Q6:} Did the person perform a flicking motion with the toothbrush? } \\
\textsf{\small \textbf{Q7:} Did droplets of paint detach from the bristles?  }\\
\textsf{\small \textbf{Q8:} Was the detachment of paint caused by the force of the flick?  }\\
\textsf{\small \textbf{Q9:} Did the paint droplets travel through the air?  }\\
\textsf{\small \textbf{Q10:} Was gravity mentioned as influencing the droplets' travel? }\\
\textsf{\small \textbf{Q11:} Did the paint droplets adhere to the canvas surface?  }\\
\textsf{\small \textbf{Q12:} Was the person's wrist motion described as controlled?  }\\
\textsf{\small \textbf{Q13:} Did the person dip the bristles into the paint? }\\ 
\textsf{\small \textbf{Q14:} Was the paint described as thick?  }\\
\textsf{\small \textbf{Q15:} Did the flicking motion require energy input from the person?}\\

\vspace{1em}

\textsf{\small \textbf{Respond only as:}}\\ 
\textsf{\small Q1: Yes}\\
\textsf{\small Q2: No}\\
\textsf{\small ...}

\end{mainbox}

{\section{A Taxonomy of Physically Grounded Failure Modes} 
\label{app:failure_modes}
To analyze \emph{why} certain prompts challenge current T2V models, we constructed a structured annotation framework that captures the physical-reasoning demands underlying each prompt, beyond surface interaction type. Every prompt is characterized along three complementary axes:

\paragraph{Action Complexity (AC).}
\begin{itemize}[leftmargin=1em]
    \item {\textbf{AC-1 (Direct):} A single, simple action.}
    \item {\textbf{AC-2 (Goal-Oriented):} A preparatory or two-step action aimed at achieving a specific result.}
    \item {\textbf{AC-3 (Procedural):} A multi-step, complex procedure where correct sequencing is essential.}
\end{itemize}

\paragraph{Scientific Principle (SP).}
\begin{itemize}[leftmargin=1em]
    \item {\textbf{SP-1 (Basic):} Straightforward mechanics with no salient principle.}
    \item {\textbf{SP-2 (Applied):} Application of a common, understandable physical principle (e.g., friction, leverage).}
    \item {\textbf{SP-3 (Complex/Hidden):} Relies on less obvious or complex, multi-stage principles (e.g., negative pressure, magnetism).}
\end{itemize}

\paragraph{Object Functionality (OF).}
\begin{itemize}[leftmargin=1em]
    \item {\textbf{OF-1 (Intended Use):} Object used for its designed purpose.}
    \item {\textbf{OF-2 (Repurposed):} Object employed as an improvised tool.}
    \item {\textbf{OF-3 (System Component):} Object functioning within a broader physical system.}
\end{itemize}

\paragraph{Illustrative Examples.}
\begin{itemize}[leftmargin=1em]
    \item {\textbf{Example 1 (AC-1, SP-1, OF-1):}}\\
    {“Hands pour a handful of mixed coins into the opening of an empty, clear plastic milk jug.”}\\
    {$\rightarrow$ \emph{Direct transfer using object as intended; straightforward use of objects.}}

    \item {\textbf{Example 2 (AC-1, SP-1, OF-2):}}\\
    {“A plastic bucket is placed upside down on someone’s head like a hat.”}\\
    {$\rightarrow$ \emph{Simple physical manipulation but repurposed functionality.}}

    \item {\textbf{Example 3 (AC-1, SP-2, OF-3):}}\\
    {“A lit flashlight inside a translucent plastic bag makes the bag glow like a lantern.”}\\
    {$\rightarrow$ \emph{Requires reasoning about light diffusion and emergent system behavior.}}

    \item {\textbf{Example 4 (AC-2, SP-3, OF-1):}}\\
    {“A person melts a toy soldier with a magnifying glass focusing sunlight.”}\\
    {$\rightarrow$ \emph{Goal-driven use of a non-obvious scientific principle (solar heat focus).}}

    \item {\textbf{Example 5 (AC-3, SP-1, OF-1):}}\\
    {“Hands pour colorful powder into a small cup, press it down firmly, then invert the cup and tap it, releasing a shaped bath bomb onto a surface.”}\\
    {$\rightarrow$ \emph{Multi-step procedural action with material handling.}}

    \item {\textbf{Example 6 (AC-3, SP-3, OF-3):}}\\
    {“Hands squeeze an empty plastic water bottle, place the opening over an egg yolk, and release the squeeze, sucking the yolk into the bottle.”}\\
    {$\rightarrow$ \emph{Complex, multi-step use of pressure dynamics and system-level interaction.}}\\
\end{itemize}

To apply this taxonomy at scale, we used Gemini 2.5 Pro to classify every prompt in our Medium, Hard, and Very Hard subsets. {As shown in Table~\ref{tab:taxonomy-scores}, the average scores increase consistently across all three axes, with harder groups involving more complex procedures, physical principles, and object functionality.}

\begin{table}[!h]
\caption{\textbf{Average taxonomy scores across model-based difficulty groups.} {AC, SP, and OF denote Action Complexity, Scientific Principle, and Object Functionality, respectively. Each measure increases consistently from Medium to Hard to Very Hard, showing that the performance-based grouping also corresponds to progressively richer and more demanding physical-reasoning structure.}}
\centering
\resizebox{\columnwidth}{!}{
\begin{tabular}{lcccc}
\toprule
{\textbf{Difficulty Subset}} & {\textbf{AC}} & {\textbf{SP}} & {\textbf{OF}} & {\textbf{Overall Score}} \\
\midrule
{Medium Prompts} & {1.48} & {1.65} & {1.62} & {4.75} \\
{Hard Prompts} & {1.69} & {1.77} & {1.90} & {5.36} \\
{Very Hard Prompts} & {2.06} & {1.94} & {2.10} & {6.10} \\
\bottomrule
\end{tabular}
}
\label{tab:taxonomy-scores}
\end{table}

{This structured analysis provides fine-grained diagnostic clarity: models often succeed on simple interactions (AC-1, SP-1, OF-1) but fail when tasks require multi-step procedures, non-obvious physical principles, or functional recontextualization. This taxonomy thus enhances PhysVidBench’s value as a diagnostic tool, revealing not only how well models perform, but why they fail.}

\section{An End-to-End Example Evaluation}
\label{app:e2e_example}
\begin{mainbox}{Prompt Generation}
\textbf{Stage 1. Input: Human Commonsense Goal + Solutions (PIQA)} \vspace{0.4em}\\
Goal: \textit{Bag}\\
Solution: \textit{can be used as a lantern with a flashlight indoors.}
\mydashrule
\textbf{Stage 2. Base Prompt:} \vspace{0.4em}\\
A lit flashlight inside a translucent plastic bag makes the bag glow like a lantern.
\mydashrule
\textbf{Stage 3. Upsampled Prompt:} \vspace{0.4em}\\
A cylindrical flashlight, currently switched on and emitting a focused beam of bright light, is positioned entirely inside a thin, flexible, translucent plastic bag. The lightweight plastic material loosely conforms around the flashlight. Light rays emanating from the flashlight's bulb strike the inner walls of the bag; due to the plastic's translucent property, the light scatters and diffuses as it passes through, causing the entire volume of the bag to glow softly from within, resembling a makeshift lantern in a dimly lit space.
\end{mainbox}

\begin{mainbox}{Evaluation}
\textbf{Stage 1. Physics-Grounded Question Generation:}\vspace{0.4em}\\
We generate questions that cover seven key physical commonsense dimensions defined in our ontology:\vspace{0.4em}\\
\textbf{AU:} Action \& Procedural Understanding, \textbf{FM:} Force and Motion, \textbf{FP:} Fundamental Physics, \textbf{MT:} Material Interaction \& Transformation, \textbf{OP:} Object Properties \& Affordances, \textbf{SR:} Spatial Reasoning, \textbf{TD:} Temporal Dynamics.
\begin{center}
\resizebox{\columnwidth}{!}{\begin{tabular}{@{}p{0.8cm}p{8cm}p{2cm}@{}}
\toprule
\textbf{\#} & \textbf{Question} & \textbf{Dimension(s)} \\ 
\midrule
Q1  & Is the flashlight emitting light energy? & FP \\
Q2  & Is the bag made of plastic? & OP \\
Q3  & Is the flashlight entirely inside the bag? & SR, OP \\
Q4  & Is the bag described as flexible? & OP \\
Q5  & Does the light scatter as it passes through the plastic? & MT, FP \\
Q6  & Does the flashlight being on cause the bag to glow? & FP \\
Q7  & Is the flashlight described as cylindrical? & SR, OP \\
Q8  & Does the bag glow softly? & FP \\
Q9  & Is the bag described as lightweight? & OP \\
Q10 & Does the setup resemble a makeshift lantern? & AU \\
Q11 & Is the light emitted by the flashlight described as a focused beam initially? & FP, OP \\
Q12 & Does the entire volume of the bag glow? & SR, FP \\
Q13 & Does the bag diffuse the light passing through it? & MT, FP \\
\bottomrule
\end{tabular}
}
\end{center}
\vspace{0.4em}
\noindent\hdashrule[0.5ex]{1.02\textwidth}{0.4pt}{1mm}\par
\textbf{Stage 2. Dense Video Captioning:} \vspace{0.4em}\\
The generated video is captioned using eight captions: one general-purpose caption and seven dimension-specific captions:\vspace{1em}\\
\textbf{General:\\}
\textit{The video features a close-up view of a bright light source, which appears to be a flashlight or a similar handheld light. The light is emitting a strong, white beam that illuminates the surrounding area with a soft glow. The light is focused and directed, creating a halo effect that enhances the visibility of the surrounding environment. The background is blurred, but hints of blue and white colors suggest a cool, possibly outdoor setting. The light source is partially obscured by a translucent, plastic bag, which adds a layer of texture to the scene. The overall composition is centered on the light, drawing attention to its brightness and the way it interacts with the environment}\vspace{1em}\\
\textbf{Fundamental Physics (FP):\\}
\textit{The video captures a moment where a light source, likely a bulb, is enveloped in a translucent material, creating a visual effect that resembles a flower. The energy transfer is evident as the light shines through the material, creating a soft glow that illuminates the surrounding area. The material appears to be in a state of tension, with the light source at the center, creating a dynamic interplay of light and shadow. The balance is slightly tilted, with the light source being the focal point, drawing the viewer's attention. The physical state of the material is fluid, allowing the light to pass through, which changes the overall appearance of the object from a solid to a more ethereal, glowing form.}\vspace{1em}\\
\textbf{Object Properties \& Affordances (OP):\\}
\textit{The main objects in the video are the plastic bag and the light bulb. The plastic bag, with its smooth and reflective surface, allows light to pass through, creating a soft glow. The light bulb, with its metallic base, emits a bright light that illuminates the bag. The actions that can be directly observed are the light bulb emitting light and the plastic bag reflecting and diffusing the light. The materials of the bag and bulb are visually confirmed as soft and metallic, respectively. The bulb's rigidity is suggested by its shape and the way it stands upright.
}\vspace{1em}\\
\textbf{Spatial Reasoning (SR):\\}
\textit{The video captures a close-up view of a light bulb, which is the main subject. The bulb is encased in a clear, plastic bag, creating a soft glow that illuminates the surrounding area. The bag appears to be slightly crumpled, adding texture to the visual. The light emanating from the bulb creates a bright, circular area that contrasts with the darker background. The bulb is positioned centrally in the frame, with the plastic bag slightly tilted, allowing the light to shine outward. The overall layout is simple, focusing on the interplay of light and shadow.
}\vspace{1em}\\
\textbf{Temporal Dynamics (TD):\\}
\textit{The video captures a moment where a bright light source, likely a flashlight, is being directed towards a translucent, reflective surface, creating a dynamic interplay of light and shadow. The light appears to be moving, as evidenced by the varying intensity of the illumination and the shifting patterns of light and dark. The surface reflects the light, creating a soft glow that enhances the visual effect. The overall atmosphere is ethereal and slightly blurred, with the focus primarily on the light source and its reflection. The light source is the main subject, with the surrounding environment remaining indistinct, allowing the viewer to concentrate on the interplay of light and shadow.
}\vspace{1em}\\
\textbf{Action \& Procedural Understanding (AU):\\}
\textit{The video captures a moment where a bright light source, possibly a flashlight, is being directed towards a translucent, reflective surface, creating a dynamic interplay of light and shadow. The light source is the main focus, with its beam creating a luminous effect that illuminates the surrounding area. The reflective surface enhances the visual impact, as it captures and distorts the light, creating a sense of depth and movement. The agent's actions are deliberate, as they manipulate the light source to achieve the desired effect. The overall goal appears to be the exploration of light and reflection, showcasing the interplay between the light source and the reflective surface.
}\vspace{1em}\\
\textbf{Material Interaction \& Transformation (MT):\\}
\textit{The video captures a moment where a clear plastic bag is being manipulated, likely by a hand, as it is folded and twisted. The bag's transparency allows the viewer to see the light source within, creating a bright, radiant effect. The light source appears to be the main focus, with its glow illuminating the bag and the surrounding environment. The bag's material, while initially clear, shows signs of distortion, possibly due to the folding and twisting actions. The light creates a soft glow that reflects off the bag's surface, enhancing the visual appeal of the scene. The overall interaction between the bag and the light source is one of visual interest, with the bag's transparency and the light's brightness creating a striking contrast.
}\vspace{1em}\\
\textbf{Force and Motion (FM):\\}
\textit{The video captures a moment where a clear plastic bag is being manipulated, likely by a hand, as evidenced by the slight blur around the edges of the bag. The bag is being stretched and folded, creating a dynamic interplay of light and shadow. The bright light source creates a lens flare effect, which adds to the visual complexity. The bag's transparency allows for a view of the light source, which is bright and white, indicating a strong light source. The bag's surface reflects the light, creating a shimmering effect. The overall scene is one of gentle interaction, with the bag responding to the touch of the hand, creating a fluid motion that is both delicate and deliberate.
}
\mydashrule
\textbf{Stage 3. QA-Based Evaluation (LLM-as-a-Judge)} \vspace{0.4em}\\
\begin{center}
\resizebox{\columnwidth}{!}{\begin{tabular}{@{}p{0.8cm}p{8cm}p{2cm}@{}}
\toprule
\textbf{\#} & \textbf{Question} & \textbf{Answer} \\ 
\midrule
Q1  & Is the flashlight emitting light energy? & Yes \\
Q2  & Is the bag made of plastic? & Yes \\
Q3  & Is the flashlight entirely inside the bag? & Yes \\
Q4  & Is the bag described as flexible? & Yes \\
Q5  & Does the light scatter as it passes through the plastic? & Yes \\
Q6  & Does the flashlight being on cause the bag to glow? & Yes \\
Q7  & Is the flashlight described as cylindrical? & No \\
Q8  & Does the bag glow softly? & Yes \\
Q9  & Is the bag described as lightweight? & No \\
Q10 & Does the setup resemble a makeshift lantern? & No \\
Q11 & Is the light emitted by the flashlight described as a focused beam initially? & Yes \\
Q12 & Does the entire volume of the bag glow? & No \\
Q13 & Does the bag diffuse the light passing through it? & Yes \\
\bottomrule
\end{tabular}
}
\end{center}
\end{mainbox}

\newpage
\section{Additional Generation Results and Qualitative Examples}
\label{app:qualitative}

\subsection{Qualitative Examples}
To illustrate the behaviors and failure modes of current text-to-video models on PhysVidBench, we present several representative qualitative cases. For each case, we show: (1) the input text prompt, (2) key frames from videos generated by two contrasting models, and (3) a subset of yes/no QA outcomes used for scoring. \textit{Note that only four questions are shown per example for brevity; the full evaluation uses a broader set of questions covering multiple physical-commonsense dimensions.} These examples highlight where models succeed or break down across different reasoning skills.

\begin{figure*}[!h]
\centering
\includegraphics[width=0.9\linewidth]{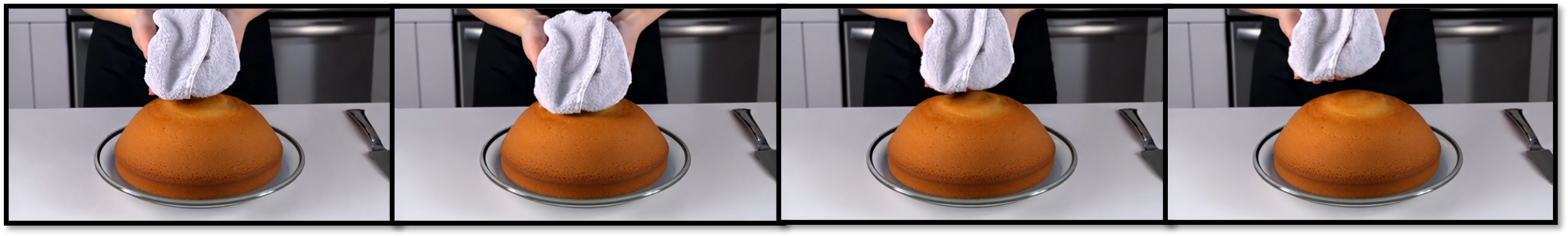}

\caption*{\textbf{Prompt:} \textsf{\small Hands place a tea towel over a freshly baked cake with a domed top and gently press down, flattening the dome.}}
\vspace{0.5em}

\begin{tcolorbox}[colback=gray!5!white, colframe=gray!50!black, title=Prompt-based Evaluation Questions, boxrule=0.5pt, arc=1mm]

\textbf{Model: Cosmos (14B)}

\vspace{0.5em}
\textsf{\small \textbf{Q1:} Does the fabric drape smoothly over the cake's dome?}\\
\textsf{\small \textbf{A1:} \blue{Yes}} 

\vspace{0.3em}
\textsf{\small \textbf{Q2:} Is the tea towel described as rectangular?}\\
\textsf{\small \textbf{A2:} \red{No}}

\vspace{0.3em}
\textsf{\small \textbf{Q3:} Does the cake visibly flatten as a result of the pressure?}\\
\textsf{\small \textbf{A3:} \blue{Yes}}

\vspace{0.3em}
\textsf{\small \textbf{Q4:} Does the warmth of the cake contribute to its structure yielding under pressure?}\\
\textsf{\small \textbf{A4:} \red{No}}

\end{tcolorbox}
\label{fig:qual_tea_towel}
\end{figure*}

\begin{figure*}[!h]
\centering
\includegraphics[width=0.9\linewidth]{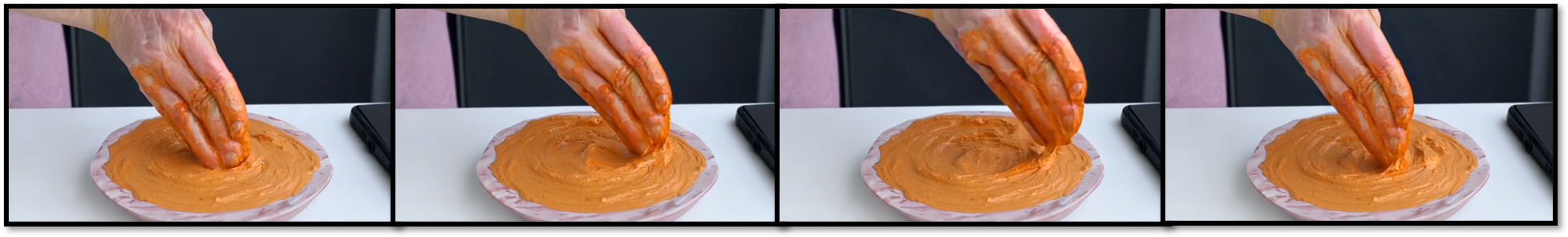}

\caption*{\textbf{Prompt:} \textsf{\small Hands mix baking soda and water in a small bowl to form a paste, then scoop some paste and rub it onto skin with orange streaks.}}
\vspace{0.5em}

\begin{tcolorbox}[colback=gray!5!white, colframe=gray!50!black, title=Prompt-based Evaluation Questions, boxrule=0.5pt, arc=1mm]

\textbf{Model: Cosmos (14B)}

\vspace{0.5em}
\textsf{\small \textbf{Q1:} Did fingers mix the contents of the bowl?}\\
\textsf{\small \textbf{A1:} \blue{Yes}} 

\vspace{0.3em}
\textsf{\small \textbf{Q2:} Was the resulting paste described as opaque?}\\
\textsf{\small \textbf{A2:} \red{No}}

\vspace{0.3em}
\textsf{\small \textbf{Q3:} Were there orange streaks on the surface where the paste was applied?}\\
\textsf{\small \textbf{A3:} \red{No}}

\vspace{0.3em}
\textsf{\small \textbf{Q4:} Was the bowl used for containment of the ingredients?}\\
\textsf{\small \textbf{A4:} \blue{Yes}}

\end{tcolorbox}
\label{fig:qual_tea_towel2}
\end{figure*}

\newpage

\begin{figure*}[!h]
\centering
\includegraphics[width=0.9\linewidth]{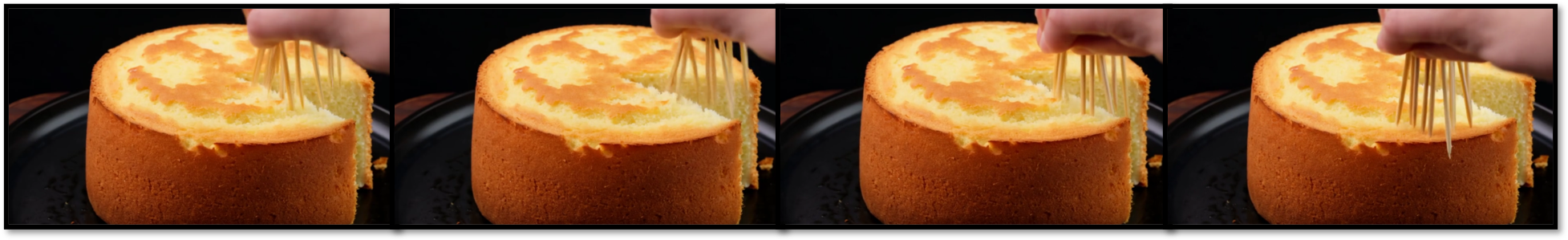}

\caption*{\textbf{Prompt:} \textsf{\small Toothpicks are pushed through slices of bread, securing them against the cut edges of a leftover cake.}}
\vspace{0.5em}

\begin{tcolorbox}[colback=gray!5!white, colframe=gray!50!black, title=Prompt-based Evaluation Questions, boxrule=0.5pt, arc=1mm]

\textbf{Model: Hunyuan}

\vspace{0.5em}
\textsf{\small \textbf{Q1:} Are the toothpicks made of wood?}\\
\textsf{\small \textbf{A1:} \blue{Yes}} 

\vspace{0.3em}
\textsf{\small \textbf{Q2:} Are the toothpicks held firmly?}\\
\textsf{\small \textbf{A2:} \red{No}}

\vspace{0.3em}
\textsf{\small \textbf{Q3:} Do the toothpicks embed themselves within the cake's interior?}\\
\textsf{\small \textbf{A3:} \blue{Yes}}

\vspace{0.3em}
\textsf{\small \textbf{Q4:} Is the bread slice secured against the cake by the action?}\\
\textsf{\small \textbf{A4:} \blue{Yes} }

\end{tcolorbox}
\label{fig:qual_tea_towel3}
\end{figure*}

\begin{figure*}[!h]
\centering
\includegraphics[width=0.9\linewidth]{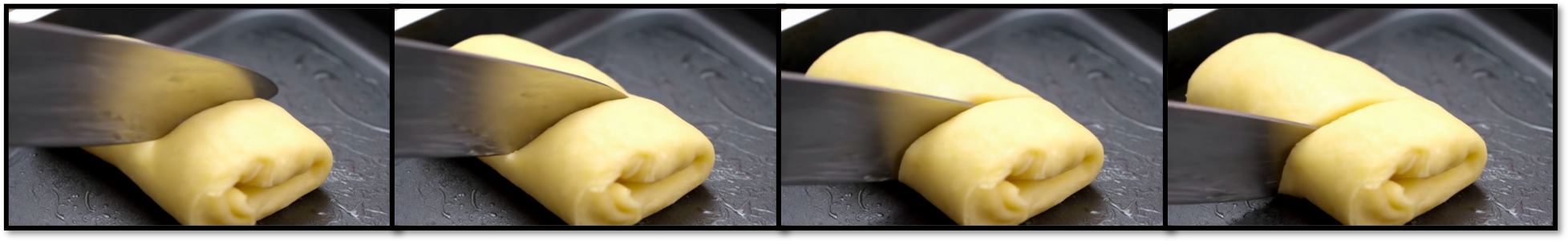}

\caption*{\textbf{Prompt:} \textsf{\small The edge of a metal spoon presses down into a roll of soft potato dough, cutting off a small piece.}}
\vspace{0.5em}

\begin{tcolorbox}[colback=gray!5!white, colframe=gray!50!black, title=Prompt-based Evaluation Questions, boxrule=0.5pt, arc=1mm]

\textbf{Model: Hunyuan}

\vspace{0.5em}
\textsf{\small \textbf{Q1:} Does the action result in separating a portion of the dough?}\\
\textsf{\small \textbf{A1:} \blue{Yes}} 

\vspace{0.3em}
\textsf{\small \textbf{Q2:} Is the dough described as having a cylindrical shape?}\\
\textsf{\small \textbf{A2:} \blue{Yes}}

\vspace{0.3em}
\textsf{\small \textbf{Q3:} Does the spoon shear through the dough's structure?}\\
\textsf{\small \textbf{A3:} \red{No}}

\vspace{0.3em}
\textsf{\small \textbf{Q4:} Is the spoon described as being made of metal?}\\
\textsf{\small \textbf{A4:} \red{No}}

\end{tcolorbox}
\label{fig:qual_tea_towel4}
\end{figure*}

\newpage

\begin{figure*}[!h]
\centering
\includegraphics[width=0.9\linewidth]{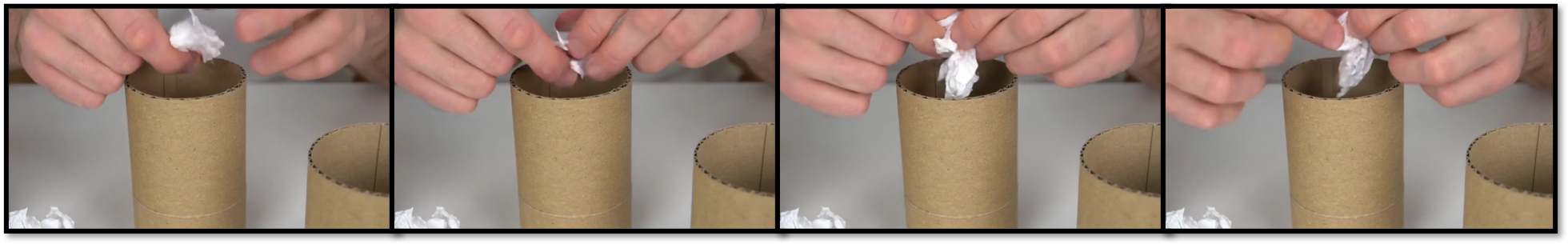}

\caption*{\textbf{Prompt:} \textsf{\small Hands stuff dryer lint into cardboard toilet paper tubes.}}
\vspace{0.5em}

\begin{tcolorbox}[colback=gray!5!white, colframe=gray!50!black, title=Prompt-based Evaluation Questions, boxrule=0.5pt, arc=1mm]

\textbf{Model: Wan2.1 (14B)}

\vspace{0.5em}
\textsf{\small \textbf{Q1:} Is the tube held steady by one hand?}\\
\textsf{\small \textbf{A1:} \red{No}} 

\vspace{0.3em}
\textsf{\small \textbf{Q2:} Are two hands involved in the overall action described?}\\
\textsf{\small \textbf{A2:} \blue{Yes}}

\vspace{0.3em}
\textsf{\small \textbf{Q3:} Does the volume occupied by the lint decrease when pushed into the tube?}\\
\textsf{\small \textbf{A3:} \blue{Yes}}

\vspace{0.3em}
\textsf{\small \textbf{Q4:} Is force applied by fingers to push the lint?}\\
\textsf{\small \textbf{A4:} \blue{Yes}}

\end{tcolorbox}
\label{fig:qual_tea_towel5}
\end{figure*}

\begin{figure*}[!h]
\centering
\includegraphics[width=0.9\linewidth]{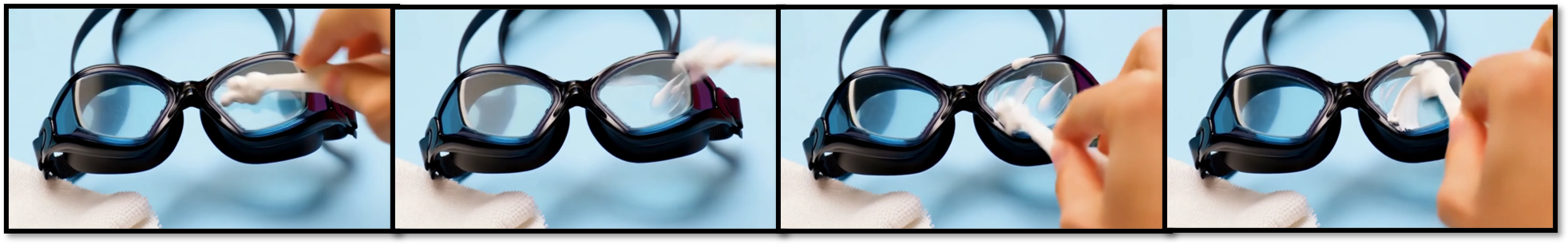}

\caption*{\textbf{Prompt:} \textsf{\small Toothpaste is smeared onto the inner lens of swimming goggles and then wiped away with a soft cloth.}}
\vspace{0.5em}

\begin{tcolorbox}[colback=gray!5!white, colframe=gray!50!black, title=Prompt-based Evaluation Questions, boxrule=0.5pt, arc=1mm]

\textbf{Model: Wan2.1 (14B)}

\vspace{0.5em}
\textsf{\small \textbf{Q1:} Did the toothpaste adhere to the lens surface after application?}\\
\textsf{\small \textbf{A1:} \blue{Yes}} 

\vspace{0.3em}
\textsf{\small \textbf{Q2:} Was the cloth pressed against the lens?}\\
\textsf{\small \textbf{A2:} \blue{Yes}}

\vspace{0.3em}
\textsf{\small \textbf{Q3:} Did the wiping action remove the toothpaste from the lens?}\\
\textsf{\small \textbf{A3:} \red{No}}

\vspace{0.3em}
\textsf{\small \textbf{Q4:} Is the goggle lens described as hard?}\\
\textsf{\small \textbf{A4:} \red{No}}

\end{tcolorbox}
\label{fig:qual_tea_towel6}
\end{figure*}

\newpage

\begin{figure*}[!h]
\centering
\includegraphics[width=0.9\linewidth]{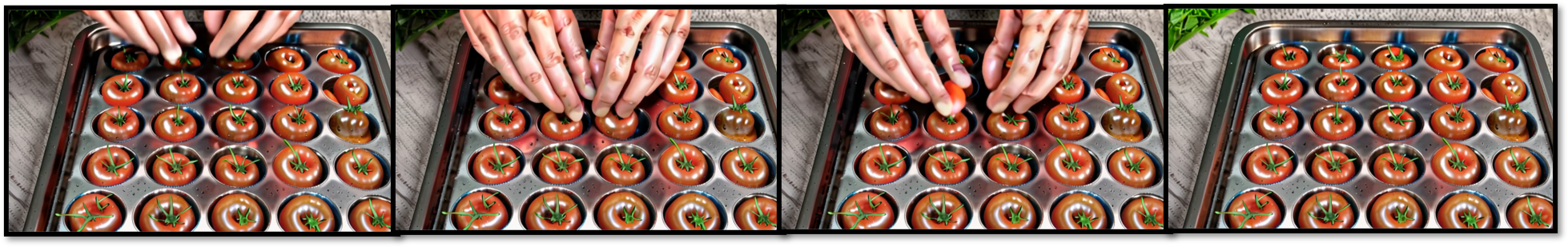}

\caption*{\textbf{Prompt:} \textsf{\small Hands place whole cherry tomatoes into the individual cups of a greased metal muffin tin.}}
\vspace{0.5em}

\begin{tcolorbox}[colback=gray!5!white, colframe=gray!50!black, title=Prompt-based Evaluation Questions, boxrule=0.5pt, arc=1mm]

\textbf{Model: LTX-Video}

\vspace{0.5em}
\textsf{\small \textbf{Q1:} Is the muffin tin made of metal?}\\
\textsf{\small \textbf{A1:} \blue{Yes}} 

\vspace{0.3em}
\textsf{\small \textbf{Q2:} Are the tomatoes described as small and round?}\\
\textsf{\small \textbf{A2:} \red{No}}

\vspace{0.3em}
\textsf{\small \textbf{Q3:} Does gravity cause the released tomato to move downwards?}\\
\textsf{\small \textbf{A3:} \red{No}}

\vspace{0.3em}
\textsf{\small \textbf{Q4:} Are the tomatoes placed into the cups one at a time?}\\
\textsf{\small \textbf{A4:} \blue{Yes}}

\end{tcolorbox}
\label{fig:qual_tea_towel7}
\end{figure*}

\begin{figure*}[!h]
\centering
\includegraphics[width=0.9\linewidth]{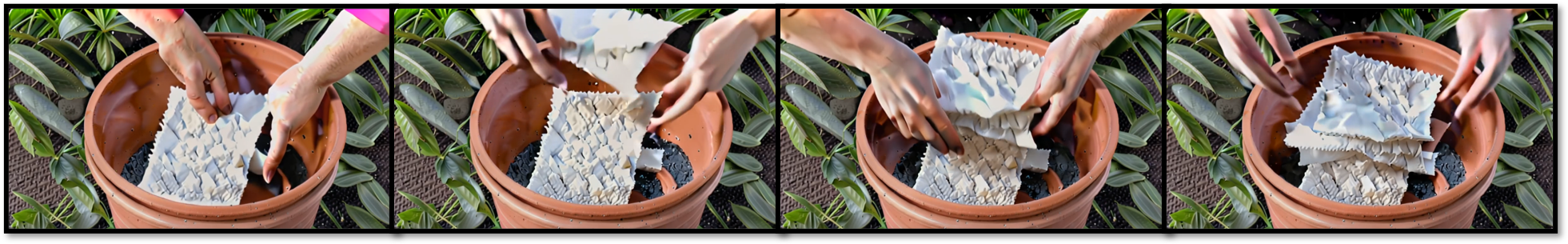}

\caption*{\textbf{Prompt:} \textsf{\small Hands place several sheets of white styrofoam into the bottom of a terracotta planter.}}
\vspace{0.5em}

\begin{tcolorbox}[colback=gray!5!white, colframe=gray!50!black, title=Prompt-based Evaluation Questions, boxrule=0.5pt, arc=1mm]

\textbf{Model: LTX-Video}

\vspace{0.5em}
\textsf{\small \textbf{Q1:} Do the hands lower the sheets into the planter?}\\
\textsf{\small \textbf{A1:} \blue{Yes}} 

\vspace{0.3em}
\textsf{\small \textbf{Q2:} Do the sheets come to rest on the bottom surface of the planter?}\\
\textsf{\small \textbf{A2:} \red{No}}

\vspace{0.3em}
\textsf{\small \textbf{Q3:} Is the styrofoam material described as porous?}\\
\textsf{\small \textbf{A3:} \red{No}}

\vspace{0.3em}
\textsf{\small \textbf{Q4:} Is the planter described as sturdy?}\\
\textsf{\small \textbf{A4:} \red{No}}

\end{tcolorbox}
\label{fig:qual_tea_towel8}
\end{figure*}

\newpage

\begin{figure*}[!h]
\centering
\includegraphics[width=0.9\linewidth]{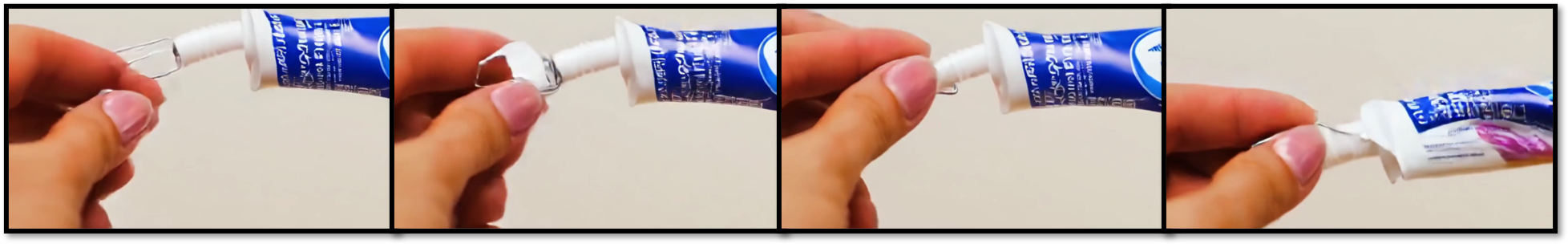}

\caption*{\textbf{Prompt:} \textsf{\small A person attaches a binder clip to the end of a toothpaste tube and squeezes it to get more toothpaste out.}}
\vspace{0.5em}

\begin{tcolorbox}[colback=gray!5!white, colframe=gray!50!black, title=Prompt-based Evaluation Questions, boxrule=0.5pt, arc=1mm]

\textbf{Model: Wan2.1 (1.3B)}

\vspace{0.5em}
\textsf{\small \textbf{Q1:} Does the clip's spring mechanism compress the tube?}\\
\textsf{\small \textbf{A1:} \red{No}} 

\vspace{0.3em}
\textsf{\small \textbf{Q2:} Is the toothpaste described as viscous?}\\
\textsf{\small \textbf{A2:} \red{No}}

\vspace{0.3em}
\textsf{\small \textbf{Q3:} Does the binder clip create a barrier preventing toothpaste from moving back towards the flattened end?}\\
\textsf{\small \textbf{A3:} \red{No}}

\vspace{0.3em}
\textsf{\small \textbf{Q4:} Does the clip help to gather the remaining toothpaste near the opening?}\\
\textsf{\small \textbf{A4:} \red{No}}

\end{tcolorbox}
\label{fig:qual_tea_towel9}
\end{figure*}

\begin{figure*}[!h]
\centering
\includegraphics[width=0.9\linewidth]{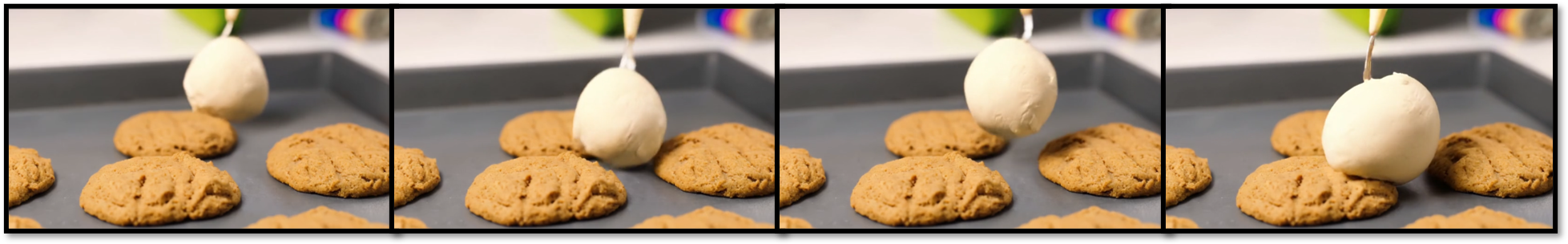}

\caption*{\textbf{Prompt:} \textsf{\small An ice cream scoop digs into cookie dough and releases a perfect ball onto a baking sheet.}}
\vspace{0.5em}

\begin{tcolorbox}[colback=gray!5!white, colframe=gray!50!black, title=Prompt-based Evaluation Questions, boxrule=0.5pt, arc=1mm]

\textbf{Model: Wan2.1 (1.3B)}

\vspace{0.5em}
\textsf{\small \textbf{Q1:} Is the cookie dough described as soft?}\\
\textsf{\small \textbf{A1:} \blue{Yes}} 

\vspace{0.3em}
\textsf{\small \textbf{Q2:} Does the dough yield under pressure from the scoop?}\\
\textsf{\small \textbf{A2:} \red{No}}

\vspace{0.3em}
\textsf{\small \textbf{Q3:} Is the scoop positioned directly above the baking sheet before the dough is released?}\\
\textsf{\small \textbf{A3:} \red{No}}

\vspace{0.3em}
\textsf{\small \textbf{Q4:} Is the ejected dough ball described as spherical?}\\
\textsf{\small \textbf{A4:} \red{No}}

\end{tcolorbox}
\label{fig:qual_tea_towel10}
\end{figure*}

\newpage

\begin{figure*}[!h]
\centering
\includegraphics[width=0.9\linewidth]{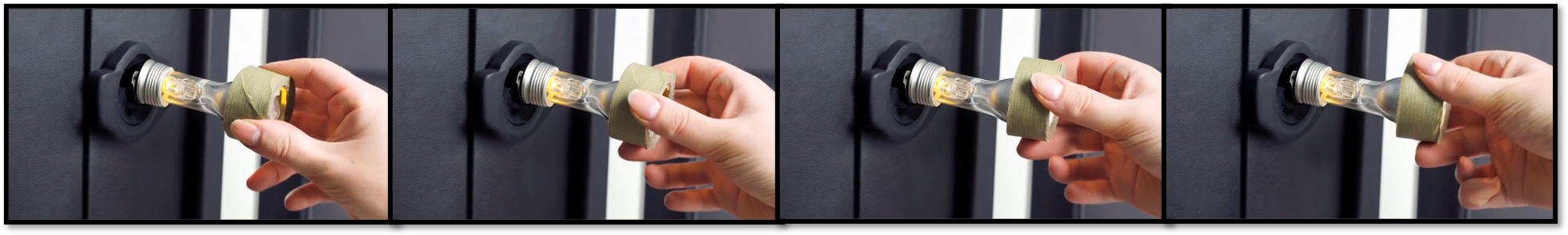}

\caption*{\textbf{Prompt:} \textsf{\small Duct tape is wrapped around a stuck light bulb in a socket, and a hand uses the tape to twist the bulb counter-clockwise, loosening it.}}

\vspace{0.5em}

\begin{tcolorbox}[colback=gray!5!white, colframe=gray!50!black, title=Prompt-based Evaluation Questions, boxrule=0.5pt, arc=1mm]

\textbf{Model: Cosmos (7B)}

\vspace{0.5em}
\textsf{\small \textbf{Q1:} Is the tape described as having a flexible fabric backing?}\\
\textsf{\small \textbf{A1:} \red{No}} 

\vspace{0.3em}
\textsf{\small \textbf{Q2:} Is the tape wrapped multiple times around the bulb?}\\
\textsf{\small \textbf{A2:} \red{No}}

\vspace{0.3em}
\textsf{\small \textbf{Q3:} Does a hand grasp the layered duct tape on the bulb?}\\
\textsf{\small \textbf{A3:} \red{No}}

\vspace{0.3em}
\textsf{\small \textbf{Q4:} Does the hand twist in a counter-clockwise direction?}\\
\textsf{\small \textbf{A4:} \red{No}}

\end{tcolorbox}
\label{fig:qual_tea_towel11}
\end{figure*}

\begin{figure*}[!h]
\centering
\includegraphics[width=0.9\linewidth]{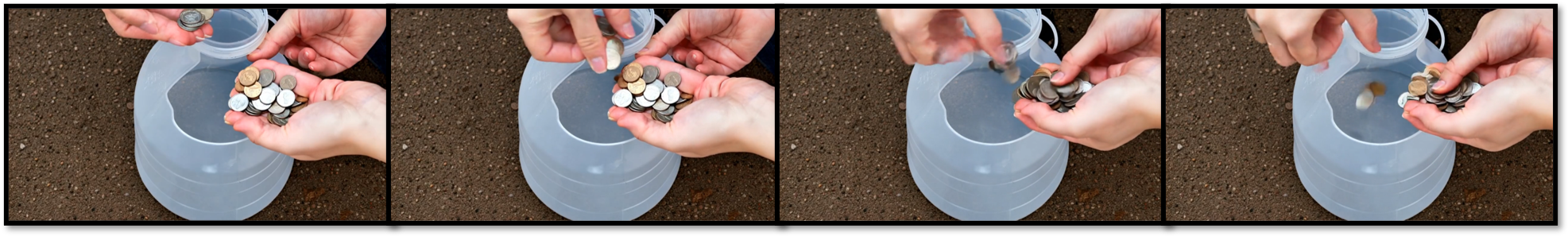}

\caption*{\textbf{Prompt:} \textsf{\small Hands pour a handful of mixed coins into the opening of an empty, clear plastic milk jug.}}

\begin{tcolorbox}[colback=gray!5!white, colframe=gray!50!black, title=Prompt-based Evaluation Questions, boxrule=0.5pt, arc=1mm]

\textbf{Model: Cosmos (7B)}

\vspace{0.5em}
\textsf{\small \textbf{Q1:} Does the action of tilting the hands happen before the coins begin to fall?}\\
\textsf{\small \textbf{A1:} \red{No}} 

\vspace{0.3em}
\textsf{\small \textbf{Q2:} Does gravity influence the movement of the coins after release?}\\
\textsf{\small \textbf{A2:} \red{No}}

\vspace{0.3em}
\textsf{\small \textbf{Q3:} Do the coins pass completely through the jug's opening to enter the body?}\\
\textsf{\small \textbf{A3:} \red{No}}

\vspace{0.3em}
\textsf{\small \textbf{Q4:} Are the coins contained within the hands before being released?}\\
\textsf{\small \textbf{A4:} \red{No}}

\end{tcolorbox}
\end{figure*}

\newpage

\begin{figure*}[!h]
\centering
\includegraphics[width=0.9\linewidth]{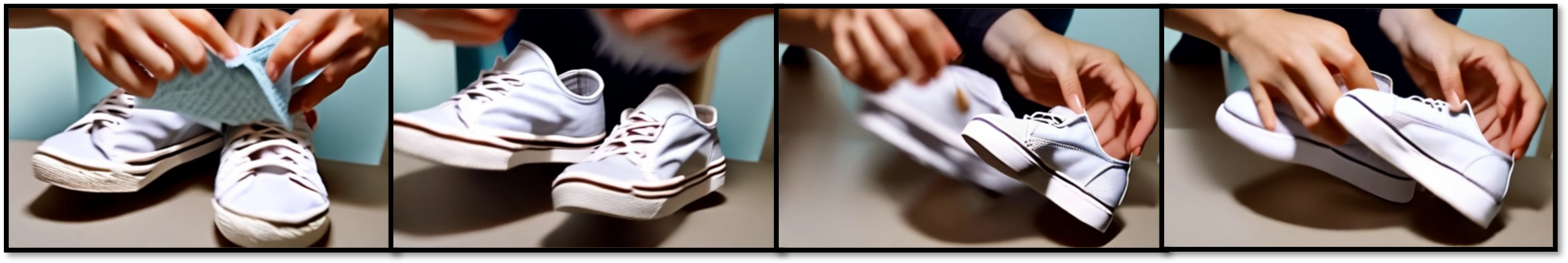}

\caption*{\textbf{Prompt:} \textsf{\small Hands place folded dryer sheets inside a pair of sneakers.}}

\begin{tcolorbox}[colback=gray!5!white, colframe=gray!50!black, title=Prompt-based Evaluation Questions, boxrule=0.5pt, arc=1mm]

\textbf{Model: VideoCrafter2}

\vspace{0.5em}
\textsf{\small \textbf{Q1:} Do the hands push a dryer sheet into the first sneaker? }\\
\textsf{\small \textbf{A1:} \red{No}} 

\vspace{0.3em}
\textsf{\small \textbf{Q2:} Is the action of inserting a dryer sheet performed on both sneakers mentioned in the prompt?}\\
\textsf{\small \textbf{A2:} \red{No}}

\vspace{0.3em}
\textsf{\small \textbf{Q3:} Is the dryer sheet described as flexible? }\\
\textsf{\small \textbf{A3:} \red{No}}

\vspace{0.3em}
\textsf{\small \textbf{Q4:} Is the dryer sheet inserted into a cavity within the sneaker? }\\
\textsf{\small \textbf{A4:} \red{No}}

\end{tcolorbox}
\end{figure*}

\begin{figure*}[!h]
\centering
\includegraphics[width=0.9\linewidth]{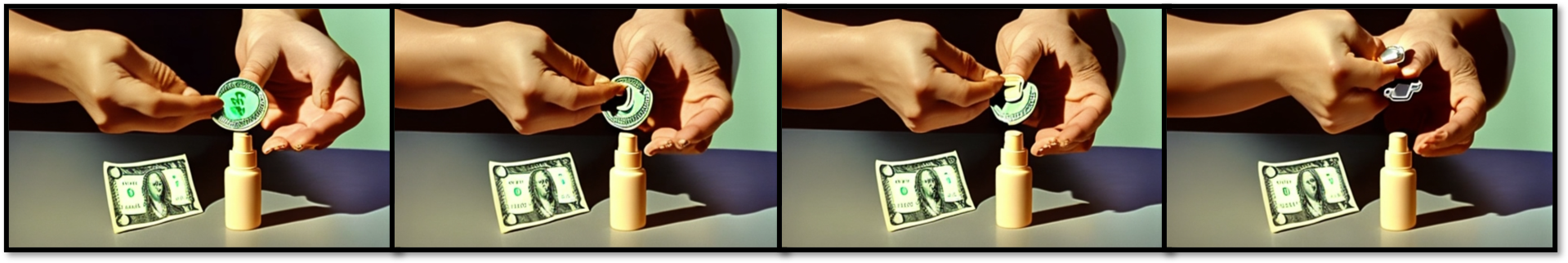}

\caption*{\textbf{Prompt:} \textsf{\small Hands place keys and folded money into a hollowed-out suntan lotion bottle and put the top back on.}}

\begin{tcolorbox}[colback=gray!5!white, colframe=gray!50!black, title=Prompt-based Evaluation Questions, boxrule=0.5pt, arc=1mm]

\textbf{Model: VideoCrafter2}

\vspace{0.5em}
\textsf{\small \textbf{Q1:} Are the keys described as rigid? }\\
\textsf{\small \textbf{A1:} \red{No}} 

\vspace{0.3em}
\textsf{\small \textbf{Q2:} Is the bottle described as hollowed-out? }\\
\textsf{\small \textbf{A2:} \blue{Yes}}

\vspace{0.3em}
\textsf{\small \textbf{Q3:} Are the keys inserted through an opening in the bottle? }\\
\textsf{\small \textbf{A3:} \red{No}}

\vspace{0.3em}
\textsf{\small \textbf{Q4:} Is the bottle made of plastic?}\\
\textsf{\small \textbf{A4:} \blue{Yes}}

\end{tcolorbox}
\end{figure*}

\newpage

\begin{figure*}[!h]
\centering
\includegraphics[width=0.9\linewidth]{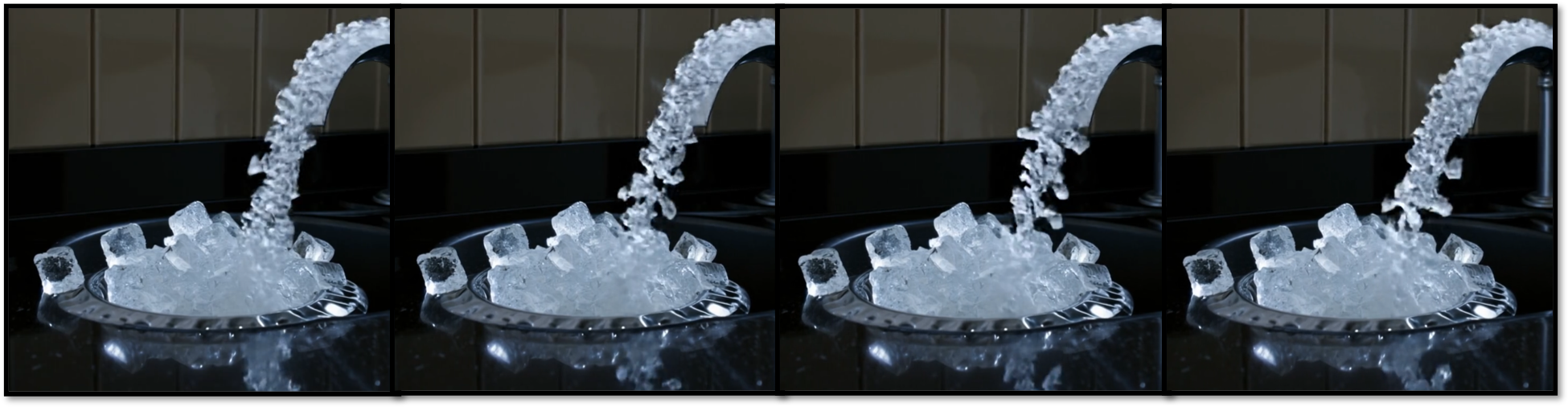}

\caption*{\textbf{Prompt:} \textsf{\small Ice cubes are poured into a kitchen sink drain and ground up by the garbage disposal.}}
\vspace{0.5em}

\begin{tcolorbox}[colback=gray!5!white, colframe=gray!50!black, title=Prompt-based Evaluation Questions, boxrule=0.5pt, arc=1mm]

\textbf{Model: MAGI-1}

\vspace{0.5em}
\textsf{\small \textbf{Q1:} Do the cubes tumble downwards into a cavity?}\\
\textsf{\small \textbf{A1:} \red{No}} 

\vspace{0.3em}
\textsf{\small \textbf{Q2:} Is the disposal chamber described as a confined space?}\\
\textsf{\small \textbf{A2:} \red{No}}

\vspace{0.3em}
\textsf{\small \textbf{Q3:} Do the spinning components collide forcefully with the ice cubes?}\\
\textsf{\small \textbf{A3:} \red{No}}

\vspace{0.3em}
\textsf{\small \textbf{Q4:} Do the internal components of the disposal spin rapidly?}\\
\textsf{\small \textbf{A4:} \red{No}}

\end{tcolorbox}
\label{fig:qual_tea_towel12}
\end{figure*}

\begin{figure*}[!h]
\centering
\includegraphics[width=0.9\linewidth]{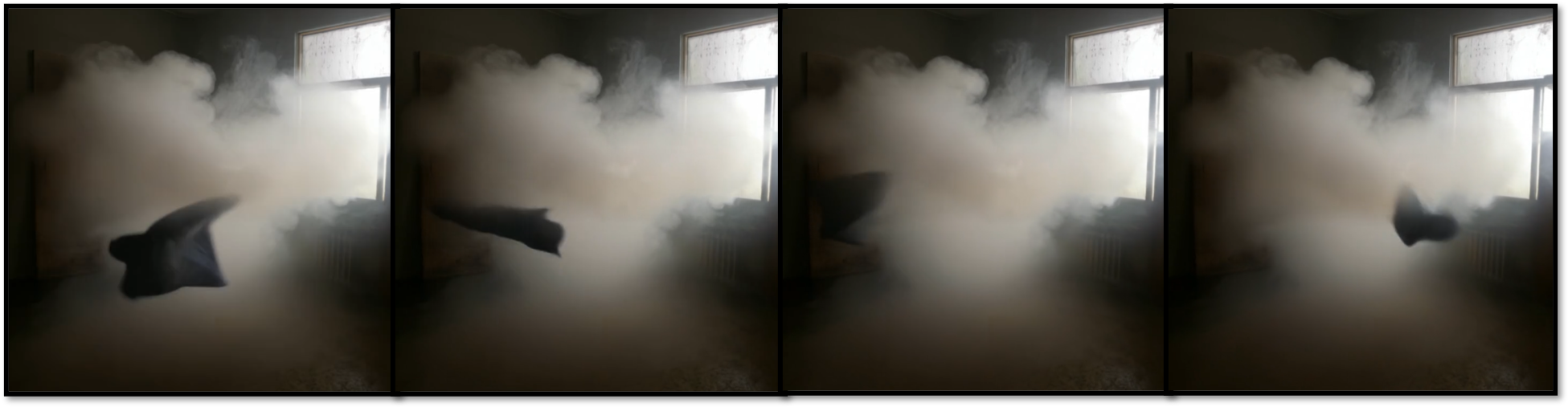}

\caption*{\textbf{Prompt:} \textsf{\small A damp towel is spun rapidly in a room filled with visible smoke, causing the smoke to swirl and dissipate slightly.}}
\vspace{0.5em}

\begin{tcolorbox}[colback=gray!5!white, colframe=gray!50!black, title=Prompt-based Evaluation Questions, boxrule=0.5pt, arc=1mm]

\textbf{Model: MAGI-1}

\vspace{0.5em}
\textsf{\small \textbf{Q1:} Was the air in the room initially still before the spinning?}\\
\textsf{\small \textbf{A1:} \blue{Yes}} 

\vspace{0.3em}
\textsf{\small \textbf{Q2:} Does the damp surface of the towel interact with the smoke particles?}\\
\textsf{\small \textbf{A2:} \red{No}}

\vspace{0.3em}
\textsf{\small \textbf{Q3:} Is the towel's motion described as being driven against air resistance?}\\
\textsf{\small \textbf{A3:} \blue{Yes}}

\vspace{0.3em}
\textsf{\small \textbf{Q4:} Does the rapid movement of the towel generate air currents?}\\
\textsf{\small \textbf{A4:} \blue{Yes}}

\end{tcolorbox}
\label{fig:qual_tea_towel13}
\end{figure*}

\newpage

\begin{figure*}[!h]
\centering
\includegraphics[width=0.9\linewidth]{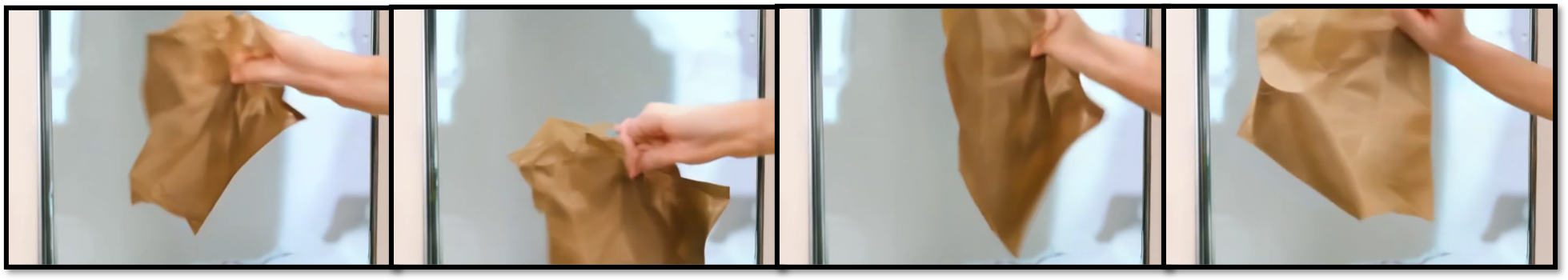}

\caption*{\textbf{Prompt:} \textsf{\small A hand wipes a glass mirror clean using a crumpled brown paper bag.}}

\begin{tcolorbox}[colback=gray!5!white, colframe=gray!50!black, title=Prompt-based Evaluation Questions, boxrule=0.5pt, arc=1mm]

\textbf{Model: CogVideoX (2B)}

\vspace{0.5em}
\textsf{\small \textbf{Q1:} Is the paper bag described as flexible?}\\
\textsf{\small \textbf{A1:} \blue{Yes}} 

\vspace{0.3em}
\textsf{\small \textbf{Q2:} Does the manipulation of the bag into a wad happen *before* it is pressed against the mirror? }\\
\textsf{\small \textbf{A2:} \red{No} }

\vspace{0.3em}
\textsf{\small \textbf{Q3:} Is the described action a wiping motion?}\\
\textsf{\small \textbf{A3:} \red{No}}

\vspace{0.3em}
\textsf{\small \textbf{Q4:} Is pressure applied by the hand onto the paper wad against the mirror? }\\
\textsf{\small \textbf{A4:} \red{No}}

\end{tcolorbox}
\end{figure*}

\begin{figure*}[!h]
\centering
\includegraphics[width=0.9\linewidth]{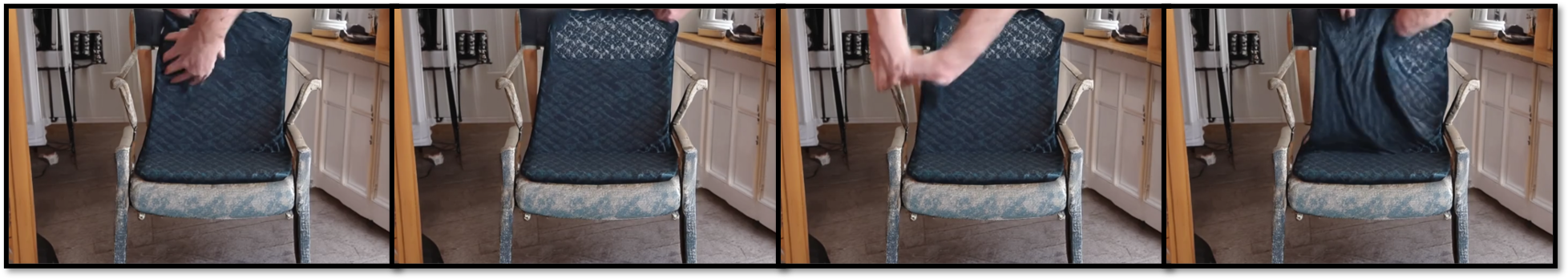}

\caption*{\textbf{Prompt:} \textsf{\small A decorative automobile bucket seat cover is stretched and fitted over an old, worn desk chair.}}

\begin{tcolorbox}[colback=gray!5!white, colframe=gray!50!black, title=Prompt-based Evaluation Questions, boxrule=0.5pt, arc=1mm]

\textbf{Model: CogVideoX (2B)}

\vspace{0.5em}
\textsf{\small \textbf{Q1:} Was the cover pulled over the backrest first? }\\
\textsf{\small \textbf{A1:} \red{No}} 

\vspace{0.3em}
\textsf{\small \textbf{Q2:} Did the application process involve stretching the cover? }\\
\textsf{\small \textbf{A2:} \blue{Yes}}

\vspace{0.3em}
\textsf{\small \textbf{Q3:} Does the cover conform closely to the chair's shape?}\\
\textsf{\small \textbf{A3:} \blue{Yes}}

\vspace{0.3em}
\textsf{\small \textbf{Q4:} Does the chair have underlying padding mentioned?}\\
\textsf{\small \textbf{A4:} \red{No}}

\end{tcolorbox}
\end{figure*}

\newpage

\begin{figure*}[!h]
\centering
\includegraphics[width=0.9\linewidth]{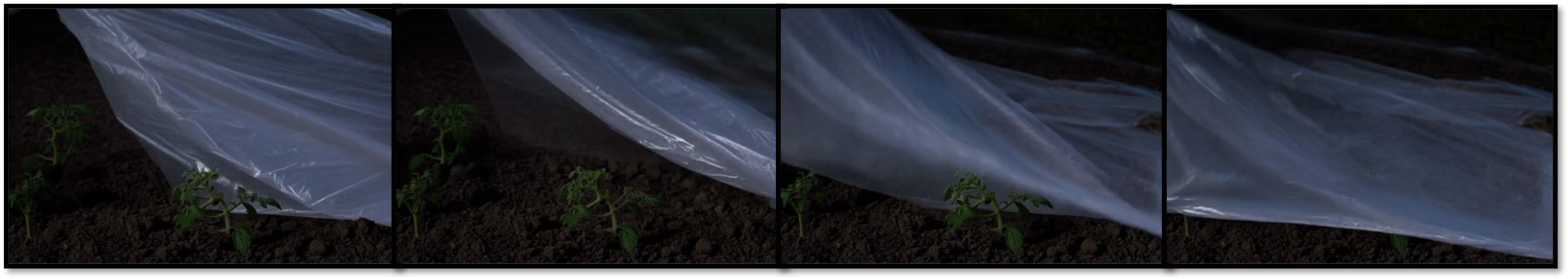}

\caption*{\textbf{Prompt:} \textsf{\small Clear plastic sheeting is carefully draped over several small tomato plants in a garden bed at dusk.}}

\begin{tcolorbox}[colback=gray!5!white, colframe=gray!50!black, title=Prompt-based Evaluation Questions, boxrule=0.5pt, arc=1mm]

\textbf{Model: CogVideoX (5B)}

\vspace{0.5em}
\textsf{\small \textbf{Q1:} Does gravity cause the plastic sheet to move downwards?}\\
\textsf{\small \textbf{A1:} \red{No}} 

\vspace{0.3em}
\textsf{\small \textbf{Q2:} Does the plastic sheet change shape as it settles on the plants?}\\
\textsf{\small \textbf{A2:} \blue{Yes}}

\vspace{0.3em}
\textsf{\small \textbf{Q3:} Does the plastic sheet cover the tomato plants?}\\
\textsf{\small \textbf{A3:} \blue{Yes}}

\vspace{0.3em}
\textsf{\small \textbf{Q4:} Does the flexibility of the plastic allow it to bend over the plants?}\\
\textsf{\small \textbf{A4:} \blue{Yes}}

\end{tcolorbox}
\end{figure*}

\begin{figure*}[!h]
\centering
\includegraphics[width=0.9\linewidth]{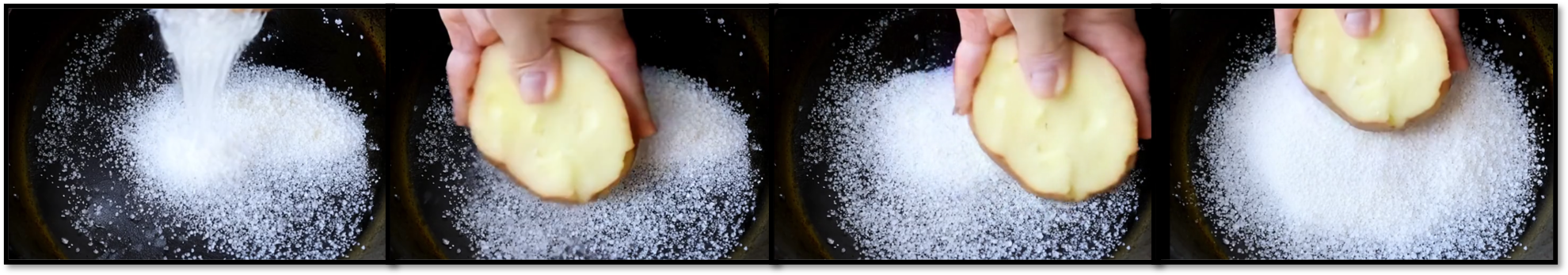}

\caption*{\textbf{Prompt:} \textsf{\small Coarse salt is poured into a dirty cast iron skillet, then scrubbed vigorously with the cut side of a potato half.}}

\begin{tcolorbox}[colback=gray!5!white, colframe=gray!50!black, title=Prompt-based Evaluation Questions, boxrule=0.5pt, arc=1mm]

\textbf{Model: CogVideoX (5B)}

\vspace{0.5em}
\textsf{\small \textbf{Q1:} Does the hand apply pressure while rubbing the potato?}\\
\textsf{\small \textbf{A1:} \red{No}} 

\vspace{0.3em}
\textsf{\small \textbf{Q2:} Is the salt ground between the potato and the skillet surface?}\\
\textsf{\small \textbf{A2:} \red{No}}

\vspace{0.3em}
\textsf{\small \textbf{Q3:} Is consistent pressure applied by the hand? }\\
\textsf{\small \textbf{A3:} \red{No}}

\vspace{0.3em}
\textsf{\small \textbf{Q4:} Is the salt contained within the skillet's interior after pouring? }\\
\textsf{\small \textbf{A4:} \red{No}}

\end{tcolorbox}
\end{figure*}

\clearpage

\begin{figure*}[!h]
\centering
\includegraphics[width=0.9\linewidth]{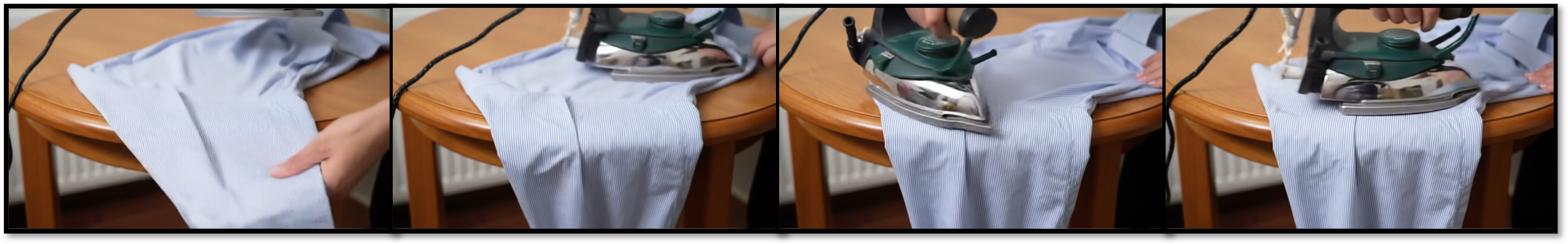}

\caption*{\textbf{Prompt:} \textsf{\small A shirt sleeve draped over the edge of a wooden table is being pressed flat by a moving iron.}}

\begin{tcolorbox}[colback=gray!5!white, colframe=gray!50!black, title=Prompt-based Evaluation Questions, boxrule=0.5pt, arc=1mm]

\textbf{Model: Sora}

\vspace{0.5em}
\textsf{\small \textbf{Q1:} Does the iron apply downward pressure onto the fabric?}\\
\textsf{\small \textbf{A1:} \blue{Yes}} 

\vspace{0.3em}
\textsf{\small \textbf{Q2:} Does the iron transfer heat to the fabric? }\\
\textsf{\small \textbf{A2:} \blue{Yes} }

\vspace{0.3em}
\textsf{\small \textbf{Q3:} Is the soleplate of the iron made of metal?}\\
\textsf{\small \textbf{A3:} \blue{Yes}}

\vspace{0.3em}
\textsf{\small \textbf{Q4:} Is the soleplate of the iron flat? }\\
\textsf{\small \textbf{A4:} \red{No}}

\end{tcolorbox}
\end{figure*}

\begin{figure*}[!h]
\centering
\includegraphics[width=0.9\linewidth]{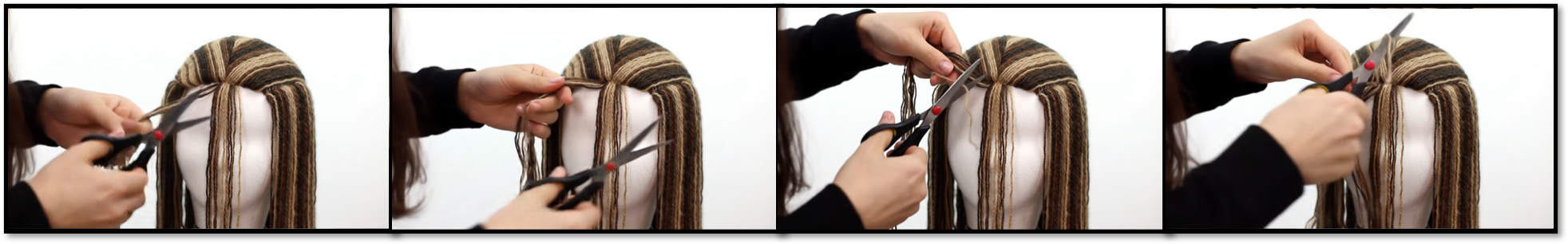}

\caption*{\textbf{Prompt:} \textsf{\small Hands cut strands of yarn with scissors, then apply glue to the ends and press them onto a mannequin head.}}

\begin{tcolorbox}[colback=gray!5!white, colframe=gray!50!black, title=Prompt-based Evaluation Questions, boxrule=0.5pt, arc=1mm]

\textbf{Model: Sora}

\vspace{0.5em}
\textsf{\small \textbf{Q1:} Does squeezing the scissor handles cause the blades to pivot? }\\
\textsf{\small \textbf{A1:} \blue{Yes}} 

\vspace{0.3em}
\textsf{\small \textbf{Q2:} Is the adhesive described as a liquid when applied? }\\
\textsf{\small \textbf{A2:} \red{No}}

\vspace{0.3em}
\textsf{\small \textbf{Q3:} Are multiple strands of yarn held by the hands initially? }\\
\textsf{\small \textbf{A3:} \red{No}}

\vspace{0.3em}
\textsf{\small \textbf{Q4:} Does the cutting action result in loose strands?}\\
\textsf{\small \textbf{A4:} \blue{Yes}}

\end{tcolorbox}
\end{figure*}

\newpage

\begin{figure*}[!h]
\centering
\includegraphics[width=0.9\linewidth]{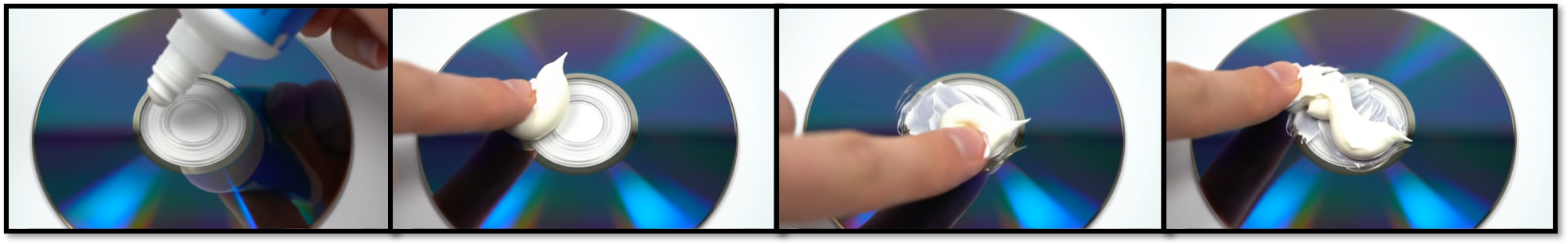}

\caption*{\textbf{Prompt:} \textsf{\small Toothpaste is squeezed onto a scratched CD, then rubbed radially outwards with a finger.}}

\begin{tcolorbox}[colback=gray!5!white, colframe=gray!50!black, title=Prompt-based Evaluation Questions, boxrule=0.5pt, arc=1mm]

\textbf{Model: Veo-2}

\vspace{0.5em}
\textsf{\small \textbf{Q1:} Is the finger used to apply force to spread the paste?}\\
\textsf{\small \textbf{A1:} \blue{Yes}} 

\vspace{0.3em}
\textsf{\small \textbf{Q2:} Is the CD's state of rest maintained throughout the toothpaste application and spreading? }\\
\textsf{\small \textbf{A2:} \red{No} }

\vspace{0.3em}
\textsf{\small \textbf{Q3:} Does the rubbing action cause the toothpaste to cover the scratched area?}\\
\textsf{\small \textbf{A3:} \red{No}}

\vspace{0.3em}
\textsf{\small \textbf{Q4:} Is the CD described as rigid? }\\
\textsf{\small \textbf{A4:} \blue{Yes}}

\end{tcolorbox}
\end{figure*}

\begin{figure*}[!h]
\centering
\includegraphics[width=0.9\linewidth]{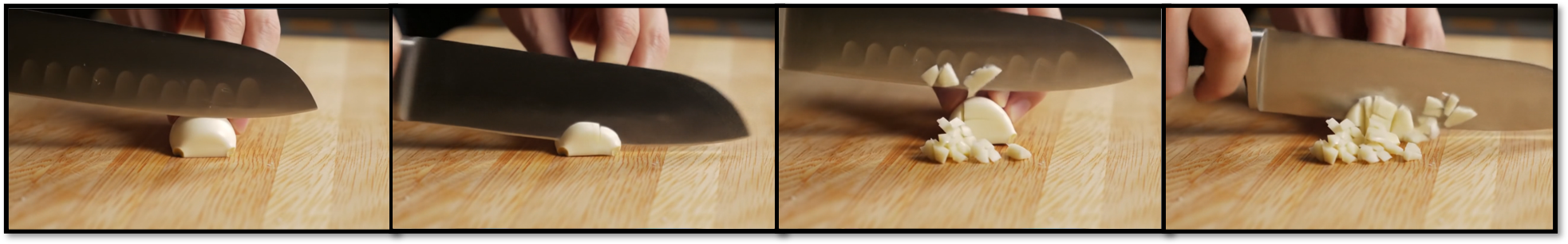}

\caption*{\textbf{Prompt:} \textsf{\small The flat side of a chef's knife is pressed down onto a garlic clove on a cutting board, crushing it.}}

\begin{tcolorbox}[colback=gray!5!white, colframe=gray!50!black, title=Prompt-based Evaluation Questions, boxrule=0.5pt, arc=1mm]

\textbf{Model: Veo-2}

\vspace{0.5em}
\textsf{\small \textbf{Q1:} Does the garlic clove audibly crack? }\\
\textsf{\small \textbf{A1:} \red{No}} 

\vspace{0.3em}
\textsf{\small \textbf{Q2:} Is downward force applied to the knife? }\\
\textsf{\small \textbf{A2:} \blue{Yes}}

\vspace{0.3em}
\textsf{\small \textbf{Q3:} Is the knife described as rigid?}\\
\textsf{\small \textbf{A3:} \blue{Yes}}

\vspace{0.3em}
\textsf{\small \textbf{Q4:} Is the knife positioned parallel to the cutting board surface?}\\
\textsf{\small \textbf{A4:} \red{No}}

\end{tcolorbox}
\end{figure*}

\clearpage
\subsection{Model Generation Specifications}
Table~\ref{tab:models} summarizes, for each evaluated text-to-video (T2V) model, the output resolution, number of frames, video duration, and guidance scale used throughout our experiments. We use these settings consistently when generating both base and enriched videos to ensure a fair comparison across models.

\begin{table}[!h]
\caption{
\textbf{Model generation specifications.} This table lists the generation settings for each text-to-video (T2V) model used in our experiments. For each model, we report the video resolution (in width and height), number of frames, total video duration (in seconds), and the guidance scale used during sampling (if applicable). These settings reflect the default or recommended configurations for each model and were held consistent across evaluations to ensure comparability in visual output and downstream analysis.
}
\centering
\resizebox{\columnwidth}{!}{
\begin{tabular}{@{}l@{$\;\;$}c@{$\;\;$}c@{$\;\;$}c@{$\;\;$}c@{}}
\toprule
\textbf{Model} 
& \textbf{Resolution} & \textbf{\#Frames} & \textbf{Duration} & \textbf{Guidance Scale} \\
\midrule

LTX-Video             & 1216$\times$704 & 121 & 4 & 3 \\
VideoCrafter2         & 512$\times$320 & 16 & 1 & 12 \\
CogVideoX (2B)        & 720$\times$480 & 49 & 6 & 6 \\
CogVideoX (5B)        & 720$\times$480 & 49 & 6 & 6 \\
Wan2.1 (1.3B)         & 832$\times$480 & 33 & 6 & 5 \\
Wan2.1 (14B)          & 832$\times$480 & 33 & 6 & 5 \\
MAGI-1                & 720$\times$720 & 48 & 2 & 7.5 \\
HunyuanVideo          & 960$\times$544 & 129 & 5 & 6 \\
Cosmos (7B)           & 1280$\times$704 & 121 & 5 & 7 \\
Cosmos (14B)          & 1280$\times$704 & 121 & 5 &  7\\
\midrule
Sora                  & 1280$\times$720 & 150 & 5 & --\\
Veo-2                 & 1280$\times$720 & 120 & 5 & --\\
\bottomrule
\end{tabular}
}
\label{tab:models}
\end{table}

\subsection{Effect of Caption Detail on Model Performance: Short vs. Dense Captions}
Caption detail strongly affects the accuracy and reliability of VLM-based evaluation. Short captions tend toward high-level summaries and omit subtle but critical details, e.g. precise object interactions or slight motions, which leads to incomplete or inaccurate physical commonsense judgments. When queried directly, VLMs may also hallucinate or exaggerate minor cues, further degrading evaluation quality. Dense captioning instead records fine-grained observations without suggestive prompting. We use AuroraCap for its ability to generate detailed, multi-faceted descriptions: for each video we produce one general-purpose caption and seven dimension-specific captions aligned with our ontology (e.g., spatial layout, force dynamics, material behavior). {A yes/no question is marked correct if any of the eight pooled captions supply the required evidence. Using complementary views reduces dependence on any single caption and provides broader coverage of the visual evidence relevant to physical reasoning.}

\begin{figure}[b!]
    \centering
    \includegraphics[width=\linewidth]{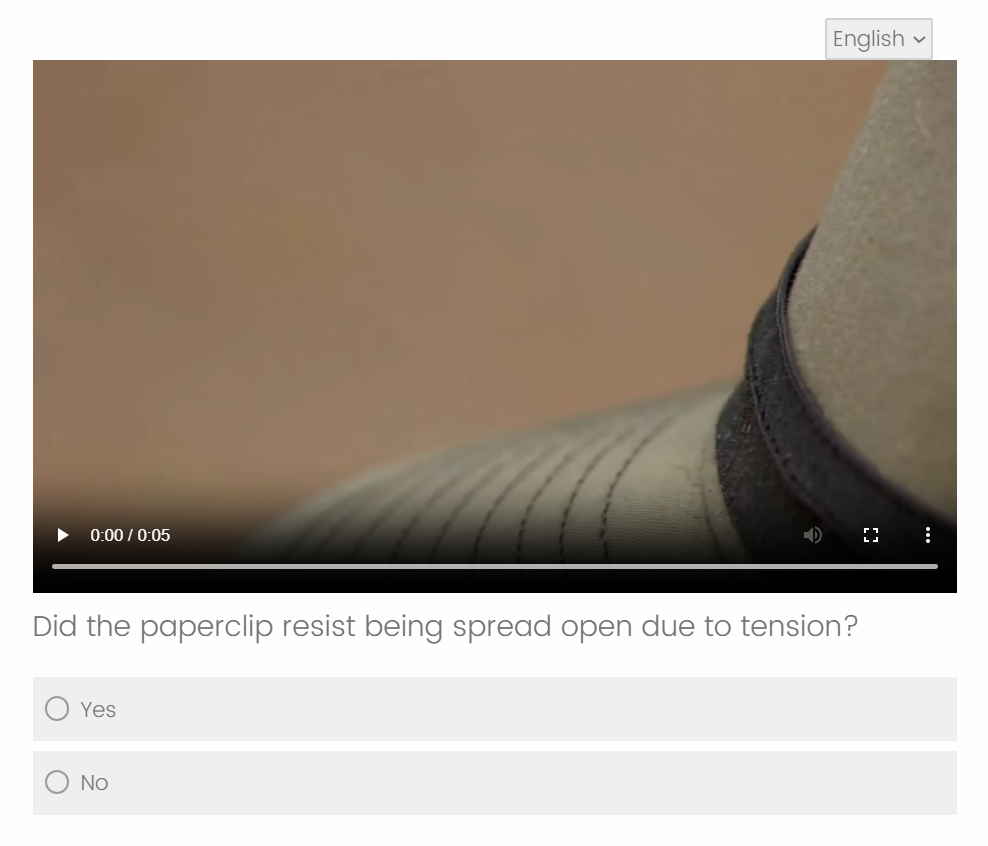}
    \caption{\textbf{User Study Interface.} Our web-based interface for the human study mimicked the auto-eval QA structure. Participants watched a video, then answered five Yes/No questions drawn from the same QA bank used in automatic evaluation. The design enforced randomization across videos and questions, while masking model and prompt identity to reduce bias.}
    \label{fig:user_interface}
\end{figure}

\begin{figure*}[!h]
    \centering
    \includegraphics[width=\textwidth]{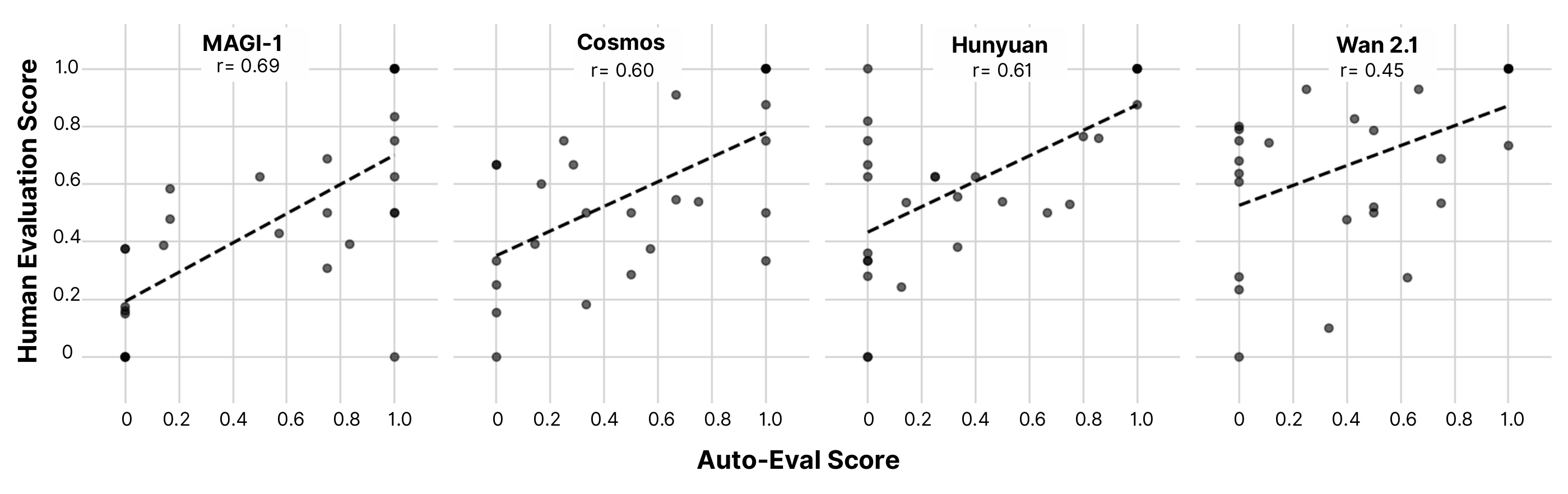}
    \caption{\textbf{Pearson correlation between human and automatic evaluations.} {Each subplot compares scores over 30 prompts and reports a linear fit for one open-source generator: MAGI-1 ($r = 0.69$), Cosmos-14B ($r = 0.60$), HunyuanVideo ($r = 0.61$), and Wan2.1-14B ($r = 0.45$).}}
    \label{fig:correlation}
\end{figure*}

{Sec.~\ref{sec:robustness:controls} further evaluates caption recall and false-positive behavior.}

\section{User Study Details and Human Evaluation }
\label{app:human_eval}
To verify that our automatic, caption‐based evaluation (auto-eval pipeline) aligns with human judgment, we carried out a user study via Qualtrics. We selected the four top‐performing open-source models, namely Cosmos‐14B, Wan2.1-14B, MAGI‐1, and HunyuanVideo, and sampled 30 prompts (out of 383 upsampled prompts) at random. Each prompt was used to generate one video per model, yielding 120 videos in total.

\parahead{Procedure} For each video, we used the yes/no questions produced by our auto‐eval pipeline and presented them in a survey interface built with Qualtrics that mimics the automated QA layout (Figure~\ref{fig:user_interface}). Participants saw each video alongside five randomly assigned questions, answered in a Yes/No format. Videos and questions were fully randomized to prevent any inference of prompt or model identity. Before beginning, participants read brief instructions on interpreting the questions; we then collected non‐identifying metadata (age range and self‐reported familiarity with deep learning).

\parahead{Participants} Fifteen individuals took part, all volunteers, collectively providing 1{,}340 binary (Yes/No) responses across the 120 videos. Owing to the randomized assignment and modest pool size, some questions were answered only once, while others received multiple responses.

\parahead{Results and Analysis} We computed Pearson correlation coefficients to measure alignment between auto‐eval scores and human judgments. As expected, correlation rose with the number of annotators. For MAGI‐1, questions answered by only a single participant gave $r{=}0.47$, while those answered by at least three reached $r{=}0.69$. Figure~\ref{fig:correlation} plots these correlations with linear fits for MAGI-1, Cosmos-14B, HunyuanVideo, and Wan2.1-14B. Although Wan2.1-14B attains the lowest model-level correlation ($r{=}0.45$, Table~\ref{tab:human-corr-vlmqa}), this remains competitive given the task’s ambiguity and binary scoring. VideoPhy-2 reports an average Pearson correlation of $r{=}0.285$ for the prior VideoCon-Physics evaluator introduced in VideoPhy and $r{=}0.42$ for its own AutoEval model on unseen prompts~\citep{bansal2024videophy,bansal2025videophy}, whereas our evaluator reaches $r{=}0.69$ at best in our human study. Together, these findings indicate that the caption$\rightarrow$LLM design is a reliable proxy for human judgment, and that adding annotators strengthens the alignment further. Inter-annotator agreement is also consistent across generators. For items with at least three responses, Krippendorff's $\alpha$ is 0.74 for MAGI-1, 0.77 for Cosmos-14B, 0.71 for HunyuanVideo, and 0.74 for Wan2.1-14B; with at least four responses, the corresponding values are 0.80, 0.85, 0.75, and 0.80.

\section{Robustness of the Automatic Evaluation Pipeline}
\label{app:robustness}

We designed our automatic caption-based evaluation (``auto-eval'') with robustness as a primary objective. This section compiles the empirical evidence for its reliability and the analyses we used to stress-test each component.

\subsection{Cross–Judge Stability}
\label{sec:robustness:judge}
To test whether results depend on a particular judge model, we replaced Gemini-Flash with DeepSeek. The induced ranking of T2V models is highly stable (Spearman $\rho=0.98$ over the full model list), showing that the pipeline measures differences between videos rather than idiosyncrasies of the judge (Table~\ref{tab:cross-judge-stability}).

\begin{table}[!h]
\centering
\small
\caption{\textbf{Cross–judge stability of our evaluation pipeline.} Replacing Gemini--Flash with DeepSeek yields nearly identical model rankings, indicating the performance signal is not judge–specific but emerges from our language-grounded evaluation design.}
\setlength{\tabcolsep}{8pt}
\resizebox{\linewidth}{!}{
\begin{tabular}{@{}l@{$\;$}c@{$\;$}c@{}}
\toprule
\textbf{Model} & \textbf{Gemini--Flash Rank} & \textbf{DeepSeek Rank} \\
\midrule
Cosmos (14B)      & 1  & 1  \\
Cosmos (7B)       & 2  & 3  \\
Wan2.1 (14B)      & 3  & 2  \\
MAGI--1           & 4  & 4  \\
Wan2.1 (1.3B)     & 5  & 5  \\
HunyuanVideo      & 6  & 6  \\
CogVideoX (2B)    & 7  & 7  \\
VideoCrafter2     & 8  & 9  \\
CogVideoX (5B)    & 9  & 8  \\
LTX--Video        & 10 & 10 \\
\multicolumn{3}{l}{\scriptsize $\quad$Spearman $\rho = 0.98$ (rank stability)} \\
\bottomrule
\end{tabular}
}
\label{tab:cross-judge-stability}
\end{table}

\subsection{Captioner Analysis and Cross–Captioner Stability}
\label{sec:robustness:captioner}
To assess whether PhysVidBench depends on a particular captioning model, we re-ran the full pipeline with two alternative captioners, Qwen2.5-VL-72B-Instruct-AWQ and Qwen3-VL-30B-A3B-Instruct, in place of AuroraCap. Agreement is strong across all ten open-source T2V models: Spearman rank correlations between captioners range from 0.80 to 1.00, so the relative ordering of models is effectively unchanged. Cosmos-14B, Wan2.1-14B, MAGI-1, and HunyuanVideo remain the top-performing models under every captioning variant, indicating that the evaluator captures properties of the videos rather than of any single captioner. Pearson correlations of 0.74--0.98 further confirm that absolute scores move coherently across captioners. Table~\ref{tab:captioner-corr} summarizes these results.

\begin{table}[!h]
\caption{\textbf{Cross-captioner agreement.} Lower triangle: Spearman rank correlation; upper triangle: Pearson correlation, over model-level accuracies. Agreement is high between the two Qwen captioners and moderate-to-high for AuroraCap, indicating PhysVidBench scores are largely stable across captioners.}
\centering
\small
\begin{tabular}{l@{$\;$}c@{$\;$}c@{$\;$}c}
\toprule
 & {AuroraCap} & {Qwen-2.5-VL} & {Qwen-3-VL} \\
\midrule
AuroraCap     & \cellcolor{gray!12}{--} & 0.83 & 0.74 \\
Qwen-2.5-VL   & 0.80 & \cellcolor{gray!12}{--} & 0.98 \\
Qwen-3-VL     & 0.80 & 1.00 & \cellcolor{gray!12}{--} \\
\bottomrule
\end{tabular}
\label{tab:captioner-corr}
\end{table}

These results show that PhysVidBench is not captioner-dependent: swapping in significantly different captioners, including models with different vision--language backbones and training corpora, yields essentially the same model rankings and similar absolute scores. {Together with the multi-caption design, this consistency indicates that the evaluation is robust to the choice of captioner.}

\subsection{Caption Evidence Recall and Mismatched-Video Control}
\label{sec:robustness:controls}
We first measure caption-stage recall on the 120 videos used in
our human study. Two expert annotators identify the physically
relevant events visible in each video and check whether each event
appears in at least one of the eight generated captions. An event
is counted as a false negative only when it is visibly present but
absent from every caption. This analysis yields 98.5\% event-level
recall, corresponding to a 1.5\% false-negative rate.

We then test the complementary false-positive pathway by pairing
each question set with a video generated by the same model from a
different, randomly selected prompt. We apply this control to all
4,123 questions for every evaluated model. As shown in
Table~\ref{tab:mismatched_control}, positive-response rates remain
between 2.8\% and 5.1\% across all models and physical-commonsense
dimensions, substantially below the matched accuracies in Table~3.
This consistent collapse indicates that the evaluator relies on
evidence from the corresponding video rather than question wording
or a general tendency to answer ``Yes.''

\begin{table*}[!h]
\centering
\small
\setlength{\tabcolsep}{5pt}
\caption{
Mismatched-video control by physical commonsense dimension.
Each question set is paired with a video generated by the
corresponding model from a different prompt. Values report
positive-response rates (\%). AU = Action \& Procedural
Understanding, FM = Force and Motion, FP = Fundamental Physics,
MT = Material Interaction \& Transformation, OP = Object
Properties \& Affordances, SR = Spatial Reasoning, and
TD = Temporal Dynamics.
}
\label{tab:mismatched_control}
\begin{tabular}{lcccccccc}
\toprule
Model & AU & FM & FP & MT & OP & SR & TD & Average \\
\midrule
LTX-Video
& 3.4 & 3.7 & 3.1 & 4.0 & 4.5 & 3.3 & 2.9 & 3.6 \\

VideoCrafter2
& 3.2 & 3.5 & 3.8 & 3.6 & 4.2 & 3.0 & 2.8 & 3.4 \\

CogVideoX (2B)
& 3.8 & 4.0 & 3.6 & 4.1 & 4.7 & 3.5 & 3.1 & 3.8 \\

CogVideoX (5B)
& 3.5 & 3.8 & 3.4 & 3.9 & 4.4 & 3.2 & 3.0 & 3.6 \\

Wan2.1 (1.3B)
& 4.0 & 4.3 & 3.9 & 4.4 & 4.9 & 3.8 & 3.4 & 4.1 \\

Wan2.1 (14B)
& 4.2 & 4.5 & 4.1 & 4.6 & 5.1 & 4.0 & 3.6 & 4.3 \\

MAGI-1
& 3.9 & 4.2 & 3.8 & 4.3 & 4.8 & 3.7 & 3.3 & 4.0 \\

HunyuanVideo
& 3.7 & 4.0 & 3.6 & 4.2 & 4.6 & 3.5 & 3.2 & 3.8 \\

Cosmos (7B)
& 4.1 & 4.4 & 4.0 & 4.5 & 5.0 & 3.9 & 3.5 & 4.2 \\

Cosmos (14B)
& 4.3 & 4.6 & 4.2 & 4.7 & 5.1 & 4.1 & 3.7 & 4.4 \\

Sora
& 3.8 & 4.1 & 3.7 & 4.3 & 4.8 & 3.6 & 3.2 & 3.9 \\

Veo-2
& 4.0 & 4.3 & 3.9 & 4.4 & 4.9 & 3.8 & 3.4 & 4.1 \\
\bottomrule
\end{tabular}
\end{table*}

Because unrelated videos may occasionally satisfy individual
questions by chance, these residual positive responses provide a
conservative estimate of the evaluator's false-positive tendency.
Together, the high caption recall and low mismatched-video response
rates show that relevant evidence is rarely omitted, while unrelated
visual evidence is rarely accepted.

\subsection{Error-Guided Prompt Refinement}
\label{sec:robustness:dev}
Beyond passive scoring, we used auto–eval as a diagnostic signal in a training–free self–refinement loop: detect failed dimensions $\rightarrow$ prompt rewrite $\rightarrow$ regenerate $\rightarrow$ re–evaluate. {On 100 prompts, CogVideoX-2B and CogVideoX-5B improved from $21.6{\to}32.7$ (+11.1 percentage points) and $17.8{\to}29.7$ (+11.9 percentage points), respectively. This iterative experiment is separate from one-shot prompt enrichment and was evaluated only on these two models.}

\begin{mainbox}{An Illustrative Example}
The example below shows how PhysVidBench feedback guided a revised prompt that led to a visibly more plausible generation: \vspace{0.4em}\\
\textbf{Base Prompt:} \textit{A screwdriver tip is placed on the lid of a metal can, and a hammer taps the screwdriver handle, puncturing the lid.}\\
\textbf{Refined Prompt (Guided by Feedback):} \textit{A person grips a hammer, delivering a sharp, controlled tap to the rigid upper end of a vertically positioned screwdriver with a solid shaft; the heavy hammer's impact force travels down the screwdriver's shaft, concentrating at its tip, which is pressed firmly against the surface of a sealed metal can lid, causing the thin metal to yield and puncture after the hammer tap, overcoming the lid's structural integrity.}
 
\end{mainbox}

{Taken together, these analyses show that our evaluation pipeline (i) aligns more strongly with human judgments than the other evaluated automatic methods, (ii) yields stable rankings across different judges and captioners, (iii) achieves high caption-event recall on the annotated human-study videos, (iv) shows a sharp drop in positive responses under the mismatched-video control for Wan2.1-14B, and (v) can provide diagnostic feedback for prompt refinement. Together, these results support the robustness of the caption-grounded evaluation pipeline for physical commonsense video generation.}

{For dimension-level uncertainty, we perform 10,000 prompt-level bootstrap resamples. Each sampled prompt retains all of its associated QA items, preserving dependence among questions derived from the same prompt. Table~\ref{tab:dimension-ci} reports the resulting 95\% interval half-widths.}

\begin{table*}[!h]
\caption{\textbf{Prompt-level bootstrap 95\% confidence intervals by physical dimension.} Values are upsampled-prompt caption-grounded QA accuracy percentages $\pm$ the 95\% bootstrap interval half-width. AU = Action \& Procedural Understanding, FM = Force and Motion, FP = Fundamental Physics, MT = Material Interaction \& Transformation, OP = Object Properties \& Affordances, SR = Spatial Reasoning, and TD = Temporal Dynamics. Prompts, rather than individual QA items, are resampled.}
\centering
\scriptsize
\setlength{\tabcolsep}{3.5pt}
\resizebox{\textwidth}{!}{%
\begin{tabular}{lccccccc}
\toprule
\textbf{Model} & \textbf{AU} & \textbf{FM} & \textbf{FP} & \textbf{MT} & \textbf{OP} & \textbf{SR} & \textbf{TD} \\
\midrule
LTX-Video          & 20.7 $\pm$ 3.40 & 19.9 $\pm$ 3.10 & 18.6 $\pm$ 3.15 & 16.9 $\pm$ 3.75 & 25.0 $\pm$ 2.95 & 17.3 $\pm$ 2.80 & 13.7 $\pm$ 4.35 \\
VideoCrafter2      & 17.7 $\pm$ 2.90 & 19.4 $\pm$ 2.95 & 23.3 $\pm$ 3.75 & 18.8 $\pm$ 4.15 & 26.6 $\pm$ 2.75 & 15.8 $\pm$ 2.55 & 12.6 $\pm$ 4.05 \\
CogVideoX (2B)     & 22.6 $\pm$ 3.15 & 24.6 $\pm$ 2.95 & 23.8 $\pm$ 3.70 & 22.6 $\pm$ 4.10 & 28.9 $\pm$ 2.75 & 19.6 $\pm$ 2.85 & 16.8 $\pm$ 4.60 \\
CogVideoX (5B)     & 18.9 $\pm$ 3.00 & 19.6 $\pm$ 2.95 & 21.3 $\pm$ 3.40 & 18.1 $\pm$ 3.90 & 25.6 $\pm$ 2.75 & 17.9 $\pm$ 2.90 & 12.6 $\pm$ 4.00 \\
Wan2.1 (1.3B)      & 30.1 $\pm$ 3.55 & 29.0 $\pm$ 3.30 & 28.5 $\pm$ 4.00 & 28.9 $\pm$ 4.60 & 35.2 $\pm$ 2.95 & 27.4 $\pm$ 3.25 & 18.7 $\pm$ 4.85 \\
Wan2.1 (14B)       & 33.6 $\pm$ 3.65 & 31.6 $\pm$ 3.50 & 32.6 $\pm$ 4.35 & 32.9 $\pm$ 4.80 & 39.0 $\pm$ 2.95 & 29.6 $\pm$ 3.30 & 23.7 $\pm$ 5.20 \\
MAGI-1             & 27.3 $\pm$ 3.35 & 26.6 $\pm$ 3.25 & 29.4 $\pm$ 4.20 & 28.5 $\pm$ 4.45 & 37.4 $\pm$ 3.00 & 30.4 $\pm$ 3.30 & 19.1 $\pm$ 4.85 \\
HunyuanVideo       & 26.7 $\pm$ 3.45 & 26.2 $\pm$ 3.15 & 28.1 $\pm$ 3.95 & 32.3 $\pm$ 4.75 & 36.1 $\pm$ 3.00 & 23.9 $\pm$ 3.10 & 21.8 $\pm$ 5.25 \\
Cosmos (7B)        & 32.5 $\pm$ 3.85 & 32.8 $\pm$ 3.70 & 33.6 $\pm$ 4.15 & 36.1 $\pm$ 4.90 & 40.1 $\pm$ 2.90 & 31.6 $\pm$ 3.55 & 21.4 $\pm$ 5.30 \\
Cosmos (14B)       & 35.6 $\pm$ 3.50 & 33.3 $\pm$ 3.35 & 31.7 $\pm$ 3.90 & 37.4 $\pm$ 4.60 & 40.9 $\pm$ 2.90 & 32.5 $\pm$ 3.25 & 21.0 $\pm$ 5.05 \\
Sora               & 28.6 $\pm$ 3.20 & 30.0 $\pm$ 3.25 & 29.9 $\pm$ 4.20 & 33.1 $\pm$ 4.80 & 37.0 $\pm$ 2.60 & 24.8 $\pm$ 2.90 & 16.7 $\pm$ 4.55 \\
Veo-2              & 34.9 $\pm$ 3.35 & 33.8 $\pm$ 3.40 & 31.7 $\pm$ 4.20 & 36.0 $\pm$ 4.55 & 38.4 $\pm$ 2.85 & 29.7 $\pm$ 3.30 & 19.5 $\pm$ 5.85 \\
\bottomrule
\end{tabular}}
\label{tab:dimension-ci}
\end{table*}

\begin{table*}[!h]
\caption{{\textbf{Bootstrap 95\% confidence intervals for model accuracies using upsampled prompts, base prompts, and their paired differences.} Positive intervals in the “Difference’’ column indicate statistically significant improvements from prompt enrichment.}}
\centering
\small
\setlength{\tabcolsep}{6pt}
\begin{tabular}{lccc}
\toprule
\textbf{Model} & \textbf{Upsampled Prompts} & \textbf{Base Prompts} & \textbf{Difference} \\
\midrule
CogVideoX (2B)  & [0.2341, 0.2784] & [0.2042, 0.2458] & [0.0068, 0.0556] \\
CogVideoX (5B)  & [0.1883, 0.2317] & [0.1735, 0.2162] & [-0.0097, 0.0399] \\
Wan2.1 (1.3B)   & [0.2831, 0.3309] & [0.2525, 0.2977] & [0.0075, 0.0561] \\
Wan2.1 (14B)    & [0.3165, 0.3658] & [0.2982, 0.3468] & [-0.0078, 0.0448] \\
VideoCrafter2   & [0.2012, 0.2405] & [0.1832, 0.2237] & [-0.0013, 0.0355] \\
LTX-Video       & [0.1883, 0.2325] & [0.1604, 0.2009] & [0.0071, 0.0526] \\
Cosmos (7B)     & [0.3248, 0.3762] & [0.2436, 0.2910] & [0.0555, 0.1109] \\
Cosmos (14B)    & [0.3398, 0.3847] & [0.2712, 0.3117] & [0.0463, 0.0956] \\
HunyuanVideo    & [0.2769, 0.3248] & [0.2675, 0.3153] & [-0.0168, 0.0352] \\
MAGI-1          & [0.3013, 0.3505] & [0.2489, 0.2961] & [0.0276, 0.0795] \\
\bottomrule
\end{tabular}
\label{tab:bootstrap1}
\end{table*}

\begin{table*}[!h]
\caption{\textbf{Bootstrap 95\% confidence intervals across model-based difficulty groups for upsampled prompts.} {Each interval reflects model accuracy uncertainty within the Very Hard, Hard, and Medium groups. The non-overlapping intervals show a consistent separation in model performance across the three groups. Together with the taxonomy analysis in Table~\ref{tab:taxonomy-scores}, this separation is associated with increasing action complexity, scientific-principle complexity, and object functionality from Medium to Very Hard.}}
\centering
\setlength{\tabcolsep}{6pt}
\small
\begin{tabular}{lccc}
\toprule
{\textbf{Model}} & {\textbf{Very Hard Prompts}} & {\textbf{Hard Prompts}} & {\textbf{Medium Prompts}} \\
\midrule
{CogVideoX (2B)}  & {[0.0961, 0.1365]} & {[0.2551, 0.3087]} & {[0.4429, 0.5555]} \\
{CogVideoX (5B)} & {[0.0897, 0.1382]} & {[0.1934, 0.2481]} & {[0.3248, 0.4596]} \\
{Wan2.1 (1.3B)}    & {[0.1227, 0.1697]} & {[0.3103, 0.3663]} & {[0.5210, 0.6285]} \\
{Wan2.1 (14B)}     & {[0.1547, 0.2060]} & {[0.3419, 0.3998]} & {[0.5645, 0.6642]} \\
{VideoCrafter2} & {[0.0876, 0.1275]} & {[0.2074, 0.2553]} & {[0.3958, 0.4938]} \\
{LTX-Video}     & {[0.0729, 0.1110]} & {[0.2009, 0.2573]} & {[0.3489, 0.4742]} \\
{Cosmos (7B)}     & {[0.1552, 0.2182] }& {[0.3574, 0.4186]} & {[0.5488, 0.6606]} \\
{Cosmos (14B)}    & {[0.1951, 0.2487]} & {[0.3651, 0.4201]} & {[0.5528, 0.6319]} \\
{HunyuanVideo} & {[0.1336, 0.1859]} & {[0.3056, 0.3646]} & {[0.4667, 0.5830]} \\
{MAGI-1}          & {[0.1447, 0.1911]} & {[0.3229, 0.3830]} & {[0.5483, 0.6589]} \\
\bottomrule
\end{tabular}
\label{tab:ci_upsampled_difficulty}
\end{table*}

\begin{table*}[!h]
\caption{\textbf{Bootstrap 95\% confidence intervals across model-based difficulty groups for base prompts.} {The non-overlapping intervals across Medium, Hard, and Very Hard groups show that the same separation in model performance persists for base prompts, indicating that the observed difficulty structure is not driven by prompt enrichment.}}
\centering
\setlength{\tabcolsep}{6pt}
\small
\begin{tabular}{lccc}
\toprule
{\textbf{Model}} & {\textbf{Very Hard Prompts}} & {\textbf{Hard Prompts}} & {\textbf{Medium Prompts}} \\
\midrule
{CogVideoX (2B)}  & {[0.0901, 0.1292]} & {[0.2251, 0.2775]} & {[0.3619, 0.4806]} \\
{CogVideoX (5B)}  & {[0.0710, 0.1151]} & {[0.1856, 0.2419]} & {[0.3026, 0.4287]} \\
{Wan2.1 (1.3B)}    & {[0.1110, 0.1577]} & {[0.2650, 0.3189]} & {[0.4849, 0.5849]} \\
{Wan2.1 (14B)}     & {[0.1543, 0.2127]} & {[0.3140, 0.3729]} & {[0.5065, 0.6238]} \\
{VideoCrafter2} & {[0.0729, 0.1105]} & {[0.1949, 0.2448]} & {[0.3501, 0.4544]} \\
{LTX-Video}     & {[0.0604, 0.0979]} & {[0.1601, 0.2083]} & {[0.3358, 0.4560]} \\
{Cosmos (7B)} & {[0.1129, 0.1602]} & {[0.2556, 0.3155]} & {[0.4314, 0.5623]} \\
{Cosmos (14B)}    & {[0.1689, 0.2254]}& {[0.2810, 0.3303]} & {[0.4147, 0.5076]} \\
{HunyuanVideo} & {[0.1108, 0.1558]} & {[0.2886, 0.3493]} & {[0.5129, 0.6097]} \\
{MAGI-1}          & {[0.1126, 0.1621]} & {[0.2675, 0.3264]} & {[0.4401, 0.5545]} \\
\bottomrule
\end{tabular}
\label{tab:ci_base_difficulty}
\end{table*}

\subsection{Statistical Significance and Confidence Interval Analysis}
Bootstrap confidence intervals provide evidence that PhysVidBench yields statistically {interpretable} evaluation signals. Using 10K resamples, we find that the confidence interval on the enriched-minus-base accuracy difference lies entirely above zero for six of ten models. For CogVideoX-5B, Wan2.1-14B, VideoCrafter2, and HunyuanVideo, the interval includes zero. {Overall, prompt enrichment yields modest, model-dependent gains, with statistically supported improvements for six of the ten models.}

{For the model-based difficulty groups, the Medium, Hard, and Very Hard intervals do not overlap for any model under either base or upsampled prompts (Tables~\ref{tab:ci_upsampled_difficulty} and~\ref{tab:ci_base_difficulty}), showing a consistent separation in model performance across the three groups. Together with the taxonomy analysis in Table~\ref{tab:taxonomy-scores}, where action complexity, scientific-principle complexity, and object functionality increase monotonically from Medium to Very Hard, this indicates that the performance-based grouping aligns with increasingly demanding physical-reasoning characteristics rather than arbitrary score partitions.}

{Overall, the bootstrap analysis quantifies uncertainty in the reported results, supports a consistent distinction across the model-based difficulty groups, and shows that the effect of prompt enrichment is modest and model-dependent.}

\section*{Acknowledgements}
We thank all reviewers for their valuable comments and suggestions, which helped improve the final version of this work. This research was supported in part by the KUIS AI Center Research Awards to Enes Sanli. Aykut Erdem acknowledges funding from the TÜBİTAK 2247-A Program under Award 123C550. The authors also gratefully acknowledge the use of Google Cloud Platform credits provided through Google's Gemini Academic Program Award. Generative AI assistants were used for language polishing, paraphrasing, and drafting assistance. All scientific content, analyses, claims, citations, and final text were reviewed and verified by the authors.

\end{document}